\documentclass{article}

\usepackage[preprint]{neurips_2026}
\usepackage{graphicx}
\usepackage{wrapfig}
\usepackage{booktabs}
\usepackage{tabularx}
\usepackage{makecell}

\usepackage[utf8]{inputenc} 
\usepackage[T1]{fontenc}    
\usepackage{hyperref}       
\usepackage{url}            
\usepackage{booktabs}       
\usepackage{amsfonts}       
\usepackage{nicefrac}       
\usepackage{microtype}      
\usepackage{xcolor}         
\usepackage[most]{tcolorbox}
\usepackage{enumitem}
\usepackage{titletoc}
\usepackage{cleveref}       
\usepackage{subcaption}
\usepackage{multirow}

\tcbset{
  takeawaybox/.style={
    colback=gray!15,
    colframe=gray!60,
    boxrule=0.8pt,
    arc=3pt,
    left=6pt,
    right=6pt,
    top=2pt,
    bottom=2pt
  }
}

\title{A Matter of Interest:\\ Understanding Interestingness Judgments of Math Problems in Humans and Language Models}

\author{Shubhra Mishra$^{1}$, Yuka Machino$^{2,3,4}$, Gabriel Poesia$^{5}$, Albert Jiang$^{4,6}$, Joy Hsu$^{2}$,\\
\textbf{Adrian Weller$^4$}, \textbf{Challenger Mishra$^4$}, \textbf{David Broman$^1$}, \textbf{Joshua B. Tenenbaum$^{3}$}, \\ \textbf{Mateja Jamnik$^{4,}$\thanks{Collaborative advising.}}\,\,, \textbf{Cedegao E. Zhang$^{3,*}$}, \textbf{Katherine M. Collins$^{3,4,*}$}\\
KTH Royal Institute of Technology$^1$, Stanford University$^2$, \\Massachusetts Institute of Technology$^{3}$, University of Cambridge$^4$, \\ Kempner Institute at Harvard University$^5$, Mistral AI$^6$
\\
\texttt{shubhram@kth.se, \{cedzhang, katiemc\}@mit.edu}
}

\begin{document}

\maketitle

\begin{abstract}
  The evolution of mathematics is shaped importantly by \emph{interestingness}: researchers choose which problems to pursue, and students choose which problems to engage with, based on expectations of interest and challenge. As AI systems, particularly large language models (LLMs) that operate flexibly over natural language and formal mathematics, are increasingly used in mathematics research and education, it becomes crucial to characterize how closely their judgments align with people from different mathematical backgrounds. We study whether LLMs align with human interestingness judgments by comparing LLM ratings with those of two populations, crowdsourced participants with college math experience and International Math Olympiad competitors. Although many LLMs broadly agree with human notions of interestingness, they largely fail to match the \emph{distribution} of human judgments. They also weakly align with \emph{why} humans find problems interesting, with low correlation to human-selected rationales. Finally, we evaluate LLMs’ ability to \emph{generate} interesting problems and find that, after filtering for validity, LLMs are able to generate engaging problems. We conclude with takeaways, including the need for multi-LLM human-AI collaborative systems, that highlight both the promise and current limits of LLMs as partners in mathematical reasoning.
\end{abstract}

\section{Introduction}
\label{sec:introduction}
Mathematical reasoning involves not just solving problems but also judging whether a problem is worth solving. Large language models (LLMs) and large reasoning models (LRMs) have substantially advanced in their ability to solve mathematics problems: they have gone from struggling to solve grade school mathematics problems to now achieving gold-medal-level performance at the International Mathematical Olympiad (IMO)~\cite{cobbe2021trainingverifierssolvemath, hendrycks2021measuringmathematicalproblemsolving, DeepMind2025GeminiGold, openai2025o3o4mini, achim2025aristotleimolevelautomatedtheorem}. While this progress is impressive, \textit{problems are often given to models by humans}. Even computer-assisted discoveries like improved bounds on the CapSet problem (by FunSearch~\cite{veličković2024amplifyinghumanperformancecombinatorial}), better matrix multiplication algorithms (by AlphaTensor~\cite{AlphaTensor2022}), or the improved bound on the kissing number (by AlphaEvolve~\cite{novikov2025alphaevolvecodingagentscientific}) have ultimately depended on human-posed targets or carefully designed heuristics. 

It is unclear whether LLMs and LRMs can adequately judge and select which problems are worth solving at all~\citep{collins2025evaluating}. This is important because problem selection is crucial for many potential applications of LLMs from education (e.g., proposing interesting problems and examples to students) to automated mathematical discovery (AMD; posing interesting conjectures to explore)~\cite{poesia2024learningformalmathematicsintrinsic, mishra2023mathematicalconjecturegenerationusing,bailey2026scaling}. Answering the question \emph{``is this problem worth solving?''} often involves estimating properties of the problem before any experience e.g., whether the problem is likely to be challenging, or generally interesting. While some prior work has explored the notion of interestingness in humans~\cite{10.1093/philmat/nku014, day1967evaluations, berlyne1963complexity}, in this paper, we take \textbf{the first step} in empirically comparing interestingness and difficulty judgments across humans and LLMs.
\begin{figure*}[!t]
    \centering
    \includegraphics[width=\linewidth]{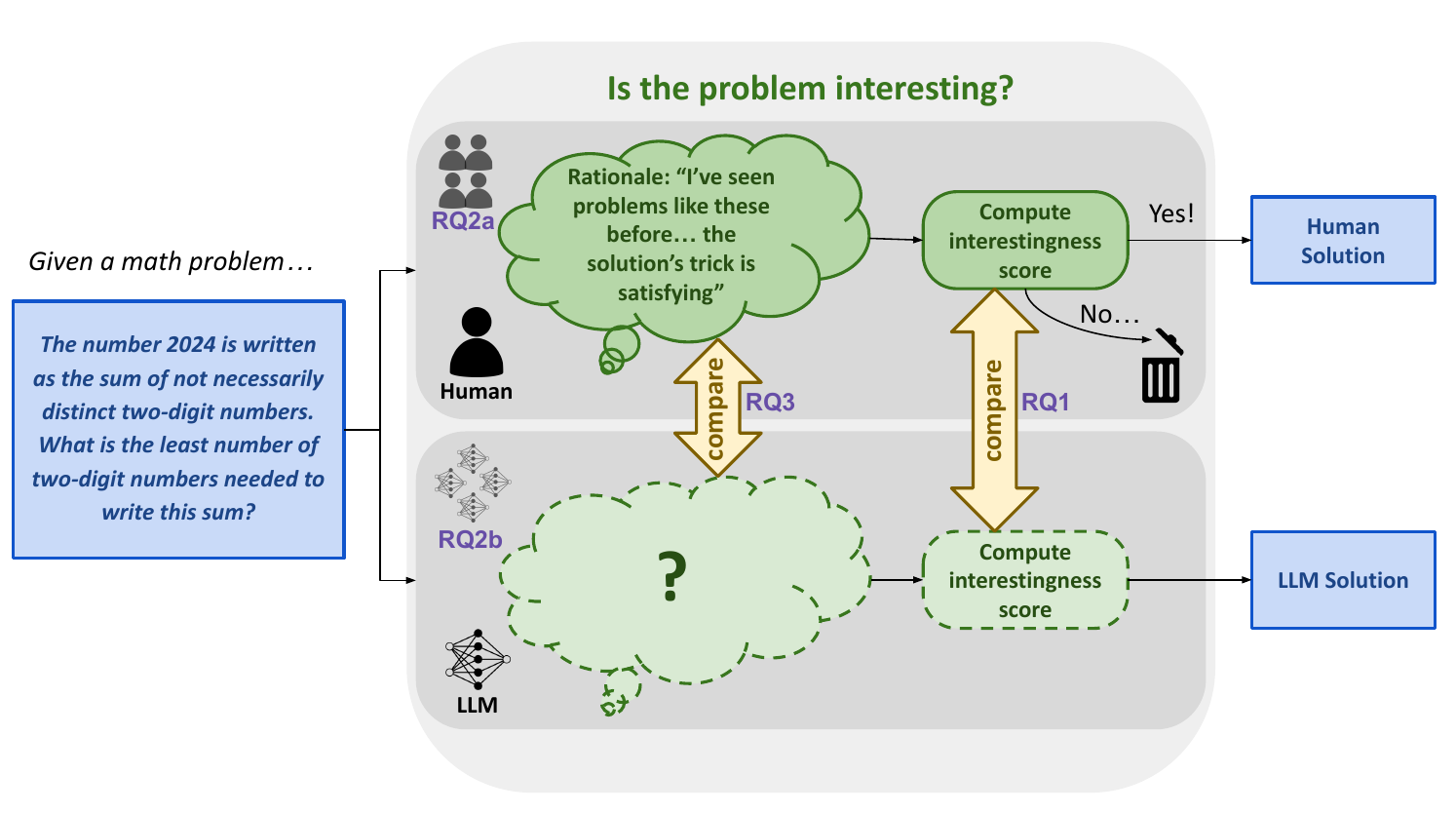}
    \caption{Our experimental pipeline for studying human-LLM alignment on interestingness is depicted above. For any problem, human participants rate \emph{how} interesting it is and \emph{why} they find it interesting (top panel). We also evaluate LLMs' ratings for these (bottom panel), comparing human-LLM alignment on how interesting a problem is (RQ1) and their rationales for interestingness (RQ3). We explore whether the \emph{distributions} of judgments between humans and LLMs align (RQ2a) and if model pooling can improve this alignment (RQ2b). While not depicted in the figure, we also study how reasoning models reason about problem interestingness (RQ4) and explore whether LLMs are capable of generating mathematical problems that interest humans (RQ5).\looseness-1}
    \label{fig:interestingness}
    \vspace{-0.8cm}
\end{figure*}

Classical work in AMD modeled interestingness using hand-coded heuristics to guide conjecture generation and theorem proving---often hundreds of them, limiting scalability~\cite{Colton1999AutomaticCF, COLTON2000351, Lenat1976AMAA, Colton1999AutomaticCF, epstein1987} . Recent work has introduced complementary approaches that structure mathematical statements to yield conjectures provable through domain expertise, but lack heuristics to guide search based on human notions of interestingness~\cite{mishra2023mathematicalconjecturegenerationusing}. Other work has also explored evolutionary search to find symbolic interestingness evaluators that lead to human-written statements being rediscovered~\cite{tsoukalas2025learninginterestingnessautomatedmathematical}. The ability for LLMs and LRMs to operate more freely over language text (e.g., informal natural language problem descriptions) opens up new possibilities for flexible modeling of problem interestingness. This makes a new approach possible---instead of manually specifying heuristics, we can directly measure how well LLMs capture human perceptions of interestingness, and use this to identify and address gaps.\looseness-1

Our work takes a step in this direction, conducting initial studies comparing LLMs and LRMs to human judgments of mathematical problem interestingness and understanding whether these models can actually \emph{generate} interesting problems. We run two controlled studies, one with crowdsourced participants and one with IMO participants rating math problems from the American Math Competition and International Math Olympiad, respectively. As we see in Figure~\ref{fig:interestingness}, participants are asked to rate how interesting they think a problem is and also why they find it interesting. We then evaluate LLMs on the same task of problem evaluation and compare the human and LLM judgments for all problems. Specifically, we make the following contributions:
\begin{itemize}[topsep=-1pt, leftmargin=20pt, itemsep=-2pt]
    \item A systematic evaluation of whether a battery of LLMs judge problem interestingness in human-aligned ways.
    \item Insights into how LLMs evaluate the interestingness and difficulty of math problems, and how they do (and don't) align with those of humans. Importantly, we make a recommendation for the use of multiple LLMs in discovery and human-AI collaboration systems and highlight a proxy for interestingness in LRMs.
    \item An initial exploration of LLMs' ability to generate problems that humans find interesting.
\end{itemize}

\section{Related work}
\label{sec:related_work}
Our work lies at the intersection of (i) AI for mathematical problem solving, (ii) evaluating AI evaluations, and (iii) human judgments of interestingness.

\textbf{AI for mathematical problem solving.}
The use of large language models for math problem solving began with grade- and high-school level mathematics with benchmarks like GSM8K and MATH~\cite{cobbe2021trainingverifierssolvemath, hendrycks2021measuringmathematicalproblemsolving}. LLMs quickly saturated these benchmarks due to increased problem-solving abilities~\cite{bubeck2023sparksartificialgeneralintelligence}. To address benchmark saturation, more challenging benchmarks like Putnam-AXIOM, PutnamBench, FrontierMath, and the AIMO prizes containing PhD and competition mathematics emerged~\cite{gulati2025putnamaxiomfunctionalstaticbenchmark, tsoukalas2024putnambenchevaluatingneuraltheoremprovers, glazer2025frontiermathbenchmarkevaluatingadvanced, ai-mathematical-olympiad-prize, ai-mathematical-olympiad-progress-prize-2}. Large Reasoning Models (LRMs), trained with Reinforcement Learning techniques to utilize additional compute at inference-time, quickly rose to this challenge. Closed LRMs from providers like DeepMind~\cite{DeepMind2025GeminiGold}, Harmonic~\cite{achim2025aristotleimolevelautomatedtheorem}, and OpenAI~\cite{openai2025o3o4mini} achieved Gold medal-level performance at the 2025 International Mathematics Olympiad, highlighting the gap between proprietary and open models. However, these LLMs were only evaluated on their ability to \textit{solve} mathematical problems. These accomplishments complement a parallel line of computer-assisted discovery systems (e.g., FunSearch~\cite{veličković2024amplifyinghumanperformancecombinatorial}, AlphaTensor~\cite{AlphaTensor2022}, and AlphaEvolve~\cite{novikov2025alphaevolvecodingagentscientific}), which demonstrated that search, learning, and symbolic tools can together yield novel or improved results. Despite the achievements, a gap remains: humans pose the problems that models solve and make discoveries for, and thus models are not evaluated at judging problems or generating interesting ones~\citep[cf.][]{zhang2023ai}. 

\textbf{Evaluating AI Evaluations.}
LLMs have mostly interacted with human-posed problems so far; there is comparatively less understanding of how well LLMs \textit{evaluate} whether problems are worth solving. This research direction goes beyond evaluating solution correctness or proof validity, and asks whether models can evaluate \emph{tasks themselves} (e.g., rating difficulty, quality, or broader subjective properties) and whether these evaluations align with people. For example, recent work has examined whether LLMs' evaluations of board games align with those of humans~\cite{collins2025evaluating}. This is in contrast to the ``LLM-as-Judge'' approach, which usually uses LLMs to evaluate \textit{solutions}, for example, to evaluate the quality of writing, code, and LLM-generated answers to math problems~\cite{zheng2023judgingllmasajudgemtbenchchatbot, jiang2025codejudgebenchbenchmarkingllmasajudgecoding, stephan2025calculationadjudicationexaminingllm, gu2025surveyllmasajudge}. 
Recent work has used LLM judgments to filter synthetic mathematical problems and conjectures at scale \cite{xin2024deepseek,bailey2026scaling}: here, we systematically study how such judgments compare with those of humans.

\textbf{Human interestingness judgments.} Previous work in cognitive science has attempted to build models of what humans find interesting, and ultimately, what drives human curiosity~\cite{Loewenstein1994Curiosity, hy_day_curiosity}. Some experiments studied students' reactions to irregular patterns~\cite{berlyne1963complexity} and complex shapes~\cite{day1967evaluations}, linking how they judged interestingness to complexity, novelty, and incongruity, and how these factors shape exploratory behavior. Studies done within mathematics specifically explore expert appraisals of what makes a proof ``beautiful''~\cite{10.1093/philmat/nku014}. Attempts to build artificially intelligent mathematical systems, including AMD systems such as Graffiti~\cite{Colton1999AutomaticCF}, AM~\cite{COLTON2000351, Lenat1976AMAA}, HR~\cite{Colton1999AutomaticCF, COLTON2000351}, and GT~\cite{epstein1987}, operationalized ``interestingness'' via large sets of hand-coded heuristics, which enabled early progress but limited scalability and adaptability across domains. While research in cognitive science has tried to explain how humans judge interestingness, work involving AMD systems often manually imbues these systems with problem features humans deem interesting. LLMs, which can be probed using natural language, reveal a new possibility. Drawing on the spaces of cognitive science and AMD, in this work, we evaluate whether LLMs are capable of evaluating interestingness in math problems and outline suggestions for how AI systems can better be aligned with how humans judge mathematical problems. While not the primary focus in this work, our data collection and analyses also raise new questions about how people decide what problems are even worth solving~\citep{getzels1982problem, nickles1981problem, chu2025makes, collins2025people, wong2025meta}.\looseness-1 

\vspace*{-2mm}
\section{Methods}
\label{sec:methods}
\vspace*{-2mm}

We design and run a suite of human and model experiments to assess interestingness judgments (Figure~\ref{fig:interestingness}). Specifically, we study six research questions to characterize LLMs’ ability to evaluate and generate interesting math problems.
\begin{enumerate}[topsep=-1pt, leftmargin=10pt, itemsep=-2pt]
    \item[] \textbf{RQ1:}~How do LLMs' judgments of interestingness align with those of humans? 
    \item[] \textbf{RQ2a:}~Are the \emph{distributions} of interestingness judgments aligned between humans and LLMs?
    \item[]\textbf{RQ2b:}~If not aligned, how can we improve alignment of LLMs judgments of interestingness with humans?
    \item[] \textbf{RQ3:}~Do humans and LLMs align on \emph{why} they find problems interesting?
    \item[] \textbf{RQ4:}~How much do LRMs reason when they judge problem interestingness?
    \item[]\textbf{RQ5:}~Are LLMs capable of generating math problems humans find interesting? 
\end{enumerate}
%
To take steps to address \textbf{RQ1-RQ4}, we study human interestingness judgments in two participant pools and two different banks of problems: (1)~crowdsourced participants reasoning about AMC problems, and (2)~IMO participants (engaged in-person at the 2024 competition) reasoning about past IMO problems. To address \textbf{RQ5}, we take initial steps to test whether LLMs can generate interesting problems by crowdsourcing human ratings for LLM-generated math problems. All studies received ethics approval by our institutional ethics review boards. 

\textbf{Prolific dataset: crowdsourcing human interestingness judgments.} 
We recruit 63 participants from Prolific, a crowdsourcing platform common in cognitive science~\citep{palan2018prolific}. Our sample size aligns with standard practice for related human participant empirical studies~\citep{10.1145/2858036.2858498}. Each participant was assigned to one of two conditions, rating~10 problems each. Each participant saw the same control problems and either a problem's original version or its variant (see below). Participants were required to think about each problem for at least one minute before rating its \emph{interestingness} and \emph{difficulty} on a scale of 0-100, and providing a \mbox{one- to three-sentence} rationale for each rating.

\begin{figure*}[!t]
    \centering
    \includegraphics[width=0.49\linewidth]{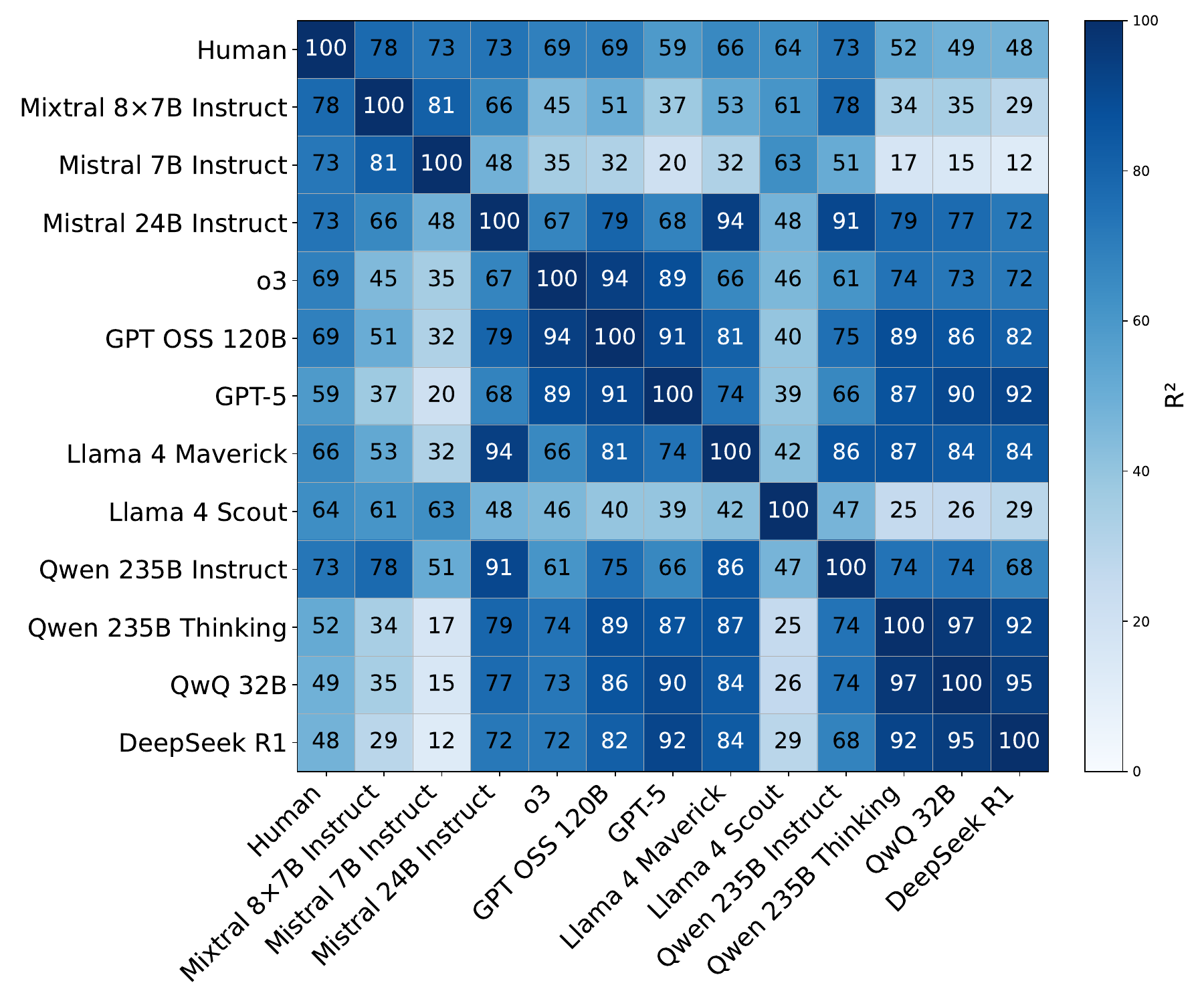}
    \includegraphics[width=0.49\linewidth]{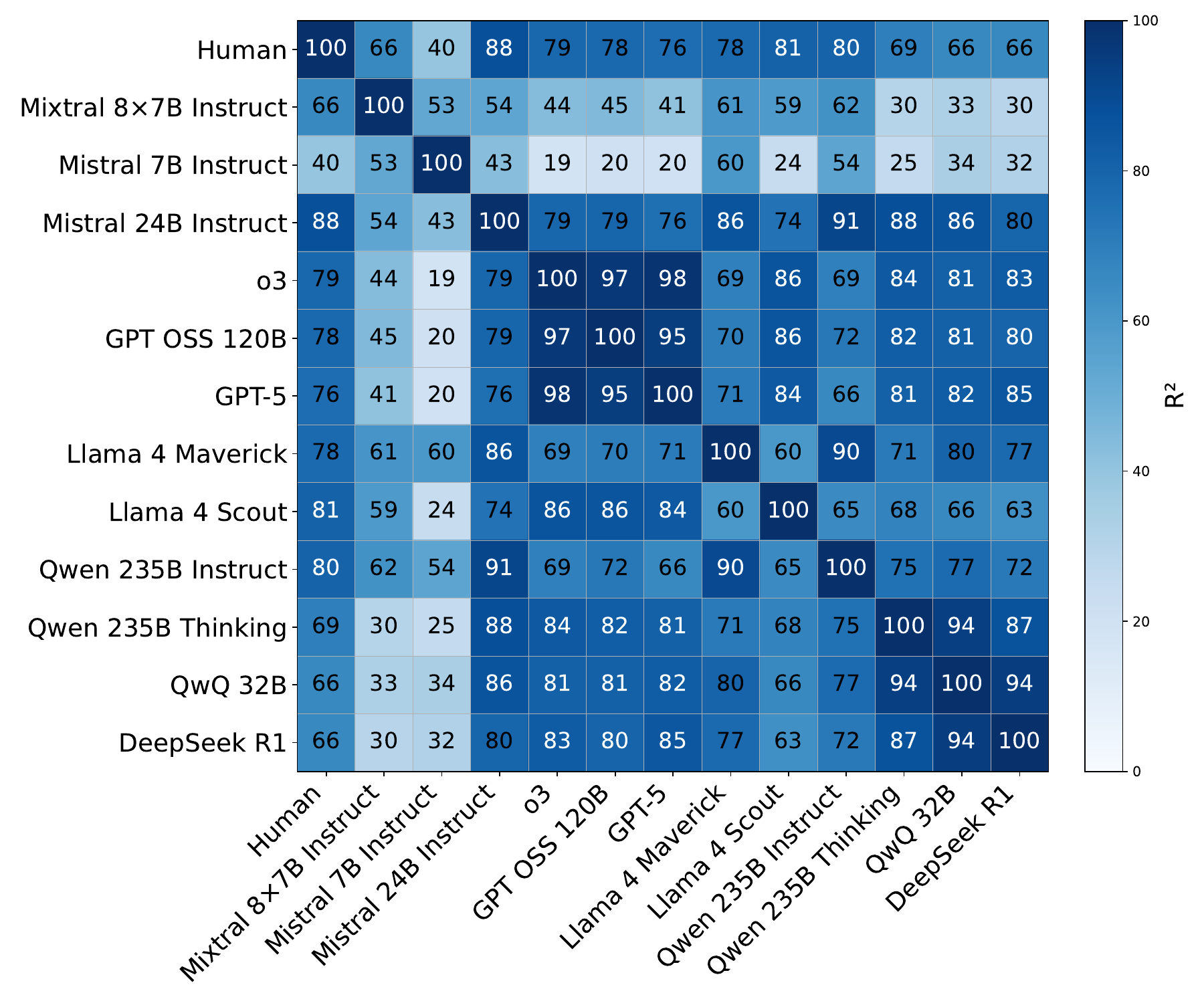}
    \caption{
      Agreement between humans' and LLMs' judgments about problem interestingness (left) and difficulty (right) on the Prolific dataset. Each cell shows the squared Pearson correlation ($R^2$ scaled by 100), between the row and column reasoners' per-problem mean ratings. The left matrix includes models from the same family together and is sorted in descending order by agreement, while the right matrix follows the order of the left matrix to enable easier comparison. Darker cells indicate higher agreement ($R^2$). Models were sampled at temperature $1.0$; additional analyses, e.g., for temperature $0.3$, are included in Appendix~\ref{app:model_correlation_prolific}. We see that different LLMs align with human judgments of interestingness and difficulty to different extents. While models from the Mistral family better approximate human judgments of interestingness, the other families excel at aligning with human judgments of difficulty. 
    }
    \label{fig:prolific-heatmaps-t1.0}
    \vspace{-1em}
\end{figure*}

We curate problems from AMC 8 and AMC 12, high school-level contests given by the Mathematical Association of America~\cite{maa-amc}. Our goal is not to construct a large-scale benchmark, but to create a controlled stimulus set for the human-model study. We therefore deliberately select a set of contest problems and design hand-written variants of base problems to systematically probe dimensions of a problem to increase problem set diversity. For each contest problem, we hand-write a new problem based on a variant type. Examples of variants include increasing/decreasing the sizes of the numbers in the problem, adding/removing steps, etc. We also create two control problems: a negative and a positive control, which are later used to filter out any unfaithful participant ratings. That is, we filter out participants that rate the negative control math problem (``What is 28 + 13?'') as having interestingness $>$ 90. The result is our final dataset for the Prolific study, which contains 18 problems (2 controls and 8 problems with one variant each). A list of all our problems and variants are provided in Appendix~\ref{app:prolific}.

\textbf{IMO data collection.} We conducted a survey of interestingness judgments made by 48 participants at the 2024 IMO. As part of the pre-survey, participants answered questions about \emph{factors of interestingness}, where participants marked how important they found dimensions of problems (e.g. 
"The problem statement is simple and elegant") to making a problem interesting. Our analysis for RQ3 utilizes this portion of the pre-survey, with additional details about the survey provided in Appendix~\ref{app:imo}.

\textbf{Comparing LLM and Human Judgments.} We evaluate 12 language models from five families. We sample 20 responses for each model at temperatures 0.3 and 1.0 for most models where temperature sampling is allowed. Due to computational costs, for reasoning models on our IMO dataset, we sampled 10 responses each. Additionally, note that GPT-5 and o3 do not allow temperature sampling and were thus only sampled at their default temperature (1.0). 

\textbf{Understanding whether LLMs can generate interesting problems.}
Finally, we conduct an initial exploration into whether LLMs can generate competition math problems that humans find interesting. We first generate 90 math problems using three models, Mistral 7B Instruct, Qwen 235B Thinking, and OpenAI o3. The LLMs are prompted specifically to generate problems that are interesting high school-level competition math problems. We then filter out invalid problems manually. Problem validity is manually determined (e.g., problems with inconsistent information, no correct answer, etc.\ are discarded). The set of 90 problems is split into 5 batches. 30 total participants evaluate problems from two of these batches, resulting in 360 judgments we evaluate.
\vspace{-1em}
\section{Results}
\vspace{-1em}
\label{sec:results}
\textbf{RQ1: How do LLMs' judgments of interestingness align with those of humans?} For each LLM, we compute the $R^2$ between per-problem mean interestingness in humans and the model (see Figure~\ref{fig:prolific-heatmaps-t1.0}). 
The human row/column reveal model-human agreement, while other row/column combinations report correlations amongst the models. On the Prolific dataset, model-human $R^2$ ranges from about 0.48 to 0.78, with the strongest agreements from the Mistral family. Split-half human $R^2$, which is our noise ceiling on explainable variance, is $0.72$ with 95\% CI: [$0.53$, $0.87$] (see Appendix~\ref{app:add_results}).  This indicates that current LLMs (especially those in the Mistral family) are able to approximate human perceptions of interestingness with surprising fidelity. For ratings of difficulty, we see the opposite: most models within the Mistral family struggle to align with human judgments of problem difficulty, with other families aligning more closely with humans. We also see that the Deepseek and Qwen reasoning models struggle in aligning with both the human judgments of interestingness and difficulty.\looseness-1


Beyond human-LLM agreement on interestingness and difficulty, respectively, we next test whether any one agent (human or model) is correlated with respect to its own ratings of interestingness and difficulty. That is, is an agent's judgment of difficulty related to its judgment of interestingness? In humans, interestingness and difficulty are not highly correlated. We define the per-participant interestingness/difficulty correlation as the $R^2$ between a participant’s interestingness and difficulty scores over the problem set, and find a mean correlation of 0.47 (95\% CI: [0.39, 0.55], standard deviation 0.32). In reasoning models, the correlation is consistently at or above 0.9. Other LLMs range from 0.41 to 0.89, with only Llama Scout and Mistral 7B Instruct below 0.80 (Appendix~\ref{app:add_results_prolific}). Prior work has used proxies like difficulty~\cite{poesia2024learningformalmathematicsintrinsic} and usefulness~\cite{KasrielEtAl2025UsefulnessDriven} for interestingness; given the low human correlation between interestingness and difficulty on the same problems, we encourage future work to consider treating interestingness as its own dimension in discovery systems and understanding the factors that drive peoples' judgments, as we begin to explore later in RQ3.
\looseness-1
\begin{tcolorbox}[takeawaybox]
\textbf{Takeaway.} LLMs broadly align with peoples' ratings of interestingness and difficulty in math problems. However, people show moderate correlation between their \emph{own} ratings of interestingness and difficulty for any given problem. LLM ratings for interestingness and difficulty for a problem, on the other hand, are typically highly correlated.
\end{tcolorbox}

\textbf{RQ2a: Are the \emph{distributions} of interestingness judgments aligned between humans and LLMs?} While LLMs capture some aspects of human interestingness judgments, in many instances, their \emph{distributions} of interestingness ratings diverge. We measure distributional similarity between human and LLM judgments using the  Wasserstein-1 Distance (WD)~\cite{villani2009optimal}, based on the L1 norm. For each problem, we calculate the WD between the human distribution of judgments and the distribution of one model's judgments. We compute the bootstrapped average WD score over our dataset of problems for each model. We report these values in Table~\ref{tab:wd_interestingness}.

\begin{table*}[!t]
\centering
\caption{Distributional alignment in interestingness judgments, as measured by Wasserstein distances (WD) between human and LLM distributions of interestingness judgments on the Prolific dataset. We use the Wasserstein-1 distance with the L1 norm, treating each individual rating with equal weighting. Lower values indicate closer alignment to human distributions. The human-human split-half baseline indicates the amount of explainable variability in the human data. LLMs are sampled at a temperature of 1.0. WD measures for other temperatures and for difficulty judgments are reported in Appendix~\ref{app:wd_tables}. We find that only one model, Mistral 7B Instruct, comes close to reflecting the distribution of human judgments observed in our study.}
\begin{tabular}{lcllcl}
\toprule
\textbf{Model} & \textbf{WD} &\textbf{ 95\% CI} & \textbf{Model} & \textbf{WD} & \textbf{95\% CI} \\
\midrule
Human-Human & 9.5 & [7.8, 11.5] & Mixtral 8×7B Instruct & 20.2 & [17.5, 23.1] \\
Mistral 7B Instruct & 12.4 & [9.3, 16.0] & Llama 4 Maverick & 20.7 & [18.1, 23.7] \\
Mistral 24B Instruct & 15.6 & [13.5, 18.0] & Llama 4 Scout & 21.1 & [18.6, 23.6] \\
DeepSeek R1 & 16.4 & [13.1, 19.7] & GPT-5 & 21.2 & [18.3, 24.2] \\
QwQ 32B & 18.1 & [15.3, 21.0] & Qwen 235B Instruct & 21.3 & [18.5, 24.1] \\
GPT OSS 120B & 18.3 & [15.9, 21.0] & o3 & 21.9 & [18.7, 25.5] \\
Qwen 235B Thinking & 19.8 & [16.4, 23.9] &  &  &  \\
\bottomrule
\end{tabular}
\label{tab:wd_interestingness}
\end{table*}

For a human comparison baseline, we calculate the bootstrapped split-halves WD by repeatedly halving the human judgment set, converting each set into a distribution, and and calculating the WD between them. The lowest WD is achieved by Mistral 7B (mean WD = 12.4, 95\% CI = [0.3, 16.0]), whose 95\% CI is the only one overlapping with the human split-half baseline (which has mean WD = 9.5, 95\% CI = [7.8, 11.5]; Table~\ref{tab:wd_interestingness}).\looseness-1

\begin{tcolorbox}[takeawaybox]
\textbf{Takeaway.} LLM's interestingness judgments generally do not well-capture the diversity across the population of human judgments. This highlights the importance of carefully considering which models we integrate in AMD and human-AI collaborative systems. 
\end{tcolorbox}

\textbf{RQ2b: How can we improve the alignment of LLMs' interestingness judgments to that of humans?}
We next explore whether multiple models, when pooled, can better capture the distribution of human judgments than any one model alone. We construct the full space of combinatorial subsets $S$ of the 12 models we experiment with ($|S| \geq 2$). We calculate the WD for the human distribution of judgments against the distribution of judgments derived from each pooled set. We compute 95\% CIs via 1000 bootstrapping iterations. In each iteration, we resample problems with replacement and draw a fresh subsample of 30 model ratings from the pooled distribution (i.e., if there are M models in a pool, where each model produced N ratings for any one problem, we repeatedly sample 30 judgments with replacement from the set of M $\times$ N judgments). Thus, this reflects both problem-level sampling variability and uncertainty from the rating subsampling. We present the top- and bottom-3 distributionally aligned subsets in \Cref{tab:best_pooled_sets_vs_hh}. The best-performing minimal subset is \{Mistral-24B, Mistral-7B, Mixtral-8x7B, OpenAI o3\}. Adding more models beyond this point tends to worsen alignment, likely because intra-family similarities reduce rather than increase judgment diversity \cite{jiang2025artificialhivemindopenendedhomogeneity}. We also conduct 5-fold cross-validation over problems, selecting the best subset of $k$ models on the training folds and evaluate WD on the held-out fold. While the set of models that achieves the lowest mean held-out WD changes, the WD estimate stays approximately the same (WD = 9.13 [7.65, 10.61]).

\begin{tcolorbox}[takeawaybox]
\textbf{Takeaway.} Pooled sets of LLMs can generally better reflect the human distribution of judgments compared to individual LLMs. This discrepancy highlights the importance for future work to explore AI-mathematics systems that rely on multiple LLMs, since most such work currently only relies on one LLM~\cite{poesia2024learningformalmathematicsintrinsic, dong2025stpselfplayllmtheorem, doi:10.1073/pnas.2318124121}.
\end{tcolorbox}

\begin{table*}[!t]
\centering
\caption{Wasserstein Distance (WD) between the human and pooled model judgments, averaged over our dataset (lower is better). The human-human WD represents a split-half baseline calculated by repeatedly halving the human judgments, calculating the WD between the resulting distributions, and bootstrapping. The top 3 and bottom 3 pooled LLM combinations ranked by pooled per-problem Wasserstein distance to human rating distributions are reported here. We see that pooling models considerably improves their alignment with the human distribution of judgments.}
\begin{tabular}{l c l}
\toprule
 Pooled model set & WD $\downarrow$ & 95\% CI \\
\midrule
\multicolumn{3}{l}{\textbf{Most aligned pooled sets (lowest WD)}} \\
\{Mistral-24B, Mistral-7B, Mixtral-8x7B, o3\} & 9.07 & [8.34, 10.9] \\
\{Mistral-7B, o3, Qwen Instruct\} & 9.27 & [8.30, 10.8] \\
\{Mistral-7B, Mixtral-8x7B, o3, Qwen Instruct \} & 9.28 & [8.26, 11.1] \\
\midrule
\multicolumn{3}{l}{\textbf{Human baseline}} \\
Human--Human (split-half baseline) & 9.5 & [7.8, 11.5] \\
\midrule
\multicolumn{3}{l}{\textbf{Least aligned pooled sets (highest WD)}}\\
\{GPT-OSS-120B, o3\} & 19.3 & [16.6, 22.3] \\
\{GPT-5, GPT OSS 120B, o3\} & 19.6 & [16.5, 22.5] \\
\{GPT-5, o3\} & 20.6 & [17.8, 24.0] \\
\bottomrule
\end{tabular}

\vspace{-3pt}
\label{tab:best_pooled_sets_vs_hh}
\end{table*}

\textbf{RQ3: Do humans and LLMs align on \emph{why} they find problems interesting?} 
To assess what factors of problems models and people find interesting, we next compared the distribution of reasons that humans and LLMs selected for problem interestingness. We focus on pre-survey questions which participants answered assigning importance to different interestingness reasons (e.g., ``Please indicate how important this factor generally is for a problem to be interesting to you: The problem statement is simple and elegant.''). The answers were collected on a four point scale of ``not important'' to ``very important'' with an option to mark if the criterion did not make sense to the participant. We replicate this experiment with all LLMs we examine, sampling 50 responses (to match the 48 participants from our IMO study). For each interestingness criterion, human participants' answers spanned the allowed range of importance options from not important to very important. However, despite sampling each LLM 50 times at temperature 1.0, most LLMs only selected one to two importance scores for each reason. We include comparisons between the human and model distributions for all models we examine in Appendix~\ref{app:add_results_imo}. Only Mistral 7B Instruct (Appendix Figure~\ref{fig:mistral_7b_interestingness_importance}) and Mistral 24B Instruct (Appendix Figure~\ref{fig:mistral_24b_interestingness_importance}) reflect the human distributions of interestingness rationales well. Future work can better understand the drivers for such differences across model families. 
\begin{tcolorbox}[takeawaybox]
\textbf{Takeaway.} People and LLMs are generally misaligned on \emph{why} they find problems interesting. This encourages future work towards a better understanding of the drivers behind this divergence.
\end{tcolorbox}

\begin{figure*}[!t]
    \centering
    \includegraphics[width=0.99\linewidth]{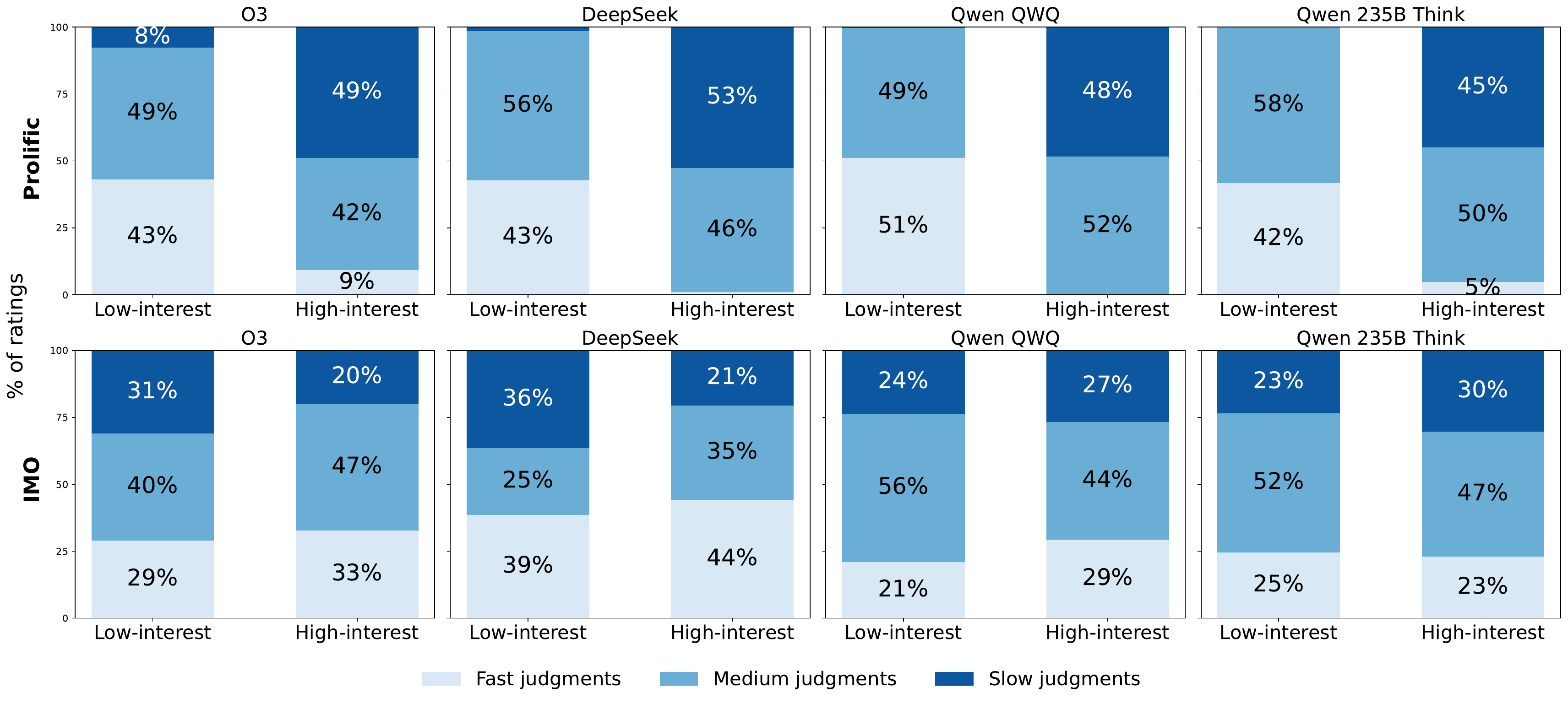}
    \caption{Judgment speed distributions across LRMs on low- vs.\ high-interest Prolific problems. A low-interest problem for a model is one that is given a below median interestingness score by the model. High-interest problems are those which are given higher than the median interestingness score. For each subset of problems, we label whether a judgment was slow, medium, or fast, based on the distribution of reasoning token counts for that model. Slow judgments occupy the bottom quartile and fast judgments occupy the top one, with medium-speed judgments covering the middle. We see that LRMs tend to engage in longer reasoning chains for problems that they ultimately label as being higher interest. We include the judgment speed distributions across LRMs for difficulty judgments in Appendix~\ref{app:add_results_prolific}.}
    \label{fig:flash_judgments}
    \vspace{-1.5em}
\end{figure*}

\textbf{RQ4: How much do LRMs reason when they judge problem interestingness?}

We examine the resource usage of LRMs when reasoning about a problem via the number of reasoning tokens used (i.e., the length of their reasoning chain). We use this to explore the distinction in reasoning time to assess problems that LRMs judge as low- vs.\ high-interest (which is judged by whether a problem's interestingness score is below or above the median of interestingness scores from judgments from that LRM). In Figure~\ref{fig:flash_judgments}, we see that for the Prolific dataset, all four LRMs make fast, ``flash'' judgments of uninterestingness while investing longer reasoning chains for problems they judged as interesting. This distinction disappears at the IMO level, where judgments made under high reasoning time are no longer correlated with higher interestingness ratings. One possible explanation is that for hard Olympiad-style problems, resource usage spent on parsing and understanding the problem dominates the total resource usage spent thinking about the problem, leading to differences in interest making minimal impact on the length of the total reasoning chain.

\begin{tcolorbox}[takeawaybox]
\textbf{Takeaway.} There is a correlation between how long LRMs think for and how they rate problem interestingness for the Prolific problems in our dataset. However, the relation between thinking time and interestingness breaks down for more challenging IMO-level problems.
\end{tcolorbox}

\textbf{RQ5: Are LLMs capable of generating math problems humans find interesting?}
So far, we explored whether LLMs' interestingness judgments align with those of humans. But ultimately, in human-AI collaborative applications, it matters whether LLMs are also capable of \emph{generating} interesting problems; this could either be to pique students' interest as an AI tutor or to complement the problems a researcher thinks about as an AI co-scientist \cite{lu2024aiscientistfullyautomated, si2025can}. As an initial case study, we begin to explore whether LLMs are capable of generating competition math problems that humans find interesting.

\begin{wrapfigure}{r}{0.48\linewidth}
    \centering
    \includegraphics[width=1\linewidth]{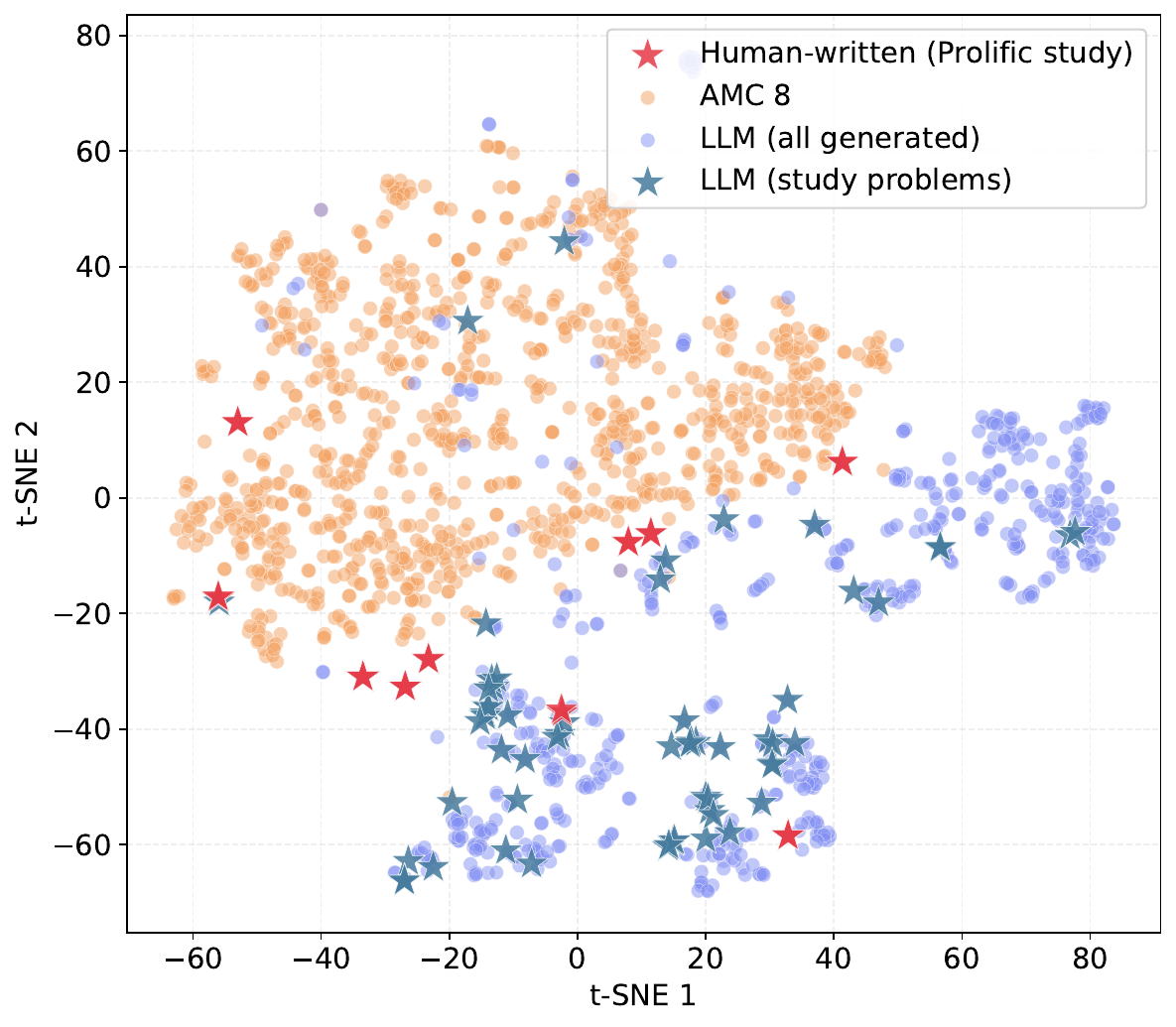}
    \caption{A t-SNE visualization of human- vs. LLM-written math problem embeddings. The LLM-generated problems from our study (purple circles) are plotted against all AMC problems (orange circles). Prolific problems (red stars), taken or adapted from the AMC, and LLM-generated problems we collect human judgments for (blue stars) are also included. We see that human- and LLM-written problems cluster separately in the semantic space, suggesting systematic differences between the two.}
    \label{fig:tsne}
    \vspace{-2em}
\end{wrapfigure}

First, we sample 30 ``interesting, high-school level competition math problems'' from the 12 models in our experiments. To understand how LLM-written problems compare to those written by humans, we embed the $360$ generated problems alongside all AMC problems (which, to the best of our knowledge, are human-authored) and the problems from our Prolific study (which are taken or adapted from the AMC). All problems are embedded with tSNE~\cite{vandermaaten2008visualizing} (see details in Appendix~\cref{app:tsne}).
We see in Figure~\ref{fig:tsne} that human- and LLM-written problems tend to cluster in different regions of the semantic space, highlighting the need for a better understanding of the gaps between the problems humans vs.\ LLMs write.

But do people find these generated problems interesting? We take initial steps towards this question via a pilot problem evaluation study with a new group of Prolific participants. The study follows our previous experimental design (to collect the judgments in RQ1-2), wherein participants now judge LLM-generated problems. Participants are only shown the problems and do not know the problems are LLMs generated to avoid bias. We choose LLMs along the range of human-alignment (as measured by WD in RQ2a): we use Mistral 7B instruct (most human aligned), Qwen 235B Thinking (somewhat human aligned), and o3 (least human aligned) as problem generators. This also allows us to assess whether models that are more aligned to people in their evaluations of problem interestingness are also better generators, or vice versa. From these 90 problems, we manually filter invalid problems (e.g., problems with inconsistent information, no correct answer, etc.). Additional information about the filtering process and ablations addressing automated filtering are provided in Appendix~\ref{app:prob_generation}.

We find substantial variation in human ratings of how interesting the LLM-generated problems are. In Appendix~\cref{tab:top_bottom_generated_problems}, we present the top- and bottom-3 problems and their generator models from both batches we run. We see no significant difference in human ratings for problems across generators. Mistral-generated problems receive a mean rating (with 95\% confidence intervals) of 59.36 [52.35, 68.09], while o3 and Qwen 235B Thinking receive scores of 62.46 [56.76, 69.00] and 62.84 [52.25, 69.83]. We also observe that no individual participant's top 3 problems originate from the same generator model. These results set up future work to explore the role of multiple problem generators, diverging from the current norm of using one base model in AI tutor and AI research systems~\cite{maurya-etal-2025-unifying, lu2024aiscientistfullyautomated}.

Since human-aligned LLM judges are not also necessarily better generators (potentially due to distribution shift in the problems we see in~\cref{fig:tsne}), we next explore whether it is possible to build systems where an LLM \emph{both generates and selects} interesting problems to present to humans. To begin to assess, we compare whether a model's ability to align with human interestingness on the human-written problems also transfers to LLM-written ones. We find no significant relationship in either direction ($r = -0.450, R^2 = 0.203, p = 0.165$) and include the plot in Appendix \Cref{fig:alignment_llm}. Even the LLM that is most aligned with human judgments on LLM-generated problems (GPT-OSS-120B), achieves a Pearson $r$ correlation with human ratings of $0.336$ ($p=0.109$), lagging vastly behind the correlations we see between humans and LLMs for human-written (Prolific) problems. A possible reason for this lack of relationship could be the divergence in the types of problems written by humans vs.\ LLMs, as we see in \Cref{fig:tsne}. This highlights an area where much additional future work is necessary. If the goal is to build AI systems that can independently teach students or assist mathematicians, they will need to be capable of both generating problems, and judging which ones are best to suggest to the user. We encourage future work to explore such generator-selector systems end-to-end.

\begin{tcolorbox}[takeawaybox]
\textbf{Takeaway.} LLMs, after filtering for problem validity, are capable of generating interesting problems. However, models that are aligned evaluators of interestingness on human-written problems are not necessarily good problem generators or evaluators of interestingness on LLM-generated problems, raising new open work. Because no singular model dominated participants' favorite problems, incorporating multiple LLMs in AMD and human-AI collaboration systems should be explored.
\end{tcolorbox}

\section{Conclusion and Limitations} 
In this work, we rigorously assess how judgments of interestingness compare between humans and LLMs, over two new datasets of people's evaluation about math problems. These datasets are collected over a varying difficulty of problems (from AMC- and IMO-level competition math), and from participants with a varying level of expertise (crowdsourced participants and IMO competitors). While LLMs' interestingness judgments generally correlate with the average judgment of people, they do not necessarily match the \textit{distribution} of human judgments. We find that model pooling can better align LLMs with humans and propose future directions in this space. We also find that resource usage in LRMs during problem judgment may be correlated with problem interestingness, but only for less difficult competition problems, highlighting thinking patterns that can serve as proxies for interestingness in the future (and raising additional questions about resource usage in models' reasoning). As an initial exploration, we also evaluate whether humans find LLM-generated math problems engaging. 

However, our study is just a first step. Future work should expand on the number and type of math problems, possibly in collaboration with educators and mathematicians. Both the Prolific and IMO dataset use competition math problems, which are a narrow subset of the problems educators and mathematicians encounter daily. Additionally, the two surveys' populations focus on participants from a crowdsourcing website (who all had a baseline interest in math) and IMO participants, which overlooks beginners and experts, who might have differing perceptions of interestingness in mathematics.

More broadly, our work raises many important questions, in particular, whether models should be aligned to the variability of human responses, which humans should those responses align to, and at what level of mathematical experience? If models are used in human-facing applications, for example, as mathematical AI thought partners~\citep{collins2024building, frieder2024data} designing curricula for students, then we may want to build interestingness measures that meaningfully correlate to human curriculum and the level of the learner~\cite{mishra2025nexttokenmathematicslearningdynamics, tankala-etal-2025-curll}. If instead models work alongside research mathematicians or autonomously discover new mathematics---and decide what mathematics problems to pursue at all---we may set a higher standard for the interestingness judgments~\cite{10.1093/philmat/nku014}. Overall, we hope our work motivates and informs future work on mathematically capable AI systems that engage with subjective notions of mathematical problem interestingness. At the same time, we believe these efforts can also inform a better understanding of what drives humans to find a problem interesting in the first place.\looseness-1

\section*{Acknowledgments}

We thank Simon Frieder, Michael Douglas, Tom Griffiths, Gizem Çaylak, Xaver Davey, Arvid Eriksson, Oscar Eriksson, Lars Hummelgren, and John Wikman for helpful discussions that informed this work. For SM and DB, this work was partially supported by the Wallenberg AI, Autonomous Systems and Software Program (WASP) funded by the Knut and Alice Wallenberg Foundation. KMC acknowledges support from the Cambridge Trust and King's College Cambridge. AW  acknowledges  support  from  a  Turing  AI  Fellowship  under grant  EP/V025279/1, The Alan Turing Institute, and the Leverhulme Trust via CFI. JBT acknowledges support from work AFOSR (FA9550-22-1-0387), the ONR Science of AI program (N00014-23-1-2355), a Schmidt AI2050 Fellowship, and the Siegel Family Quest for Intelligence at MIT.

\bibliography{bib}
\bibliographystyle{plain}

\newpage

\renewcommand{\thesection}{A\arabic{section}}

\startcontents[appendix]
\printcontents[appendix]{l}{1}{\section*{Appendix}}
\clearpage

\newpage

\appendix
\section{Impact Statement}
\label{app:impact}
The goal of this paper is to drive a better understanding of \emph{interestingness} in mathematical problems and whether language models are capable of understanding how humans judge problem interestingness. AI systems are becoming an increasingly larger part of students' and researchers' interactions with mathematics. AI tutor chatbots decide which problems are shown to students, and AI co-scientists often suggest directions for exploration to mathematicians. If people increasingly rely on AI systems to guide problem choice, these systems may shift the landscape of research and education. It is also an open question whether the kinds of selections models endorse for what problem to solve then also influence people's own future evaluations of problems. 

\section{Additional Results}
\label{app:add_results}
\subsection{Prolific Survey Results}

We next include additional details on the Prolific crowdsourced participant survey and related model predictions.

\label{app:add_results_prolific}
\subsubsection{Model Family Correlation}
\label{app:model_correlation_prolific}
\begin{figure}[t]
  \centering
  \begin{subfigure}[t]{\linewidth}
    \centering
    \includegraphics[width=0.65\linewidth]{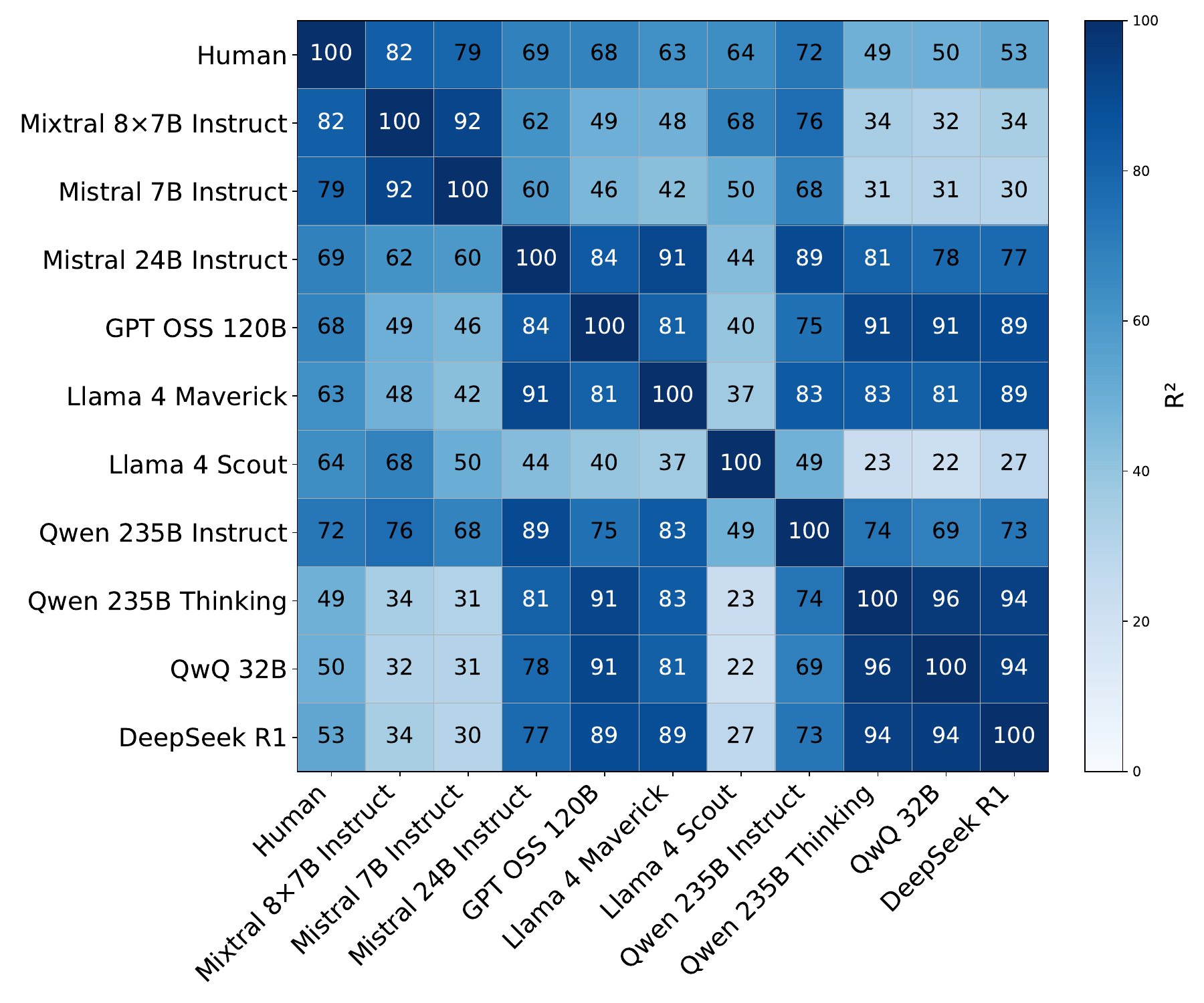}
    \caption{Interestingness (temp = 0.3)}
    \label{fig:heatmap-prolific-int-0.3}
  \end{subfigure}
  \hfill
  \begin{subfigure}[t]{\linewidth}
    \centering
    \includegraphics[width=0.65\linewidth]{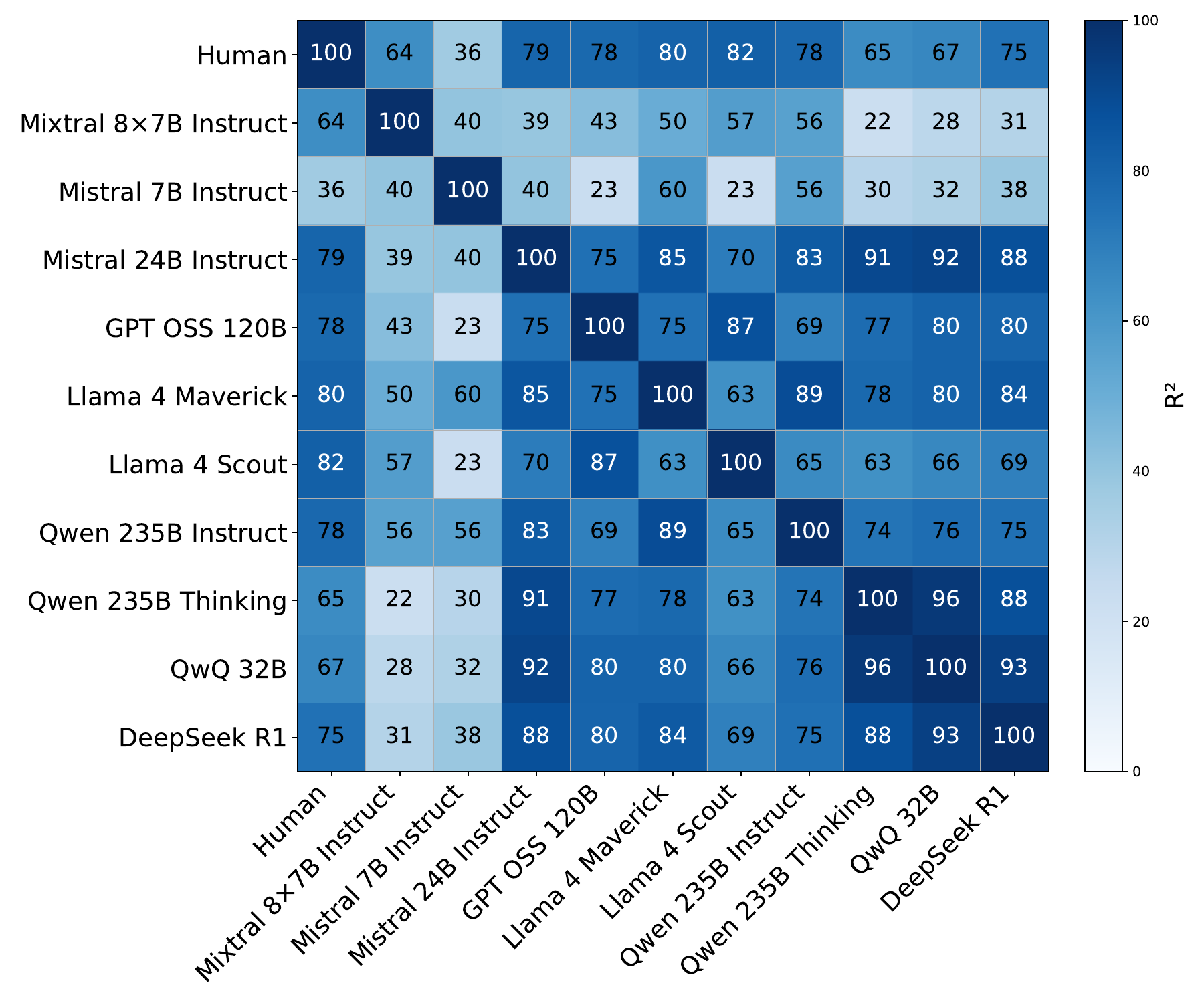}
    \caption{Difficulty (temp = 0.3)}
    \label{fig:heatmap-prolific-chal-0.3}
  \end{subfigure}

  \caption{
  Agent–agent agreement on the \textbf{Prolific} dataset at temperature $0.30$.
  Each cell shows the squared Pearson correlation ($R^2$) between the row and column agents' per-problem mean ratings.
  Darker cells indicate higher agreement; the diagonal is 1.00 by definition.
  The \emph{Human} row/column gives model–human agreement.
  Top: interestingness ratings; bottom: difficulty ratings.
  }
  \label{fig:app_prolific-heatmaps-t0.3}
\end{figure}

\begin{figure}[t]
  \centering
  \begin{subfigure}[t]{\linewidth}
    \centering
    \includegraphics[width=0.65\linewidth]{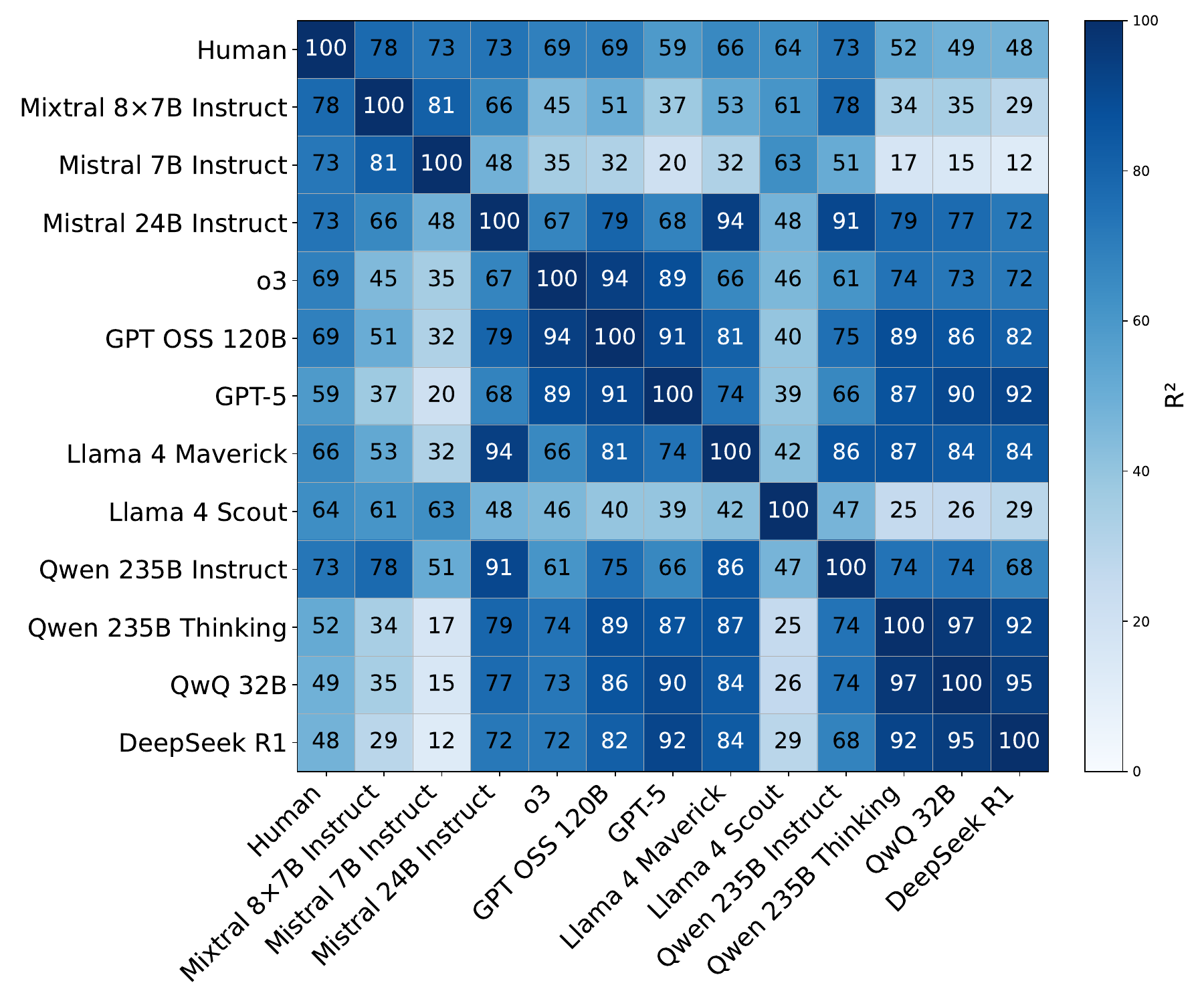}
    \caption{Interestingness (temp = 1.0)}
    \label{fig:heatmap-prolific-int-1.0}
  \end{subfigure}
  \hfill
  \begin{subfigure}[t]{\linewidth}
    \centering
    \includegraphics[width=0.65\linewidth]{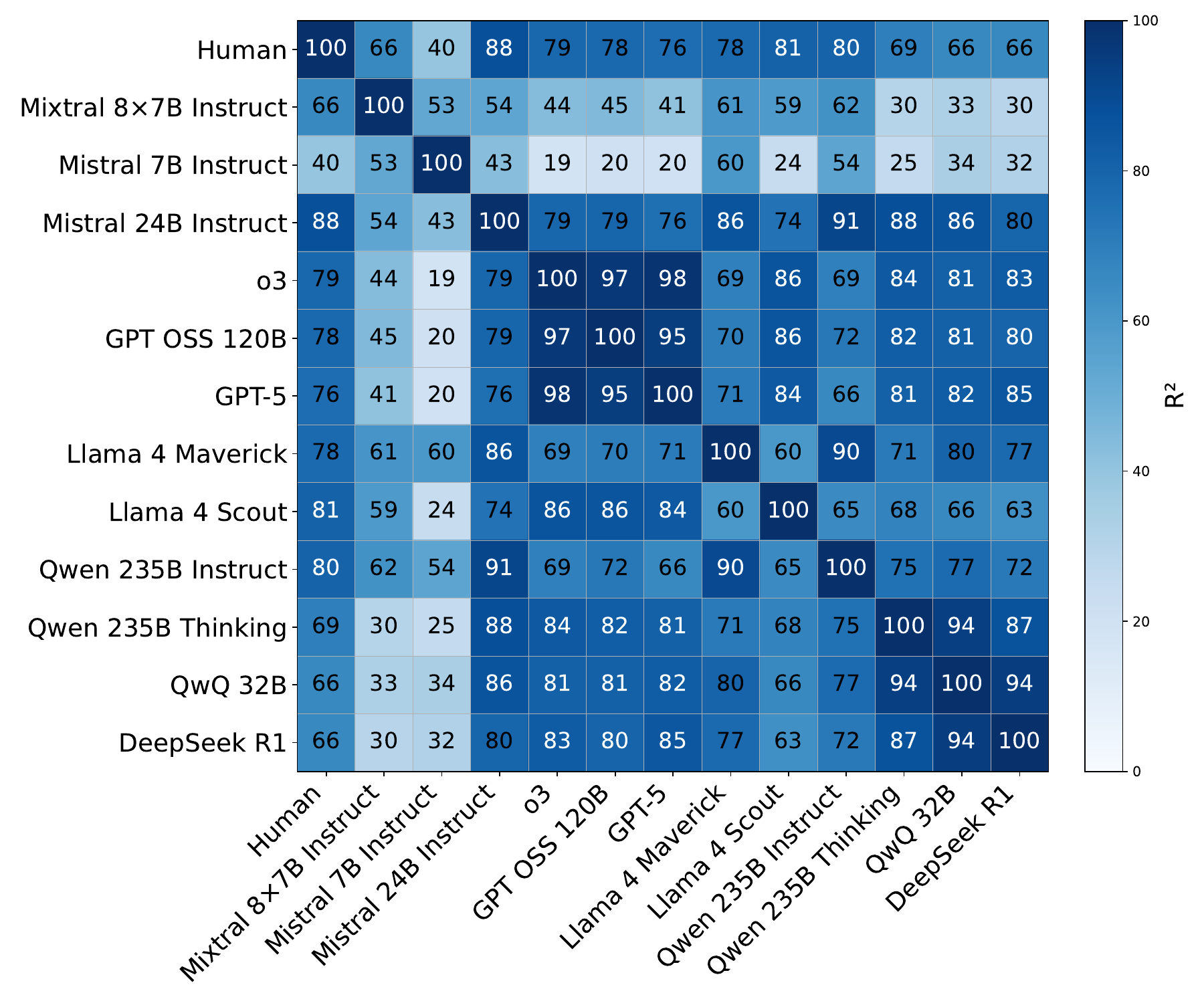}
    \caption{Difficulty (temp = 1.0)}
    \label{fig:heatmap-prolific-chal-1.0}
  \end{subfigure}

  \caption{
  Agent–agent agreement on the \textbf{Prolific} dataset at temperature $1.0$.
  Each cell shows the squared Pearson correlation ($R^2$) between the row and column agents' per-problem mean ratings.
  Darker cells indicate higher agreement.
  The \emph{Human} row/column gives model–human agreement. (Top) interestingness ratings; (bottom)difficulty ratings.
  }
  \label{fig:app_prolific-heatmaps-t1.0}
\end{figure}

\begin{figure}[htbp]
    \centering
    \begin{tabular}{c|c}
        & \textbf{Temp 1.0} \\ \hline
        \rotatebox{90}{Interestingness} &
        \includegraphics[width=0.48\linewidth]{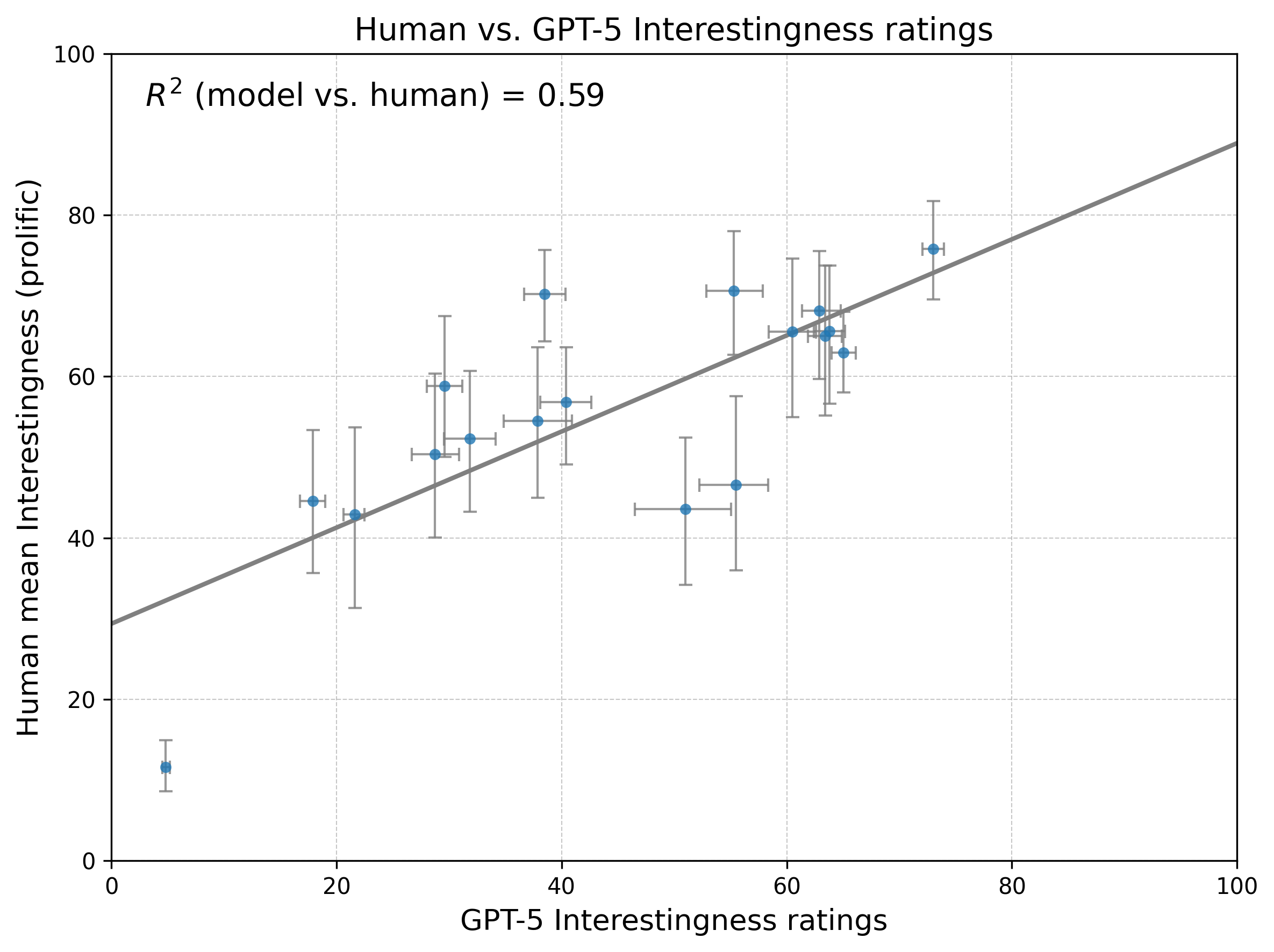} \\
        \rotatebox{90}{Difficulty} &
        \includegraphics[width=0.48\linewidth]{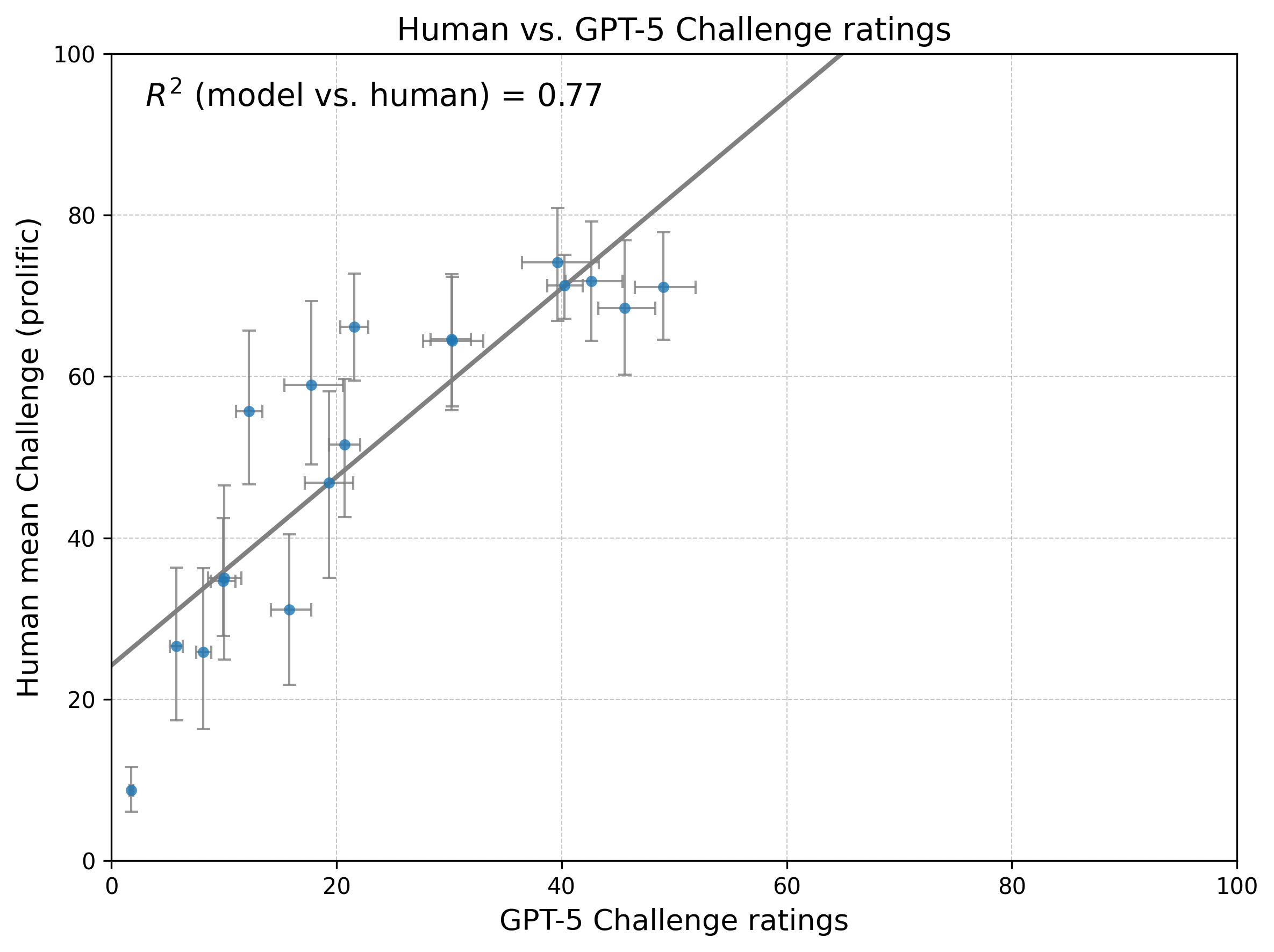} \\
    \end{tabular}
    \caption{GPT-5: Human vs LLM ratings}
    \label{fig:gpt5_r2}
\end{figure}

\begin{figure}[htbp]
    \centering
    \begin{tabular}{c|cc}
        & \textbf{Temp 0.3} & \textbf{Temp 1.0} \\ \hline
        \rotatebox{90}{Interestingness} &
        \includegraphics[width=0.48\linewidth]{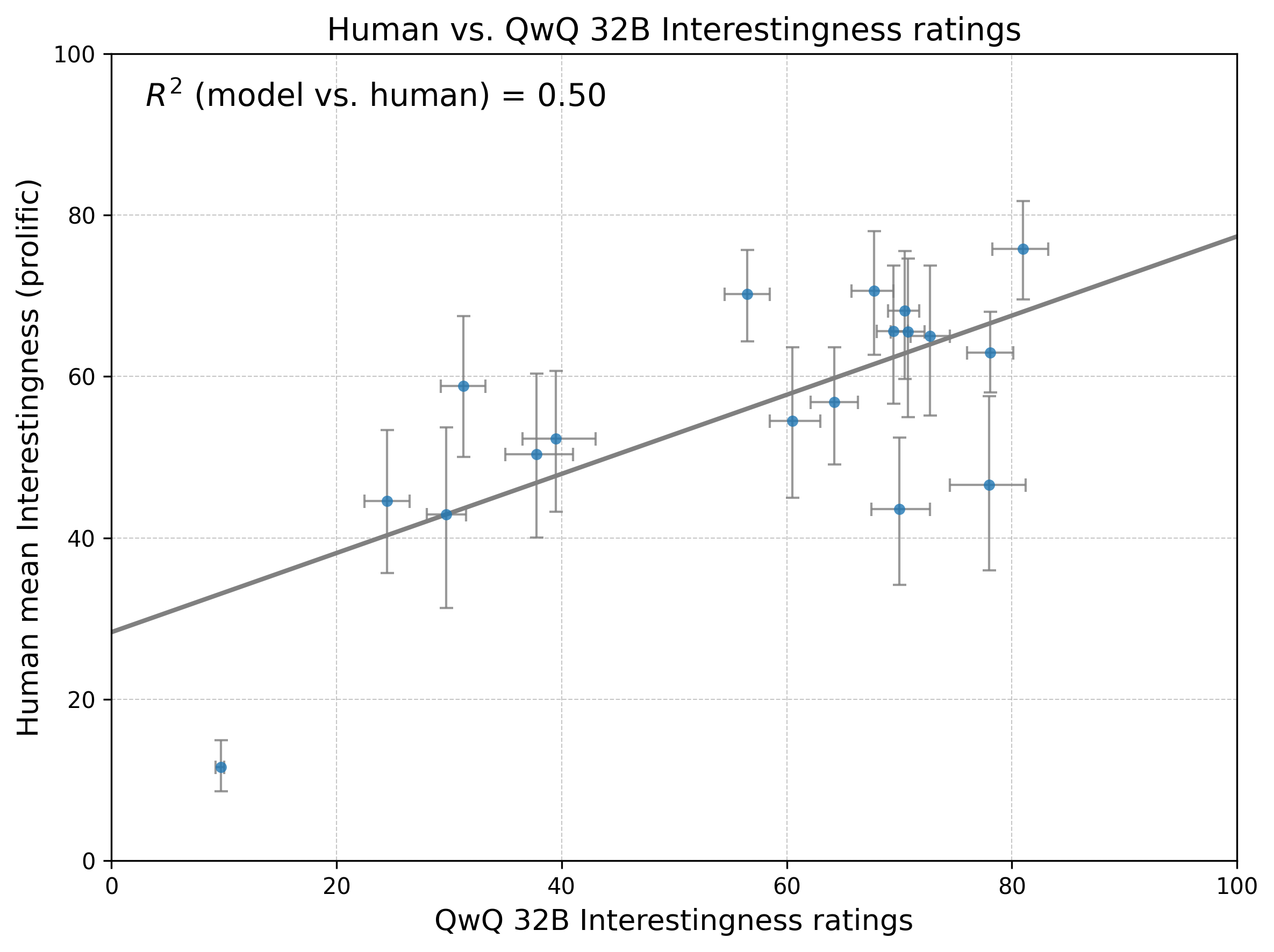} &
        \includegraphics[width=0.48\linewidth]{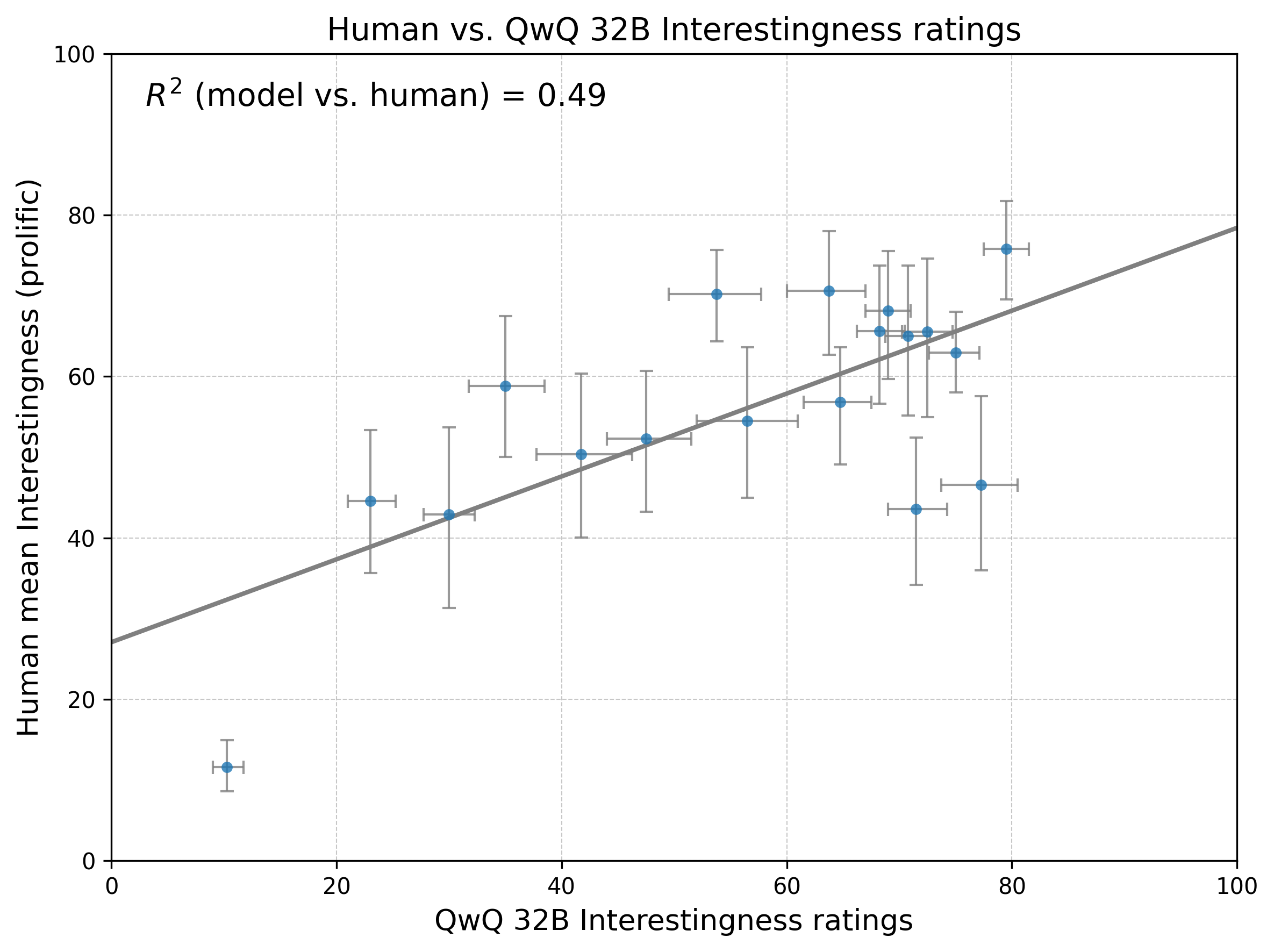} \\
        \rotatebox{90}{Difficulty} &
        \includegraphics[width=0.48\linewidth]{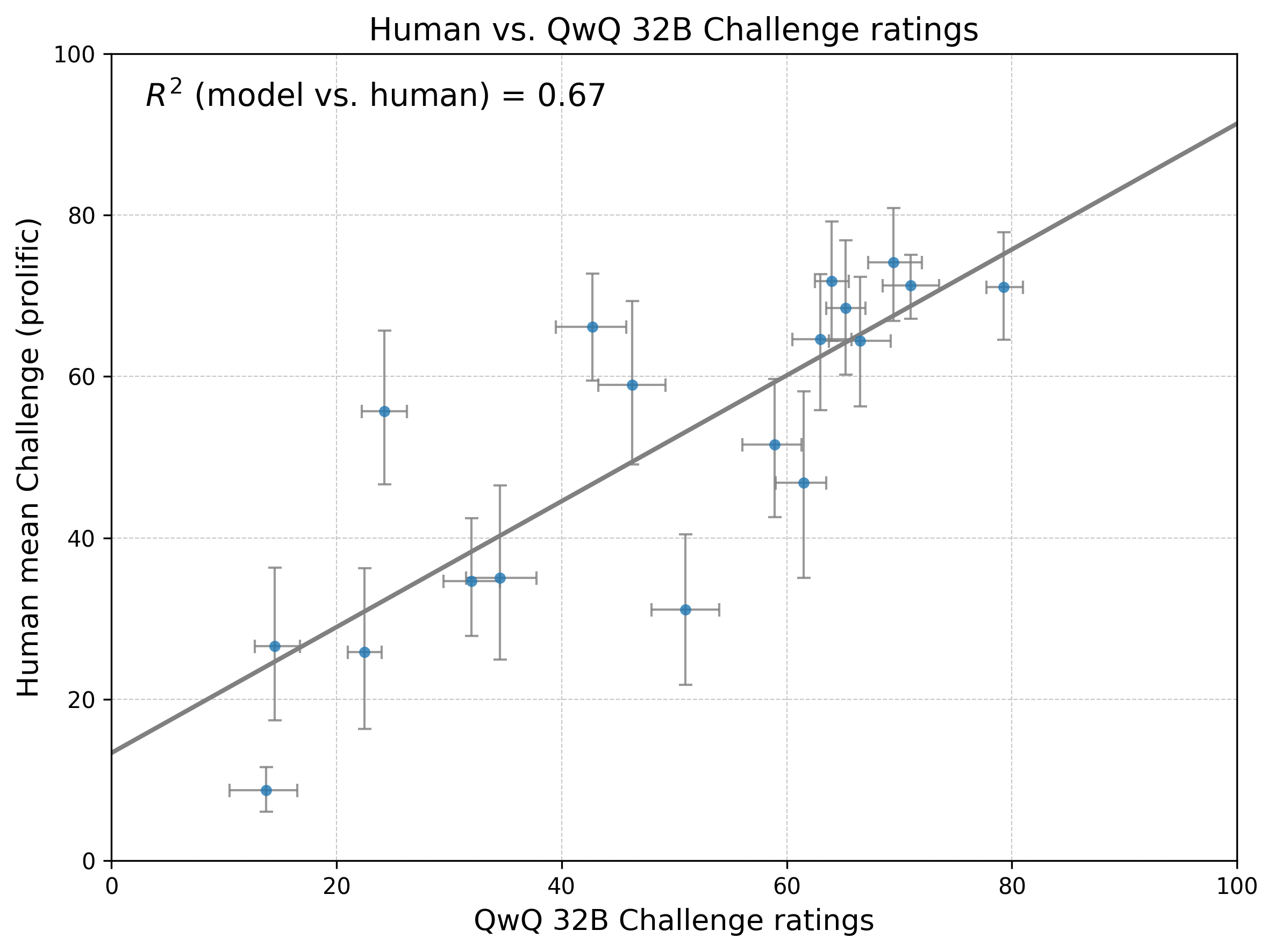} &
        \includegraphics[width=0.48\linewidth]{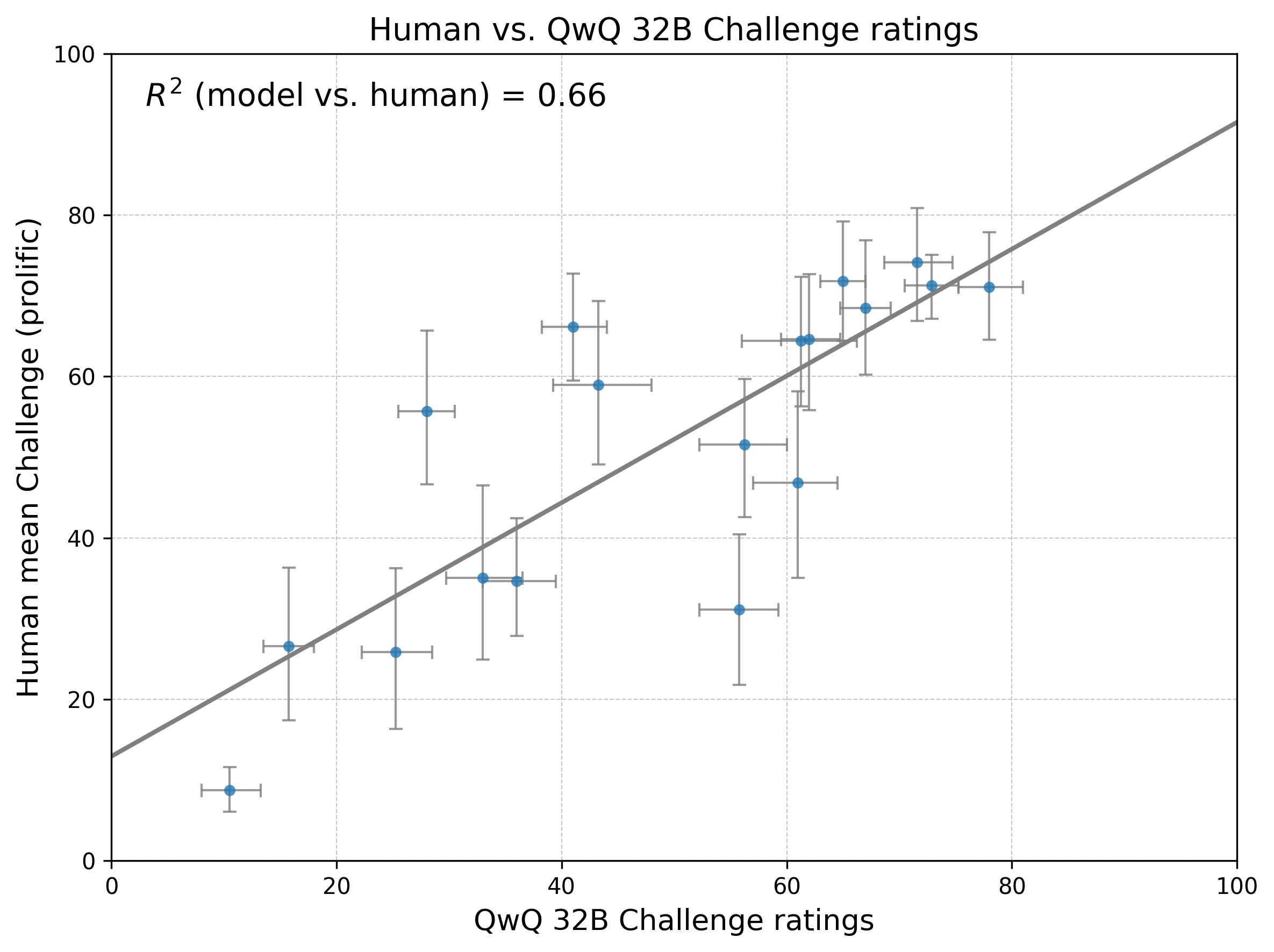} \\
    \end{tabular}
    \caption{QwQ-32B: Human vs LLM ratings}
    \label{fig:qwq_r2}
\end{figure}

\begin{figure}[htbp]
    \centering
    \begin{tabular}{c|cc}
        & \textbf{Temp 0.3} & \textbf{Temp 1.0} \\ \hline
        \rotatebox{90}{Interestingness} &
        \includegraphics[width=0.48\linewidth]{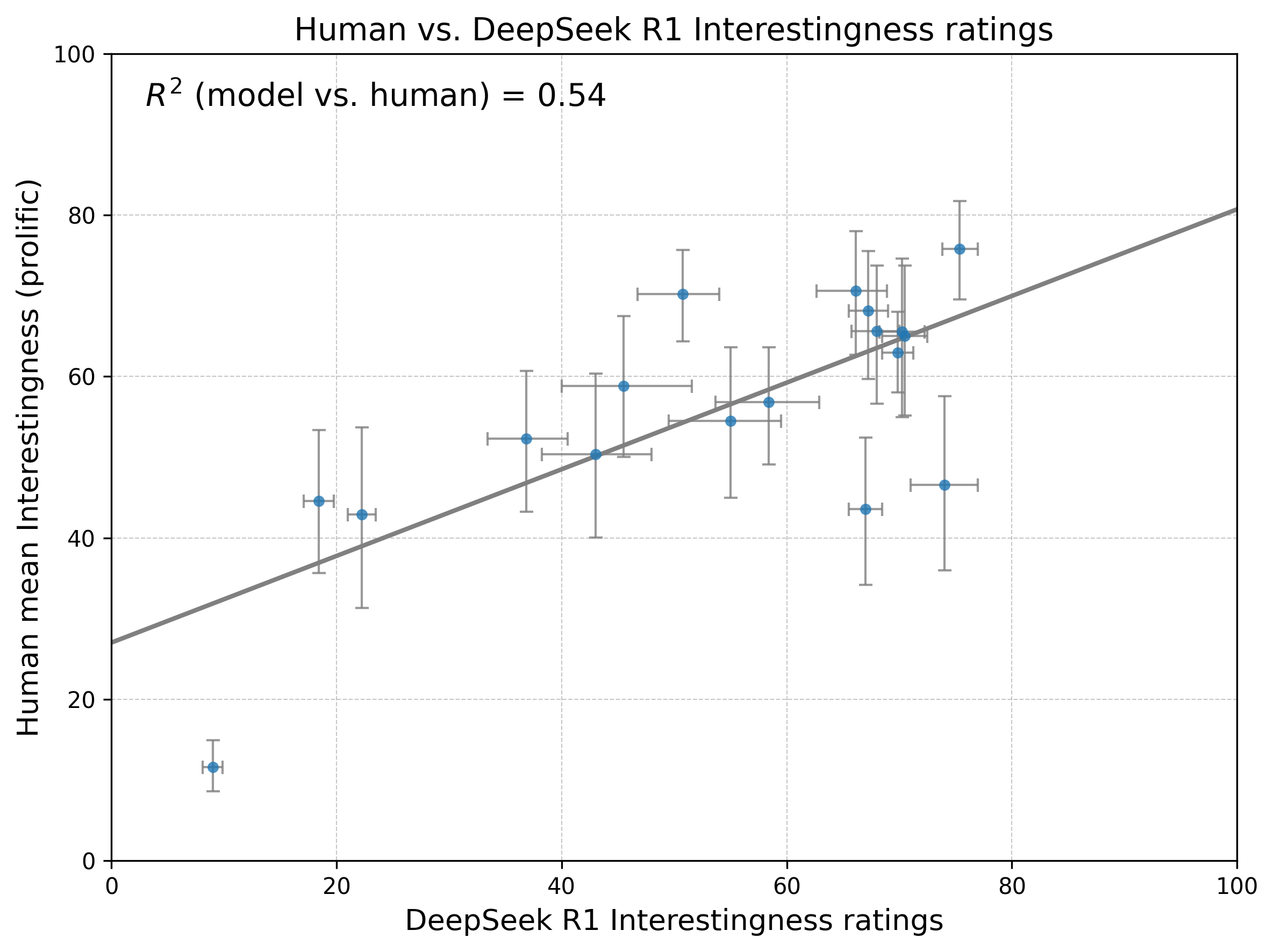} &
        \includegraphics[width=0.48\linewidth]{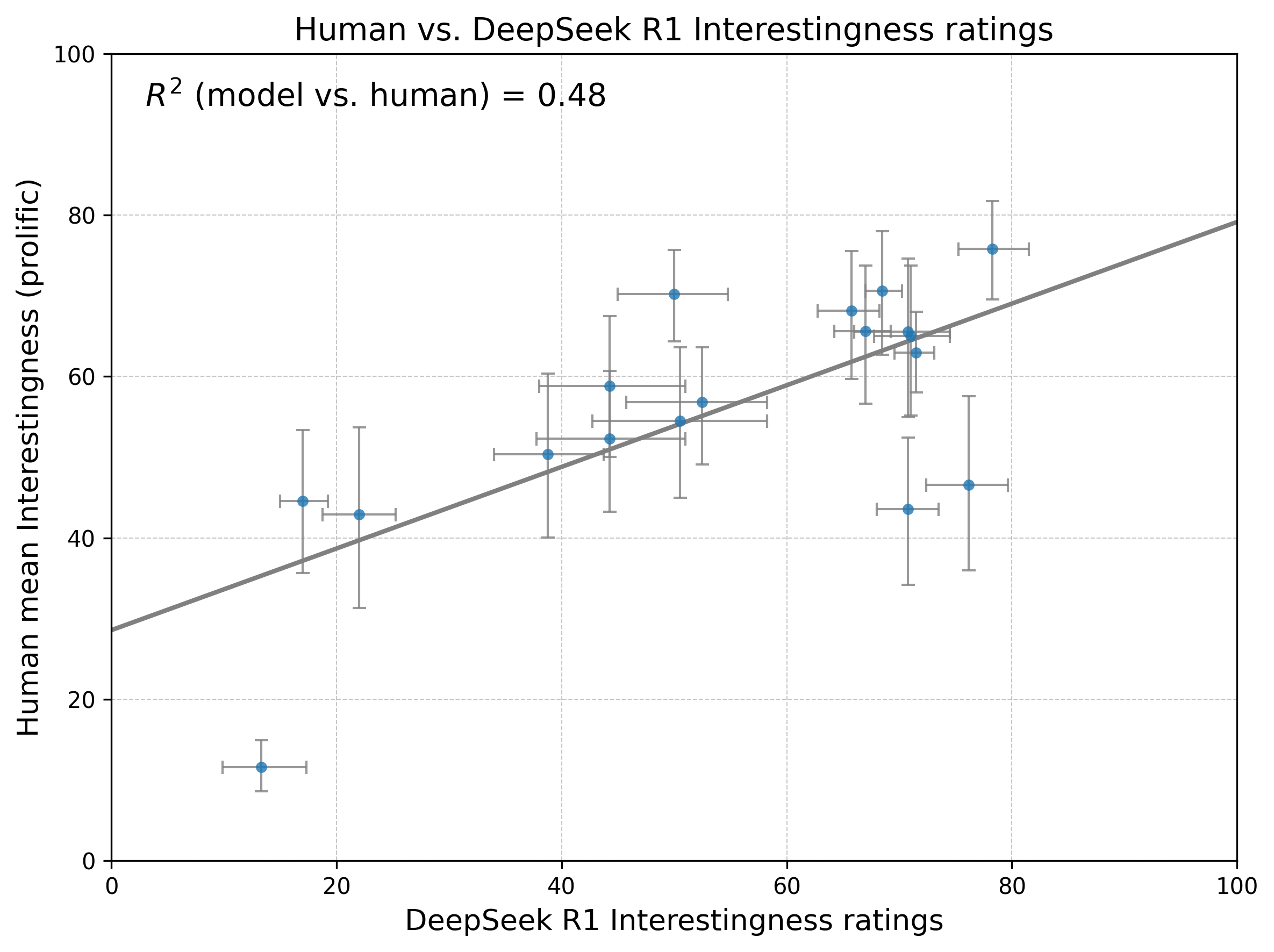} \\
        \rotatebox{90}{Difficulty} &
        \includegraphics[width=0.48\linewidth]{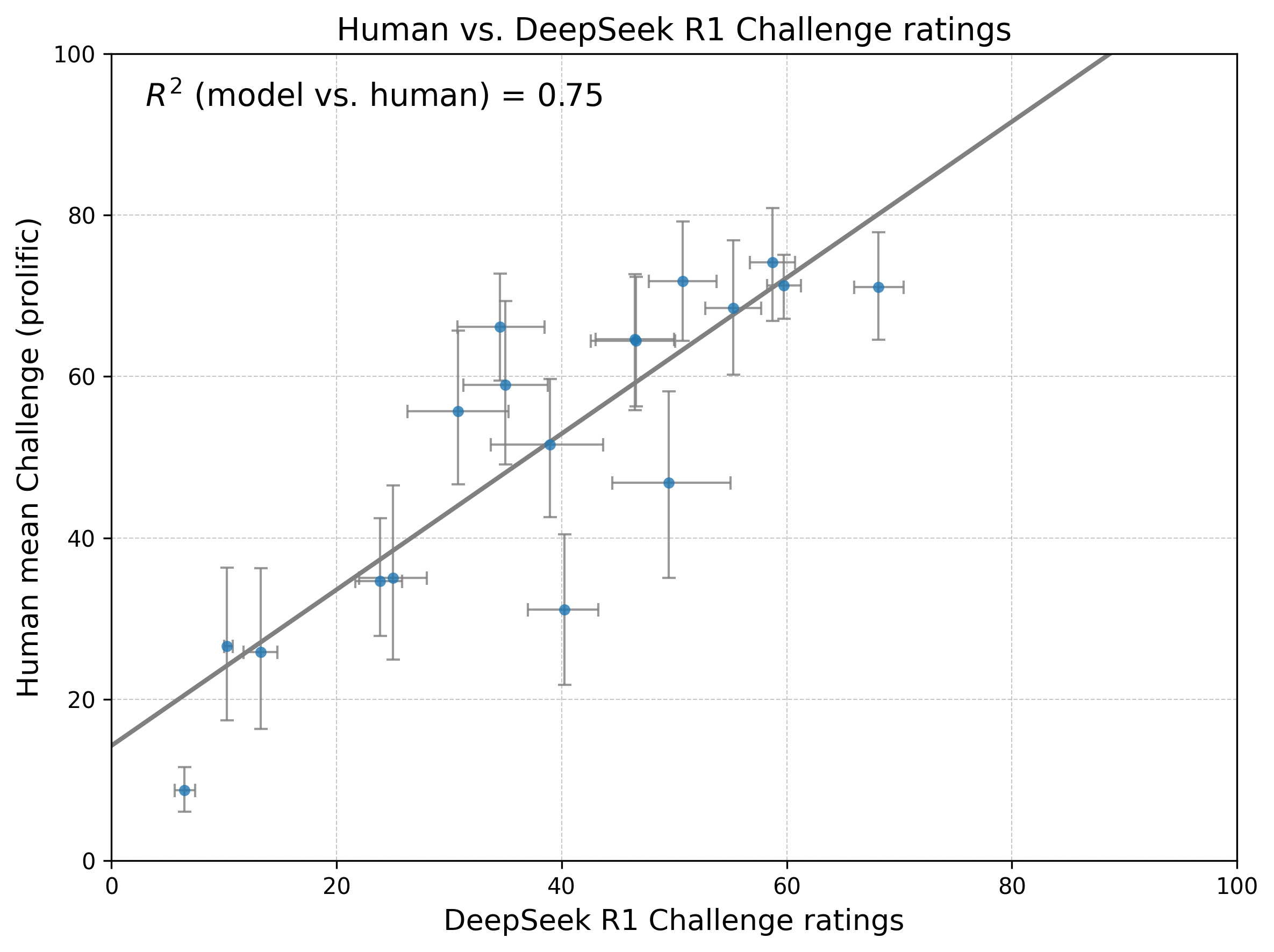} &
        \includegraphics[width=0.48\linewidth]{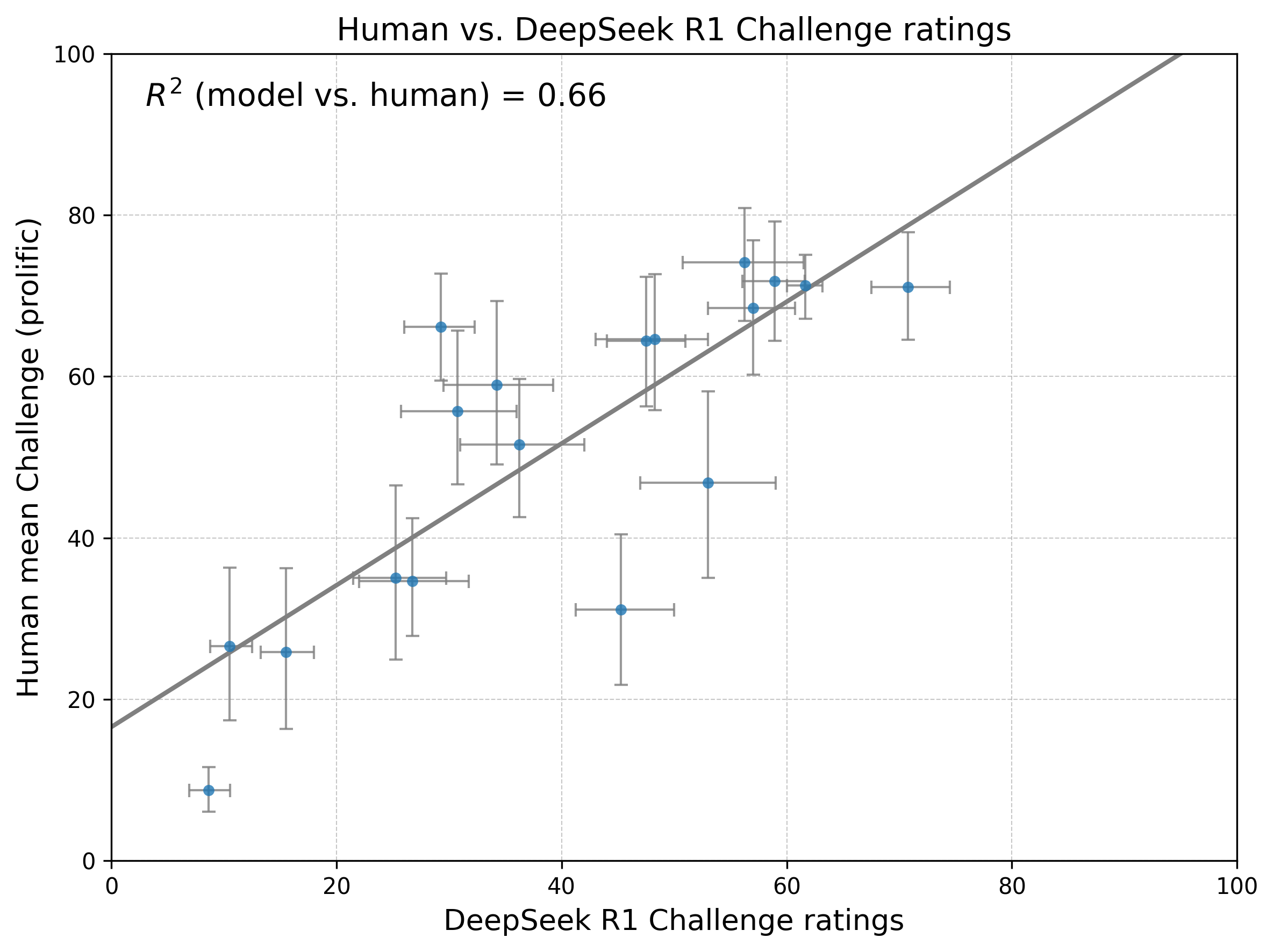} \\
    \end{tabular}
    \caption{Deepseek R1: Human vs LLM ratings}
    \label{fig:deepseek_r1_r2}
\end{figure}

We include the $R^2$ heatmaps for temperatures 0.3 and 1.0 in Figures~\ref{fig:app_prolific-heatmaps-t0.3} and ~\ref{fig:app_prolific-heatmaps-t1.0}. We also include information about the correlation between interestingness and difficulty scores between LLM families in Table~\ref{tab:llm_r2}. All correlation plots for interestingness judgments are included in Figures \ref{fig:gpt5_r2} through \ref{fig:deepseek_r1_r2}. We sample at two temperatures and also plot the $95\%$ confidence interval.

\begin{table}[b]
\centering
\small
\setlength{\tabcolsep}{6pt}
\caption{Correlation among each model family for interestingness and difficulty judgments. The “Mean pairwise $r$” column reports the mean pairwise Pearson correlation between models within their own family (computed over per-problem predictions). The "thinking" tag indicates reasoning models.}
\begin{tabular}{llcc}
\toprule
\textbf{Family} & \textbf{Model (tag)} & \textbf{Mean pairwise $r$} & \textbf{Mean pairwise $r$} \\
 &  & \textbf{(Interestingness)} & \textbf{(Difficulty)}  \\
 &  & \textbf{temp. 0.3, 1.0}  &  \textbf{temp. 0.3, 1.0}  \\
\midrule
\multirow{3}{*}{OpenAI} & o3 (thinking) & \multirow{3}{*}{-, 0.95} & \multirow{3}{*}{-, 0.98} \\
                         & GPT-5        &                        \\
                         & gpt-oss       &                       \\
\midrule
\multirow{2}{*}{Llama 4} & Llama-4 Maverick & \multirow{2}{*}{0.61, 0.65} & \multirow{2}{*}{0.79, 0.77}\\
                         & Llama-4 Scout    &                      \\
\midrule
\multirow{3}{*}{Qwen} & QwQ-32B (thinking)        & \multirow{3}{*}{0.89, 0.90} & \multirow{3}{*}{0.90, 0.90} \\
\multirow{3}{*}{} & Qwen3-235B-A22B-Instruct        & \multirow{3}{*}{}  \\
                      & Qwen3-235B-A22B-Thinking               &                       \\
\midrule
\multirow{3}{*}{Mistral} & Mistral-7B-Instruct-v0.1    & \multirow{3}{*}{0.96, 0.90} & \multirow{3}{*}{0.63, 0.73}\\
                       & Mistral-Small-24B-Instruct   &                        &  \\
                       & Mixtral-8x7B-Instruct-v0.1   &                        & \\
\midrule
\multirow{1}{*}{DeepSeek} & DeepSeek-R1 (thinking) & \multirow{1}{*}{-} & \multirow{1}{*}{-} \\
\bottomrule
\end{tabular}
\label{tab:llm_r2}
\end{table}

\subsubsection{Interestingness/Difficulty Correlations}

In Table~\ref{tab:llm_int_diff}, we report the per-problem correlation between interestingness and difficulty scores for each LLM.

\begin{table}[!t]
\centering
\caption{LLM interestingness--difficulty correlation (Pearson $R^2$) by model and temperature.}
\label{tab:llm_int_diff}
\small
\setlength{\tabcolsep}{8pt}
\begin{tabular}{l c c}
\toprule
\textbf{Model name} & \textbf{Temperature} & \textbf{$R^2$} \\
\midrule

\multirow{2}{*}{Llama-Scout}    & 0.30 & 0.58 \\
                               & 1.00 & 0.62 \\
\multirow{2}{*}{Llama-Maverick} & 0.30 & 0.84 \\
                               & 1.00 & 0.88 \\
\midrule

\multirow{2}{*}{DeepSeek-R1}    & 0.30 & 0.91 \\
                               & 1.00 & 0.90 \\
\midrule

o3                             & 1.00 & 0.96 \\
GPT-5                          & 1.00 & 0.86 \\
\multirow{2}{*}{GPT-OSS-120B}  & 0.30 & 0.84 \\
                               & 1.00 & 0.87 \\
\midrule

\multirow{2}{*}{Mixtral-8x7B-Instruct} & 0.30 & 0.87 \\
                                       & 1.00 & 0.89 \\
\multirow{2}{*}{Mistral-7B-Instruct}   & 0.30 & 0.61 \\
                                       & 1.00 & 0.41 \\
\multirow{2}{*}{Mistral-24B-Instruct}  & 0.30 & 0.83 \\
                                       & 1.00 & 0.87 \\
\midrule

\multirow{2}{*}{QwQ-32B}               & 0.30 & 0.93 \\
                                       & 1.00 & 0.93 \\
\multirow{2}{*}{Qwen-235B-Instruct}     & 0.30 & 0.84 \\
                                       & 1.00 & 0.86 \\
\multirow{2}{*}{Qwen-235B-Thinking}     & 0.30 & 0.92 \\
                                       & 1.00 & 0.90 \\

\bottomrule
\end{tabular}
\end{table}

\subsubsection{Wasserstein Distance Tables}
\label{app:wd_tables}
In Tables \ref{tab:wd_interestingness_temp03}, \ref{tab:wd_difficulty_temp03}, and \ref{tab:wd_difficulty_temp10}, we report the Wasserstein distances between the LLM and human distributions for interestingness and difficulty ate temperatures $0.3$ and $1.0$.

\begin{table}
\centering
\caption{Distributional alignment (WD) over \textbf{interestingness} judgments between human and LLM distributions of interestingness judgments on the Prolific dataset. Lower values indicate closer alignment to human distributions. The human-human split-half baseline indicates the amount of explainable variability in the human data. LLMs are sampled with a temperature of \textbf{0.3}.}
\begin{tabular}{lcllcl}
\toprule
\textbf{Model} & \textbf{WD} & \textbf{95\% CI} & \textbf{Model} & \textbf{WD} & \textbf{95\% CI} \\
\midrule
Human-Human & 9.5 & [7.8, 11.5] & Mixtral 8×7B Instruct & 21.3 & [18.9, 23.9] \\
DeepSeek R1 & 17.5 & [14.7, 20.5] & Llama 4 Scout & 21.4 & [19.0, 23.9] \\
Mistral 24B Instruct & 18.5 & [16.2, 20.9] & GPT-5 & 21.5 & [18.7, 24.6] \\
Mistral 7B Instruct & 19.0 & [16.5, 21.5] & Qwen 235B Instruct & 22.0 & [19.1, 24.9] \\
QwQ 32B & 19.3 & [16.6, 22.3] & Llama 4 Maverick & 22.0 & [19.0, 25.3] \\
GPT OSS 120B & 19.5 & [17.0, 22.2] & Qwen 235B Thinking & 20.4 & [16.9, 24.1] \\
\bottomrule
\end{tabular}
\label{tab:wd_interestingness_temp03}
\end{table}

\begin{table}
\centering
\caption{Distributional alignment (WD) over \textbf{difficulty} judgments between human and LLM distributions of difficulty judgments on the Prolific dataset. Lower values indicate closer alignment to human distributions. The human-human split-half baseline indicates the amount of explainable variability in the human data. LLMs are sampled with a temperature of \textbf{0.3}.}
\begin{tabular}{lcllcl}
\toprule
\textbf{Model} & \textbf{WD} & \textbf{95\% CI} & \textbf{Model} & \textbf{WD} & \textbf{95\% CI} \\
\midrule
Human–Human & 9.2 & [7.5, 11.0] & Mixtral 8×7B Instruct & 20.1 & [17.5, 23.1] \\
Mistral 24B Instruct & 17.3 & [15.2, 19.8] & Mistral 7B Instruct & 20.2 & [15.9, 24.7] \\
QwQ 32B & 17.6 & [15.3, 20.3] & Qwen 235B Instruct & 20.5 & [17.7, 23.6] \\
Qwen 235B Thinking & 18.3 & [15.2, 22.1] & GPT OSS 120B & 21.3 & [17.7, 25.6] \\
Llama 4 Maverick & 19.0 & [16.3, 21.9] & GPT-5 & 29.1 & [24.7, 33.3] \\
Llama 4 Scout & 19.2 & [17.1, 21.4] & DeepSeek R1 & 19.2 & [16.8, 21.8] \\
\bottomrule
\end{tabular}
\label{tab:wd_difficulty_temp03}
\end{table}

\begin{table}
\centering
\caption{Distributional alignment (WD) over \textbf{difficulty} judgments between human and LLM distributions of difficulty judgments on the Prolific dataset. Lower values indicate closer alignment to human distributions. The human-human split-half baseline indicates the amount of explainable variability in the human data. LLMs are sampled at a temperature of \textbf{1.0}.}
\begin{tabular}{lcllcl}
\toprule
\textbf{Model} & \textbf{WD} & \textbf{95\% CI} & \textbf{Model} & \textbf{WD} & \textbf{95\% CI} \\
\midrule
Human–Human & 9.2 & [7.5, 11.0] & Mixtral 8×7B Instruct & 18.2 & [15.5, 21.3] \\
Mistral 24B Instruct & 13.3 & [11.6, 15.0] & Llama 4 Scout & 18.6 & [16.6, 20.6] \\
Mistral 7B Instruct & 16.0 & [11.8, 20.5] & Llama 4 Maverick & 19.1 & [16.3, 21.7] \\
QwQ 32B & 16.4 & [13.8, 19.3] & Qwen 235B Instruct & 20.2 & [17.6, 22.9] \\
Qwen 235B Thinking & 17.4 & [14.5, 20.9] & GPT OSS 120B & 20.7 & [17.2, 25.0] \\
DeepSeek R1 & 18.0 & [15.2, 21.3] & o3 & 23.7 & [20.0, 27.8] \\
\bottomrule
\end{tabular}
\label{tab:wd_difficulty_temp10}
\end{table}

\subsubsection{LRM Judgment Lengths}
\label{app:flash_judgment}
\begin{figure*}[!t]
    \centering
    \includegraphics[width=0.99\linewidth]{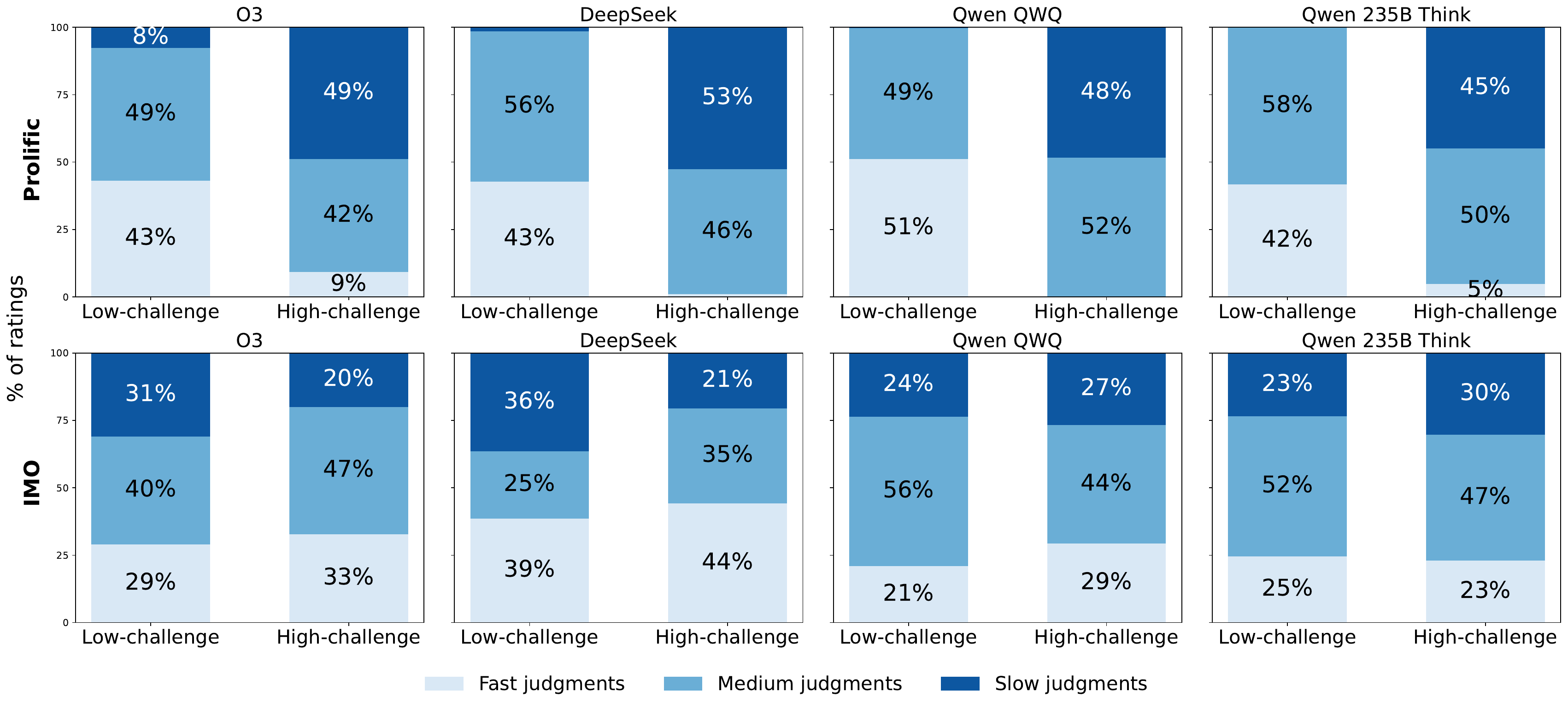}
    \caption{Judgment speed distributions across LRMs on low- vs.\ high-interest Prolific problems. A low-challenge problem for a model is one that is given a below median challenge score by the model. High-interest problems are those which are given higher than the median challenge score. For each subset of problems, we label whether a judgment was slow, medium, or fast, based on the distribution of reasoning token counts for that model. Slow judgments occupy the bottom quartile and fast judgments occupy the top one, with medium-speed judgments covering the middle. We see that LRMs tend to engage in longer reasoning chains for problems that they ultimately label as being more difficult.}
    \label{fig:flash_judgments_difficulty}
\end{figure*}

In Figure~\ref{fig:flash_judgments_difficulty}, we examine the resource usage LRMs engaged when reasoning about a problem, which we measure via the amount of reasoning tokens used (i.e., the length of their reasoning chain). We use this to explore the distinction in resources used to assess problems that LRMs judge as low- vs.\ high-difficulty (which is judged by whether a problem's difficulty score is below or above the median of difficulty scores from judgments from that LRM). 

\subsubsection{Split-half $R^2$ between independent human raters.}
\label{app:human_noise_ceiling}
To compare the LLM-human $R^2$ to the correlation among independent groups of humans, we calculate the split-half $R^2$, for which we repeatedly and randomly split human interestingness scores for each problem into two groups, and calculate the correlation between those two groups. The mean split-half $R^2$ among humans was $0.71$, with a $95\%$ confidence interval of $[0.53, 0.87]$.

\subsection{IMO Survey Results}
\label{app:add_results_imo}
\textbf{Elegance played a key role in human interestingness judgments of the IMO problems.} For the IMO survey, each participant rated the interestingness of four problems and selected interestingness/uninterestingness rationales for their rating. In Appendix Figure \ref{fig:hist_int_unint}, we include a histogram of the frequency of different rationales for interestingness/uninterestingness. The three most frequently marked reasons for interestingness were ``the problem statement is simple and elegant'', ``the solution does not require any sophisticated techniques/theorems'', and ``the solution is elegant''. We also include a correlation matrix depicting when people chose multiple reasons for interestingness/uninterestingness for the same problem in Appendix Figure~\ref{fig:int_unint_corr}.\looseness-1

\begin{figure}[ht]
    \centering
    \includegraphics[width=\linewidth]{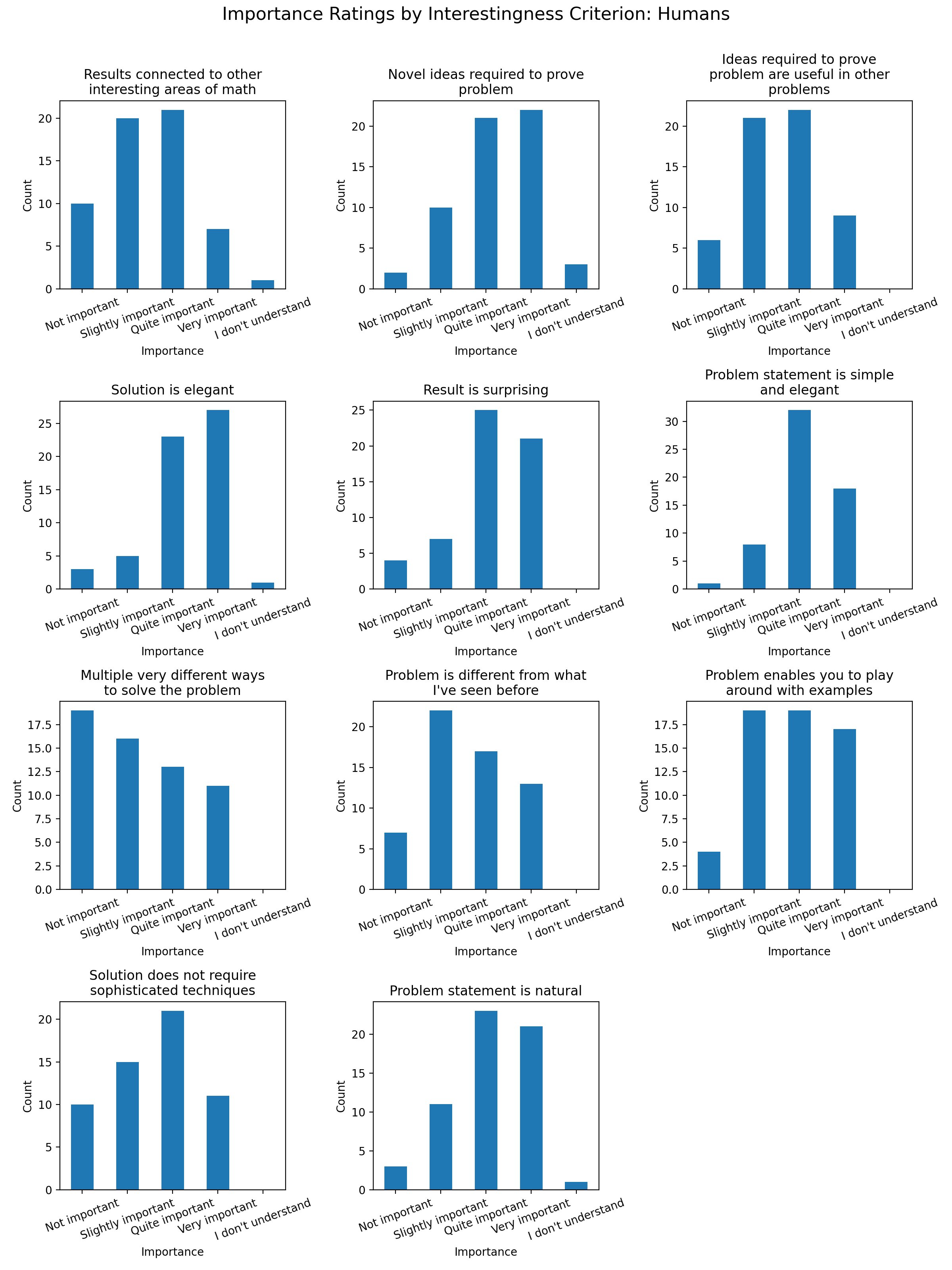}
    \caption{Distribution of importance ratings from human participants across 11 interestingness criteria.}
    \label{fig:human_interestingness_importance}
\end{figure}

\begin{figure}[ht]
    \centering
    \includegraphics[width=\linewidth]{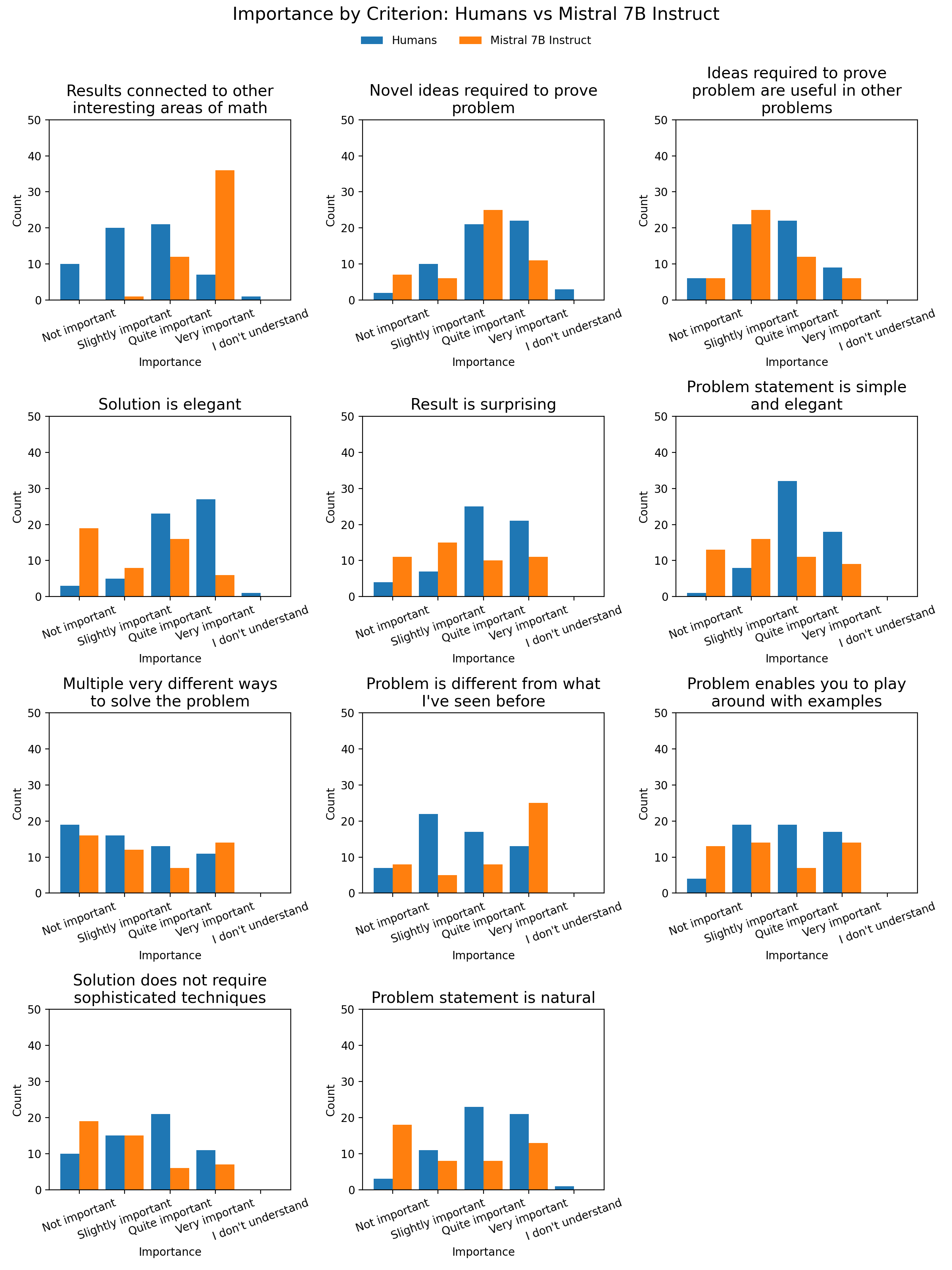}
    \caption{Importance ratings from Mistral 7B Instruct (orange) across 11 interestingness criteria compared to averaged human importance ratings (blue).}
    \label{fig:mistral_7b_interestingness_importance}
\end{figure}

\begin{figure}[ht]
    \centering
    \includegraphics[width=\linewidth]{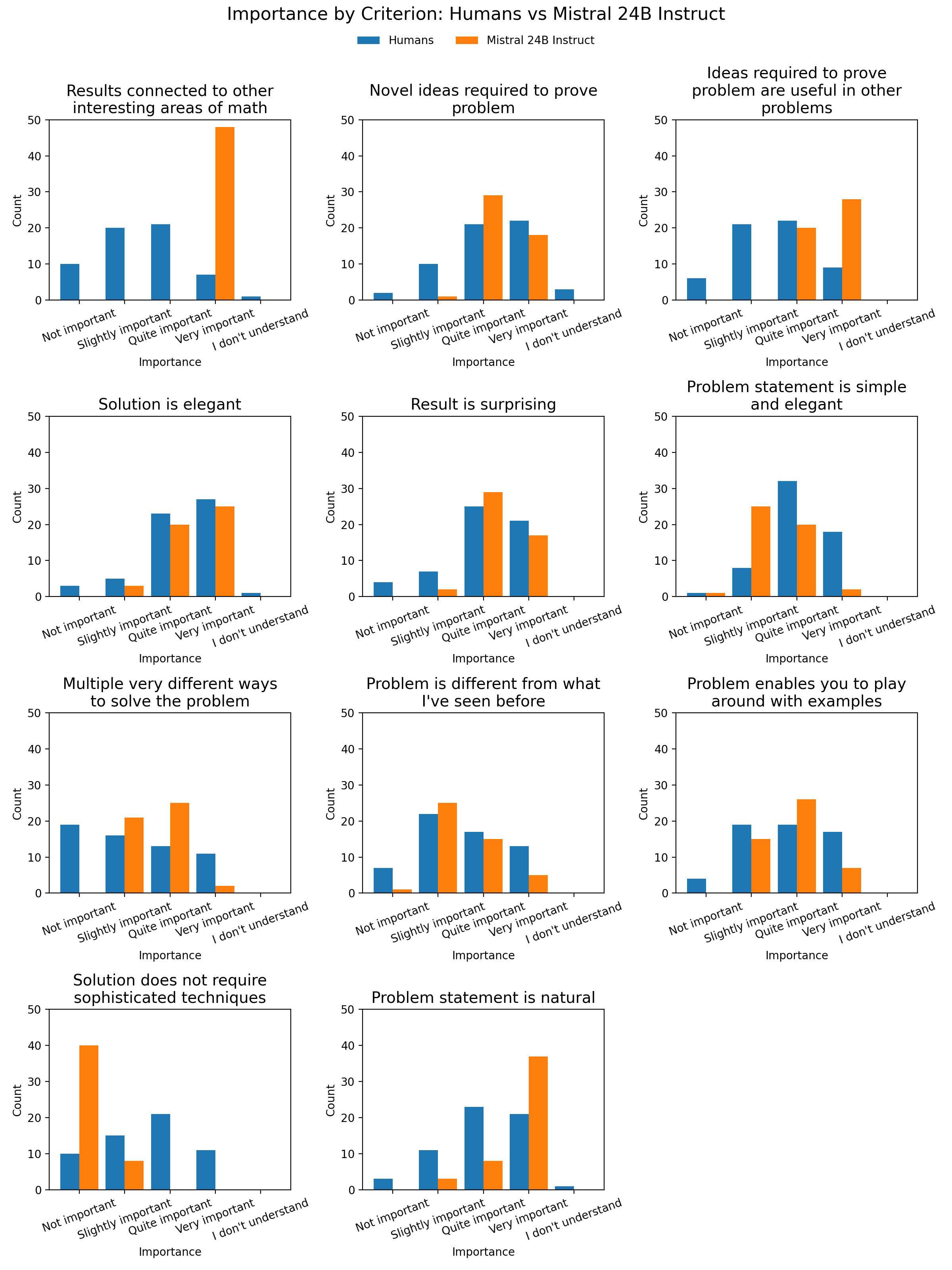}
    \caption{Importance ratings from \textbf{Mistral 24B Instruct}.}
    \label{fig:mistral_24b_interestingness_importance}
\end{figure}

\begin{figure}[ht]
    \centering
    \includegraphics[width=\linewidth]{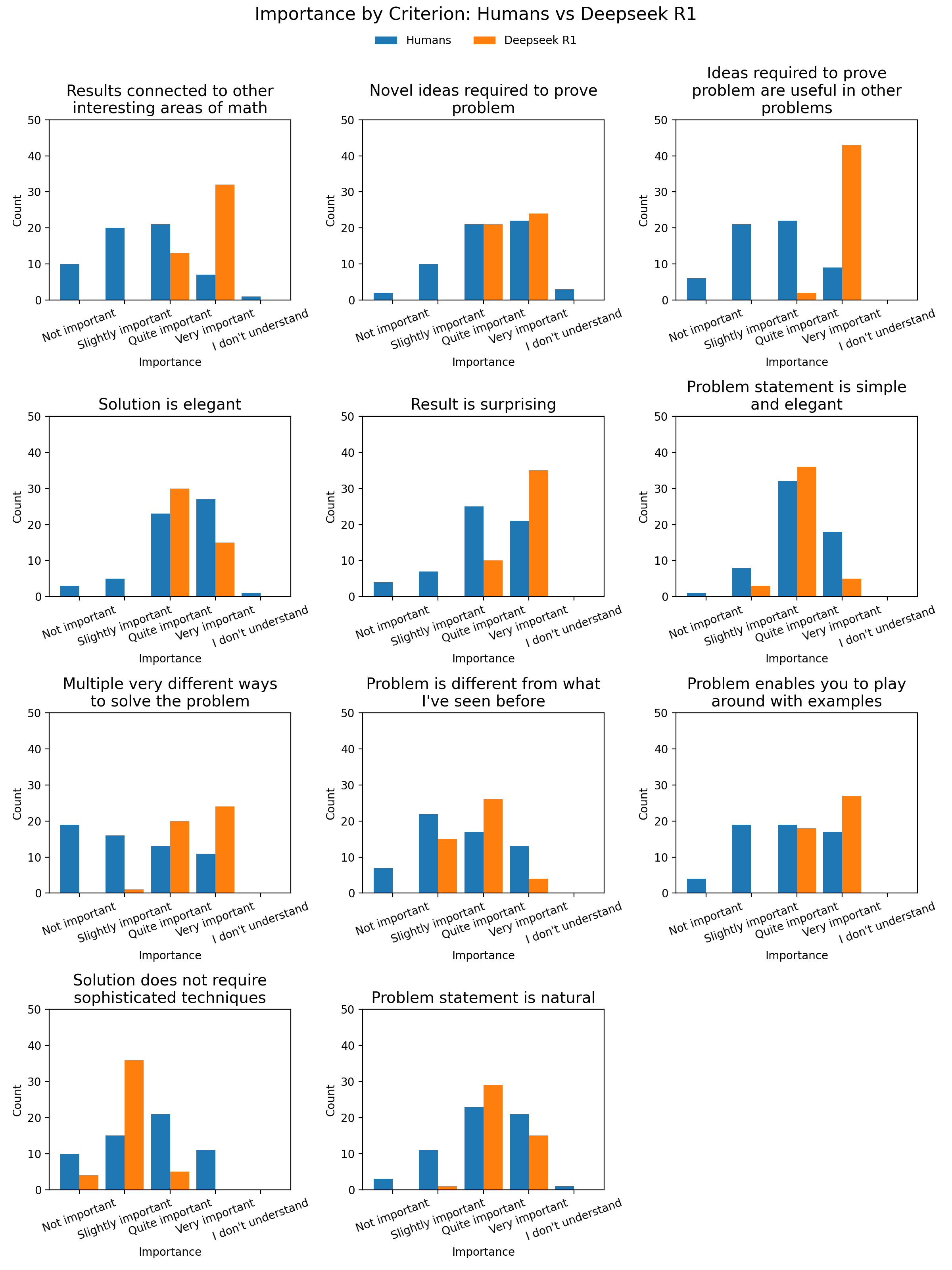}
    \caption{Importance ratings from \textbf{DeepSeek R1}.}
    \label{fig:r1_interestingness_importance}
\end{figure}

\paragraph{Reasons for interestingness across LLMs.} In Figure~\ref{fig:human_interestingness_importance}, we show histograms of human participants importance ratings for various interestingness criteria. Figures~\ref{fig:mistral_7b_interestingness_importance} through \ref{fig:r1_interestingness_importance} includes this for the LLMs we examine. 

\begin{figure}[htbp]
    \centering
    \begin{tabular}{c|cc}
        & \textbf{Temp 0.3} & \textbf{Temp 1.0} \\ \hline
        \rotatebox{90}{Interestingness} &
        \includegraphics[width=0.48\linewidth]{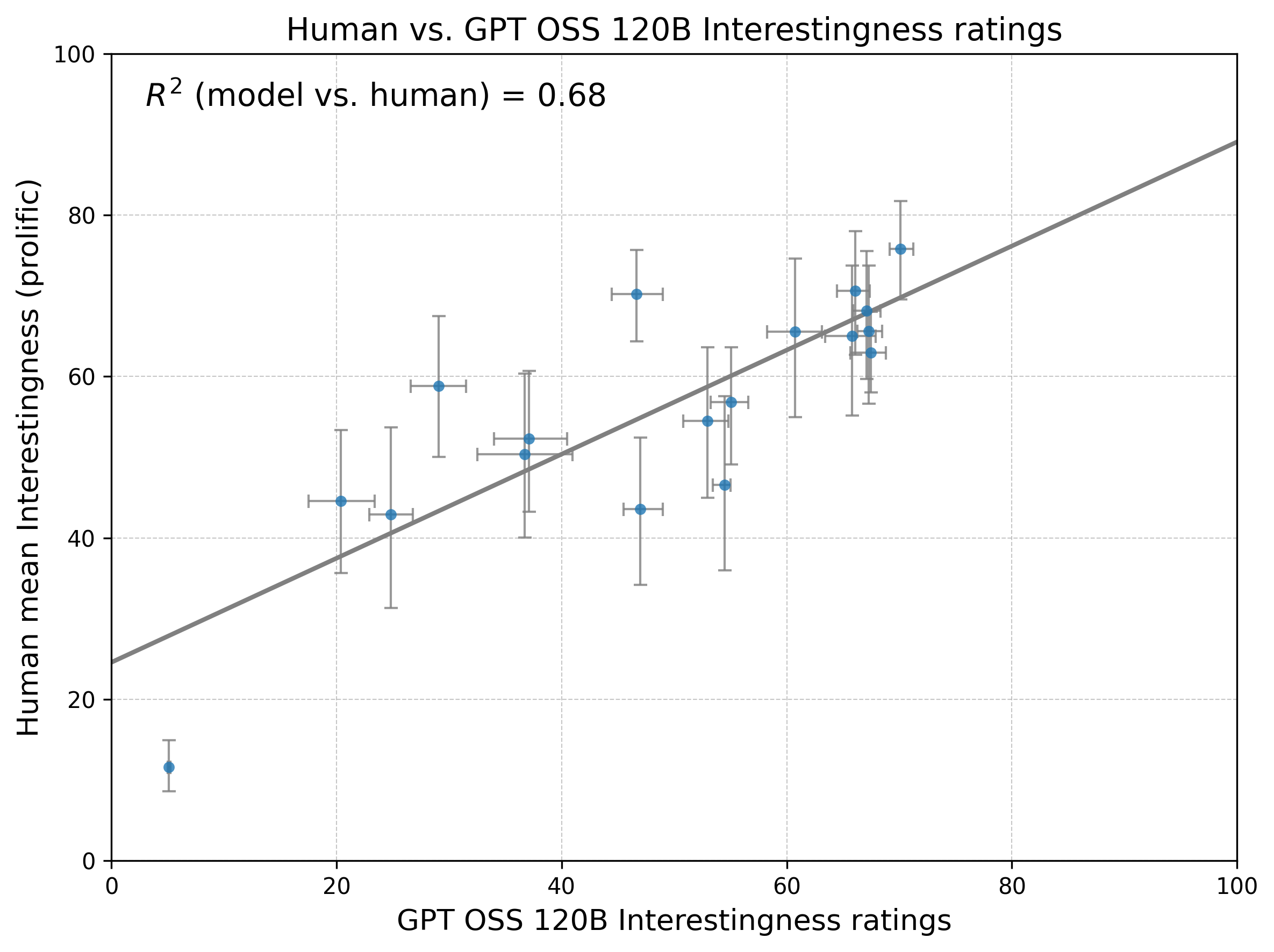} &
        \includegraphics[width=0.48\linewidth]{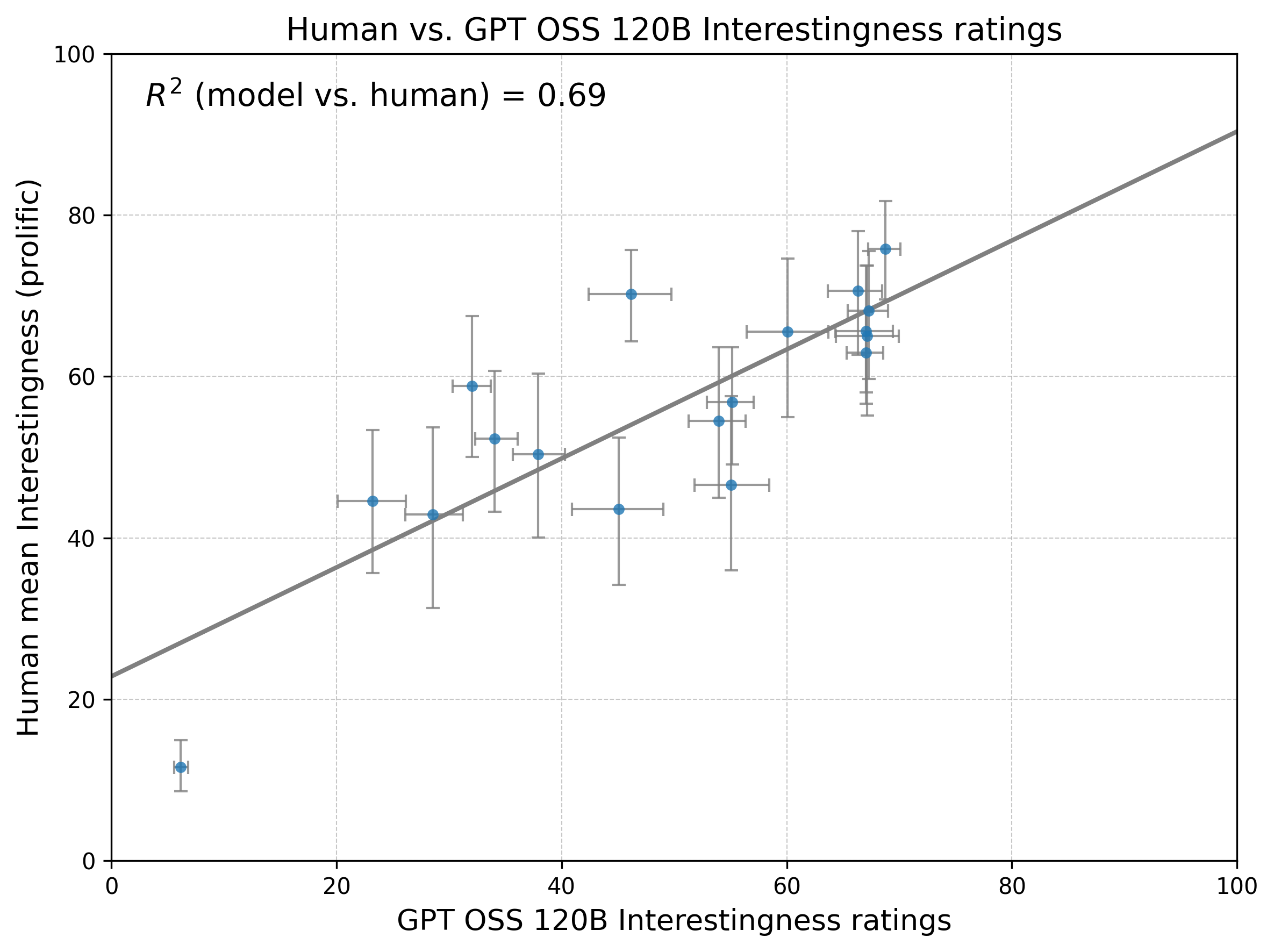} \\
        \rotatebox{90}{Difficulty} &
        \includegraphics[width=0.48\linewidth]{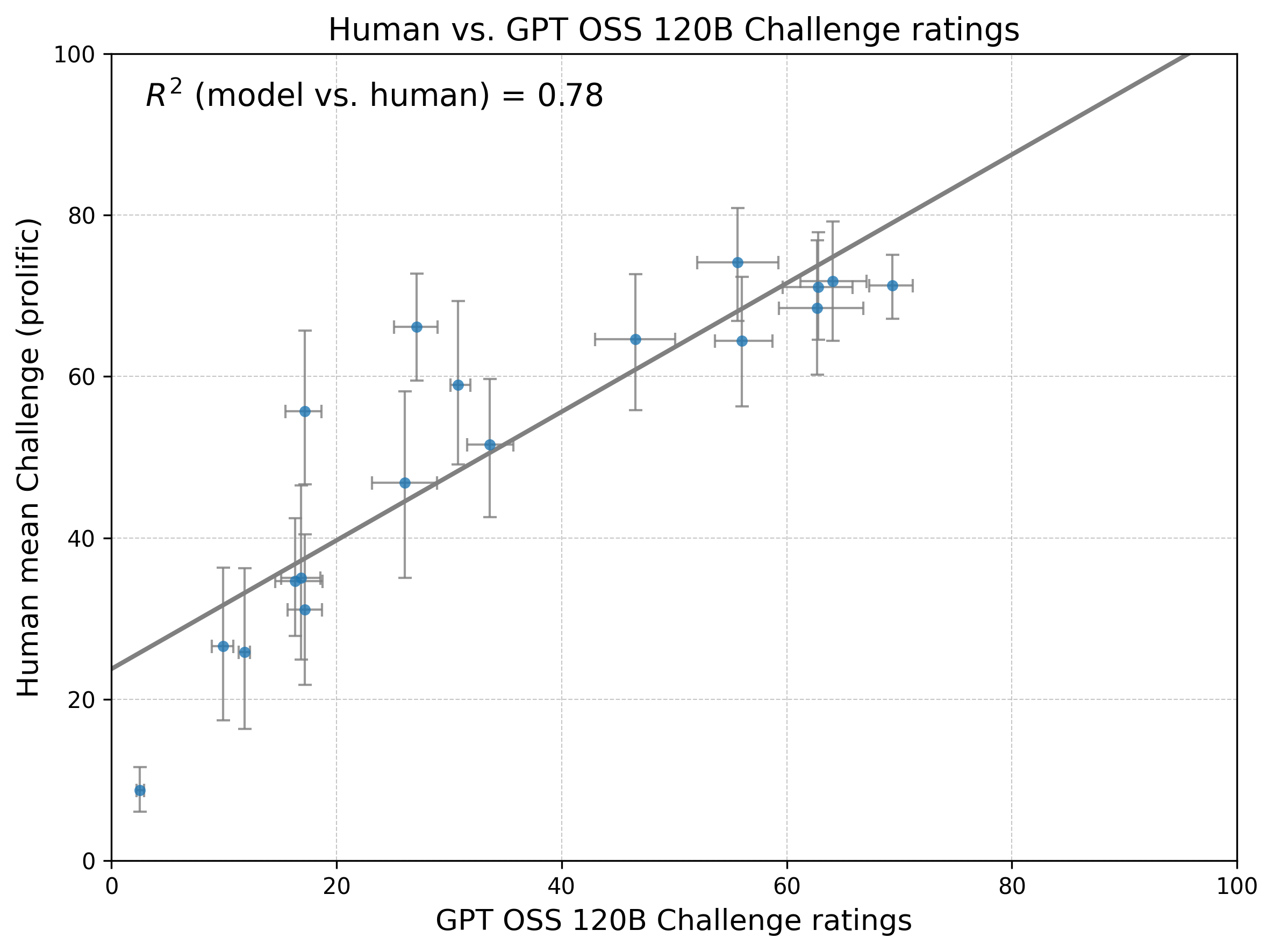} &
        \includegraphics[width=0.48\linewidth]{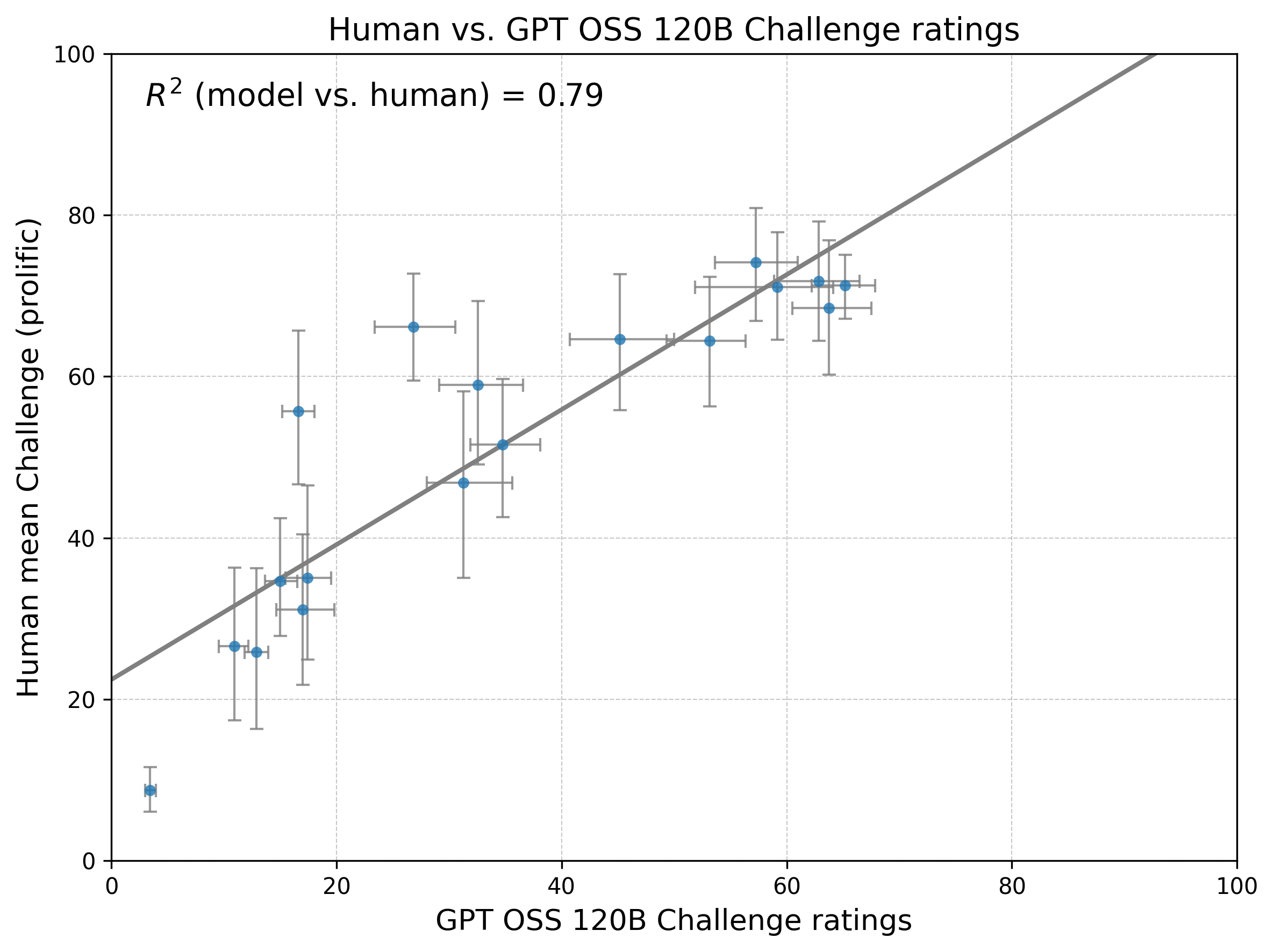} \\
    \end{tabular}
    \caption{GPT-OSS-120B: Human vs LLM ratings}
\end{figure}

\begin{figure}[htbp]
    \centering
    \begin{tabular}{c|cc}
        & \textbf{Temp 0.3} & \textbf{Temp 1.0} \\ \hline
        \rotatebox{90}{Interestingness} &
        \includegraphics[width=0.48\linewidth]{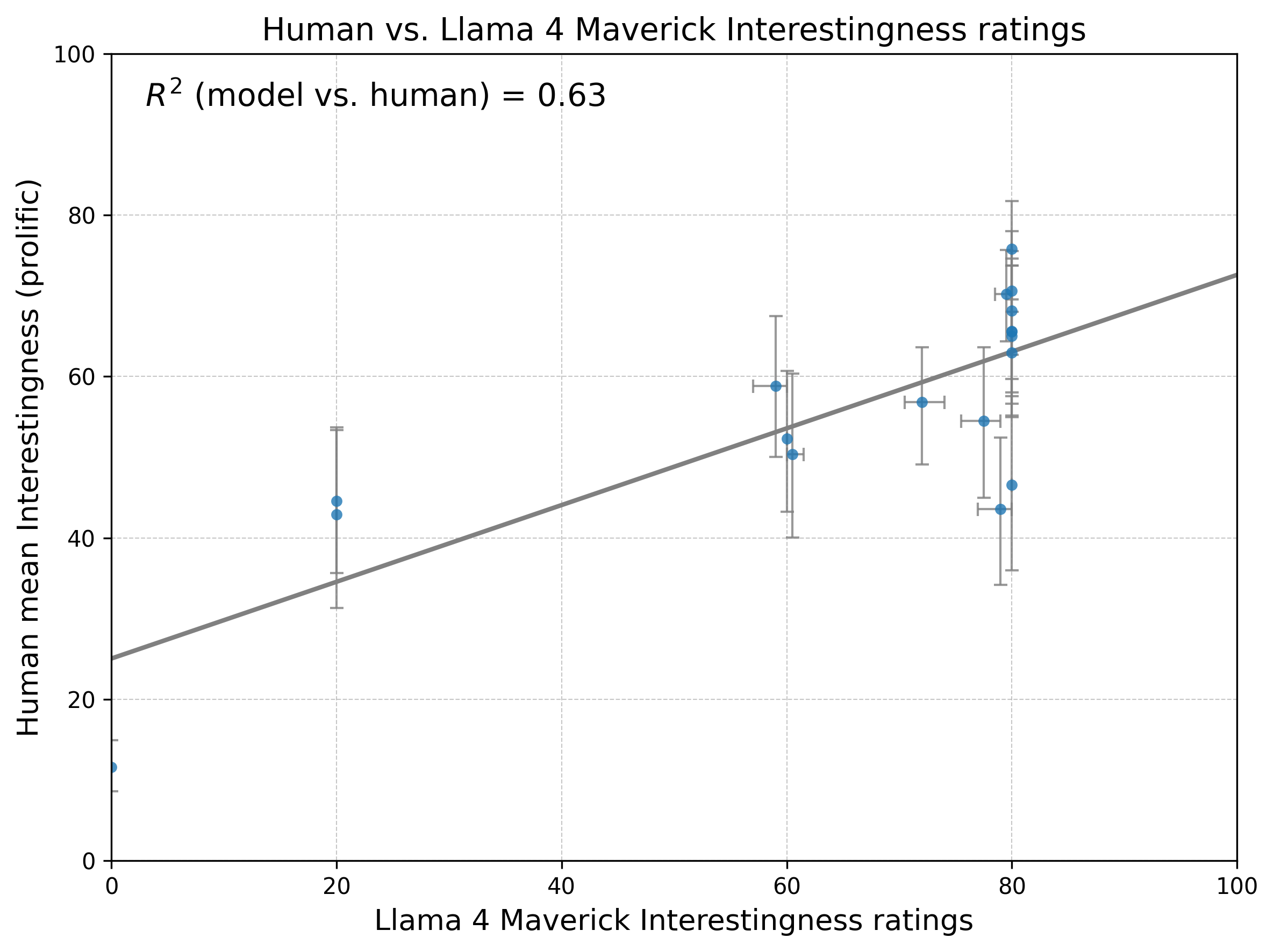} &
        \includegraphics[width=0.48\linewidth]{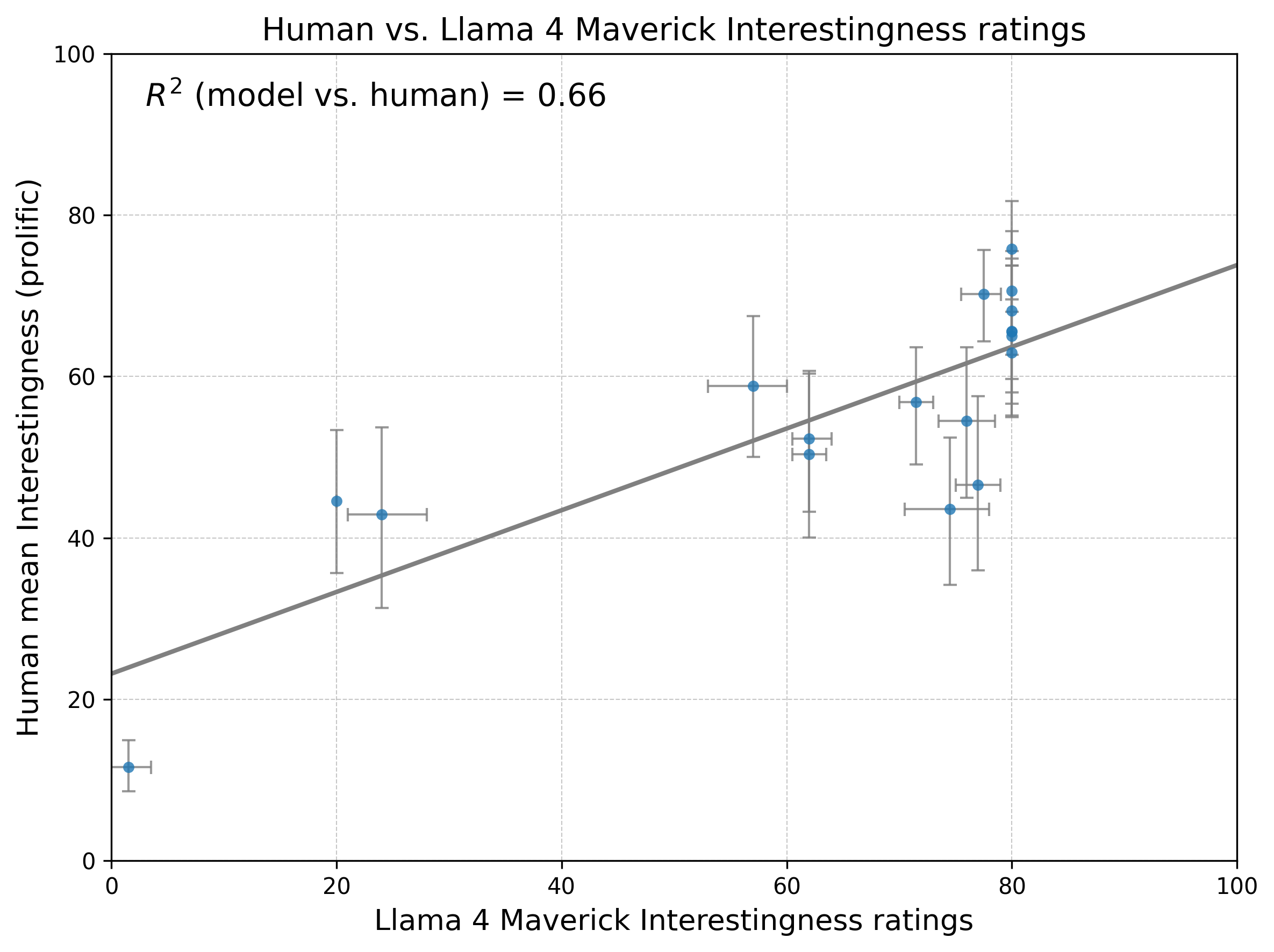} \\
        \rotatebox{90}{Difficulty} &
        \includegraphics[width=0.48\linewidth]{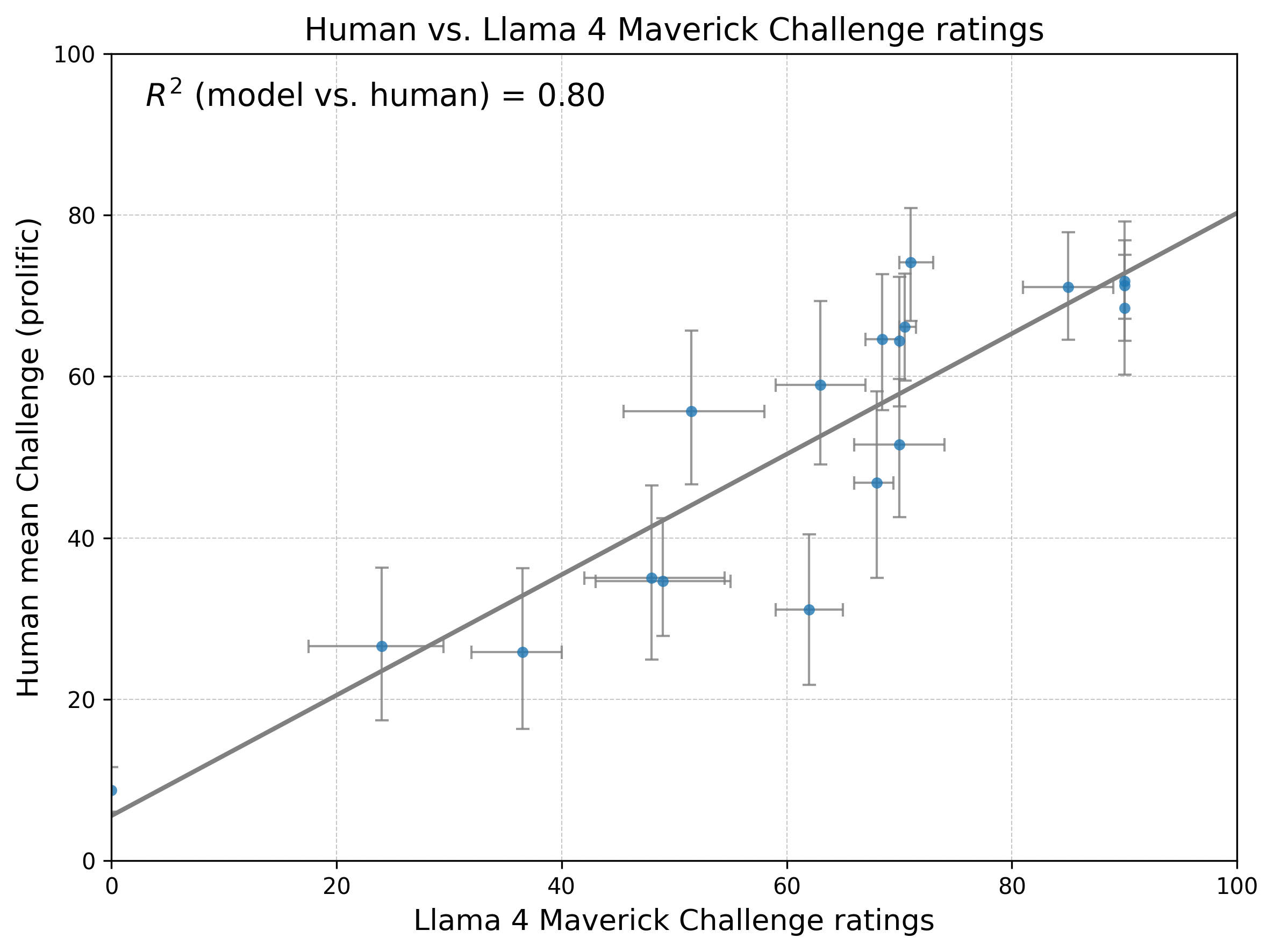} &
        \includegraphics[width=0.48\linewidth]{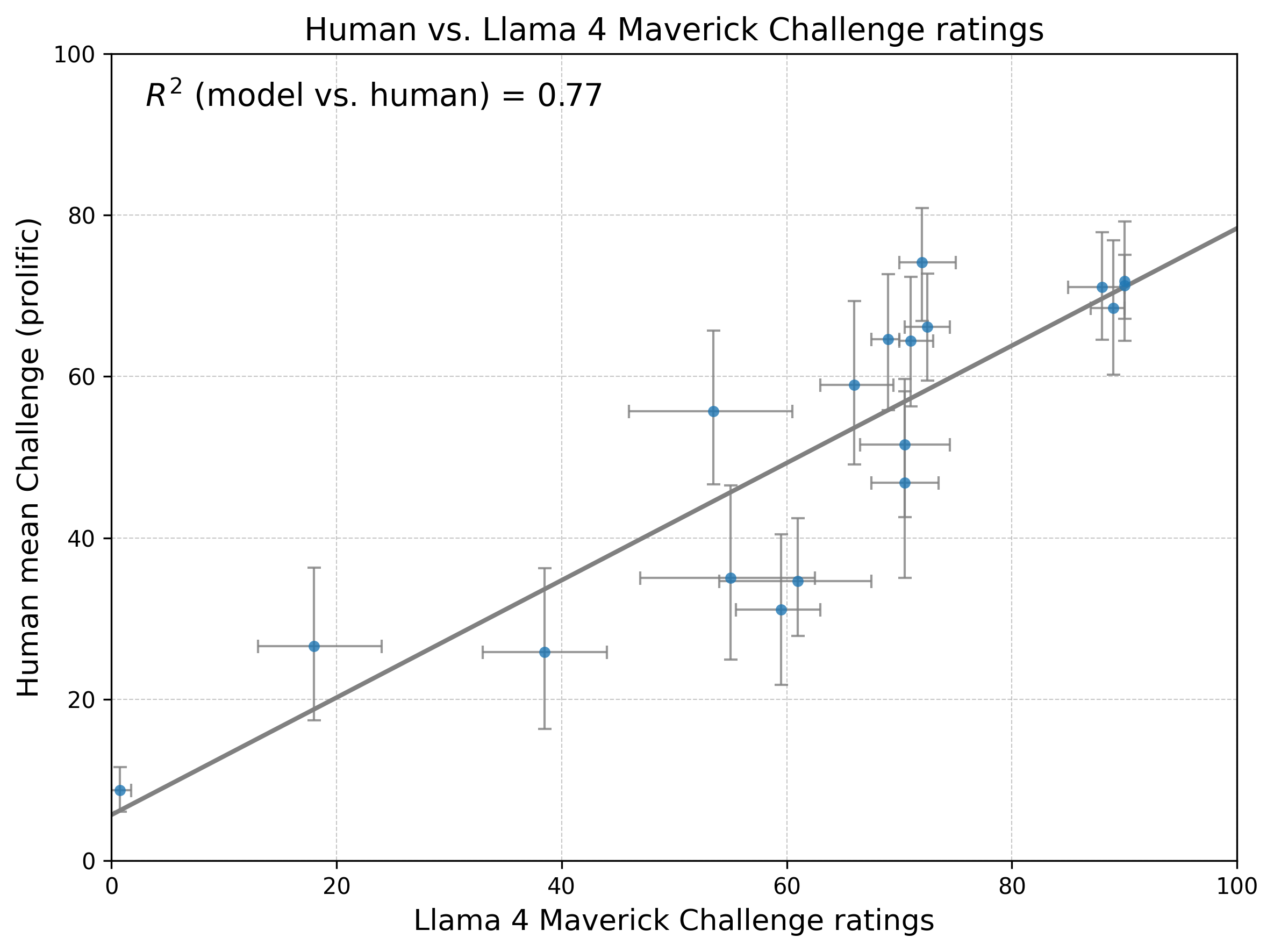} \\
    \end{tabular}
    \caption{LLaMA-Maverick: Human vs LLM ratings}
\end{figure}

\begin{figure}[htbp]
    \centering
    \begin{tabular}{c|cc}
        & \textbf{Temp 0.3} & \textbf{Temp 1.0} \\ \hline
        \rotatebox{90}{Interestingness} &
        \includegraphics[width=0.48\linewidth]{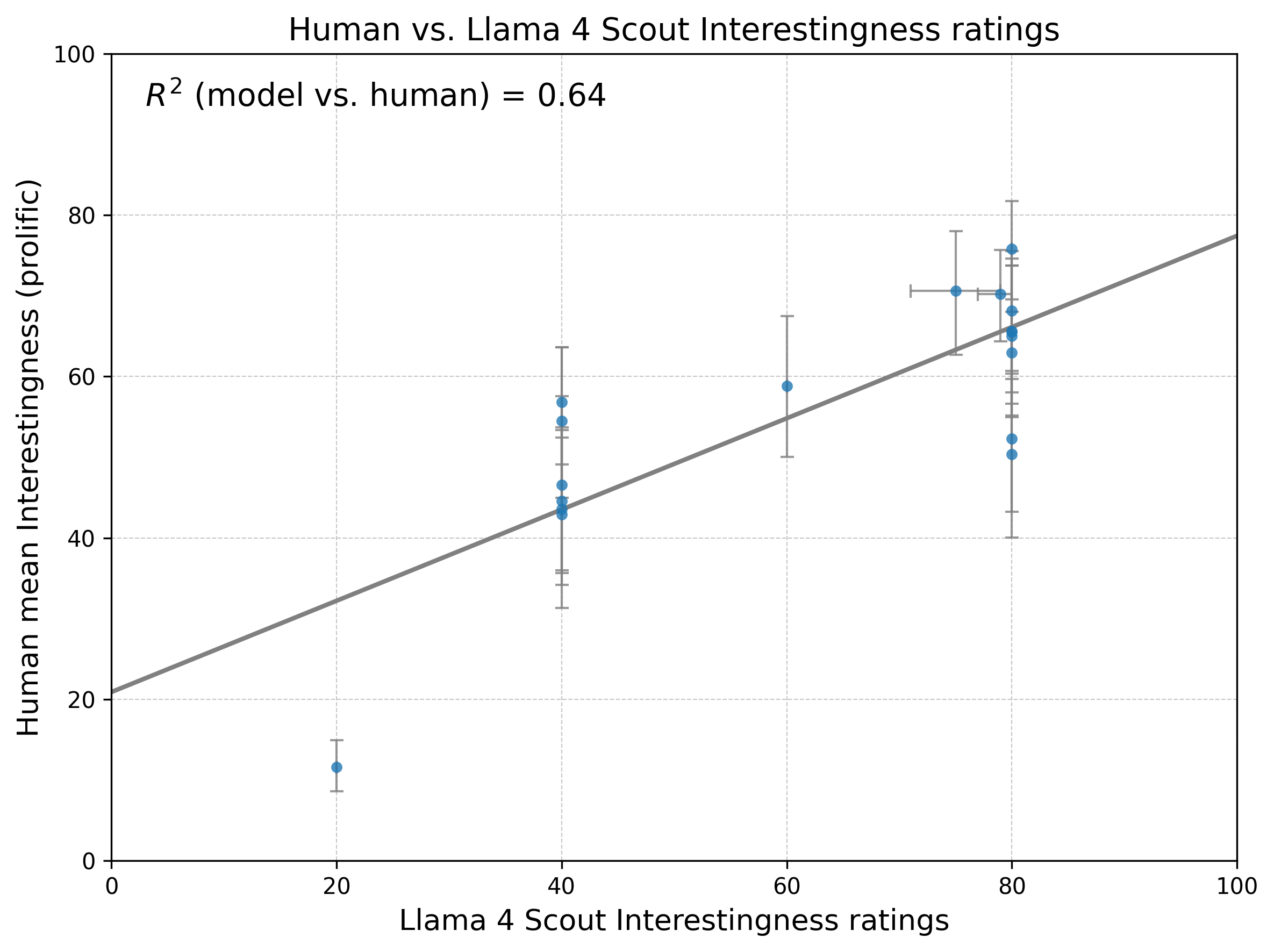} &
        \includegraphics[width=0.48\linewidth]{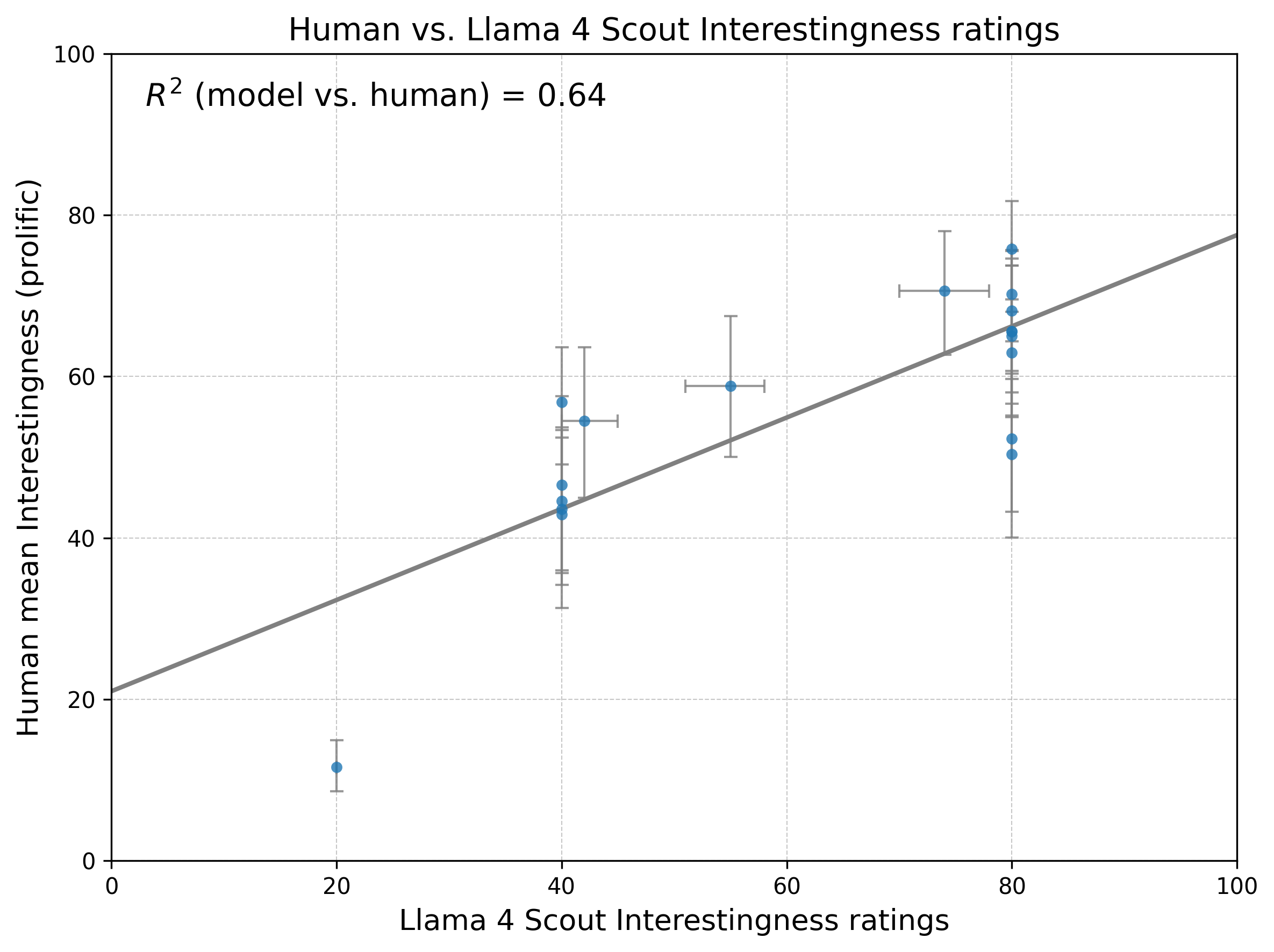} \\
        \rotatebox{90}{Difficulty} &
        \includegraphics[width=0.48\linewidth]{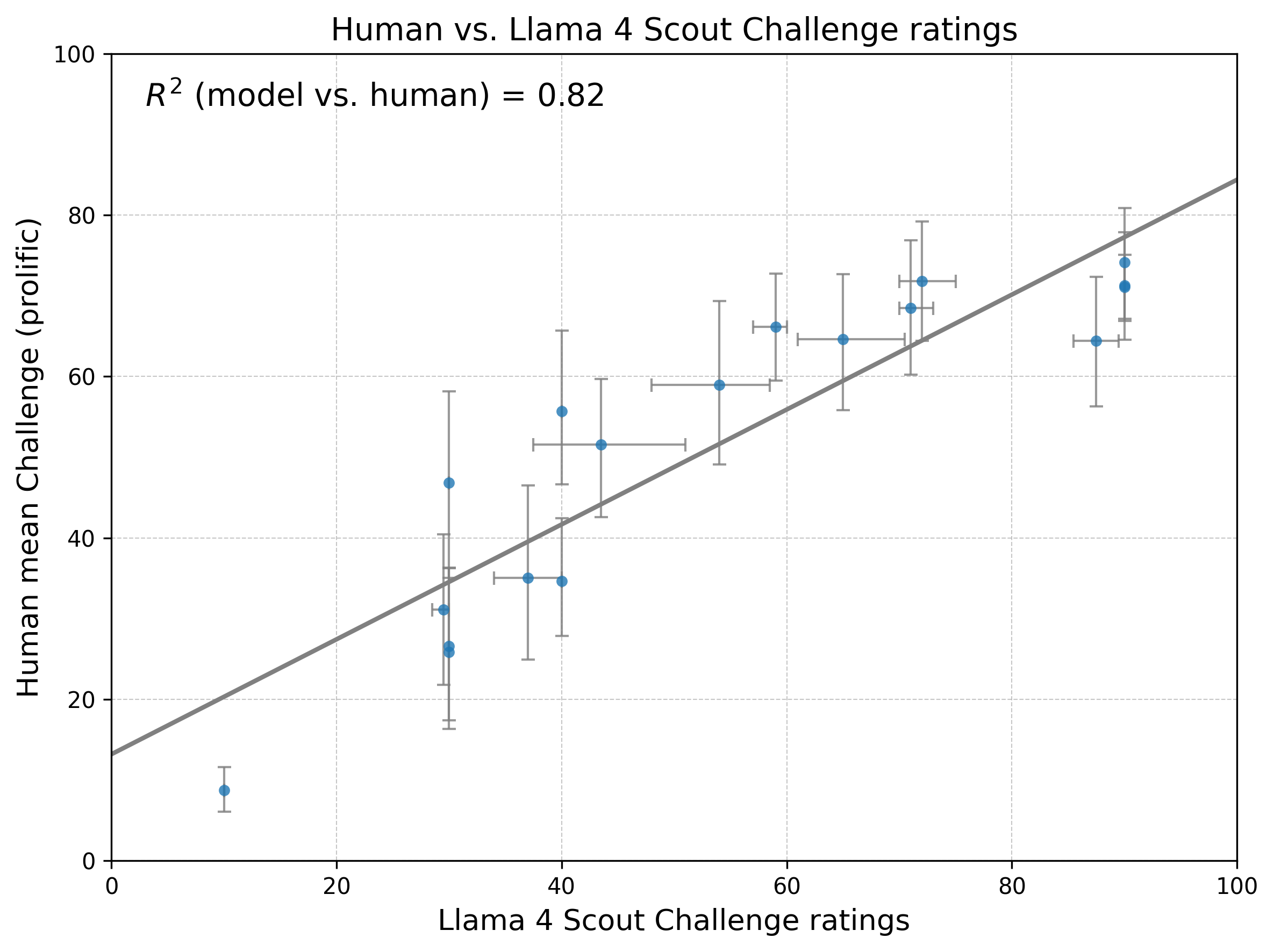} &
        \includegraphics[width=0.48\linewidth]{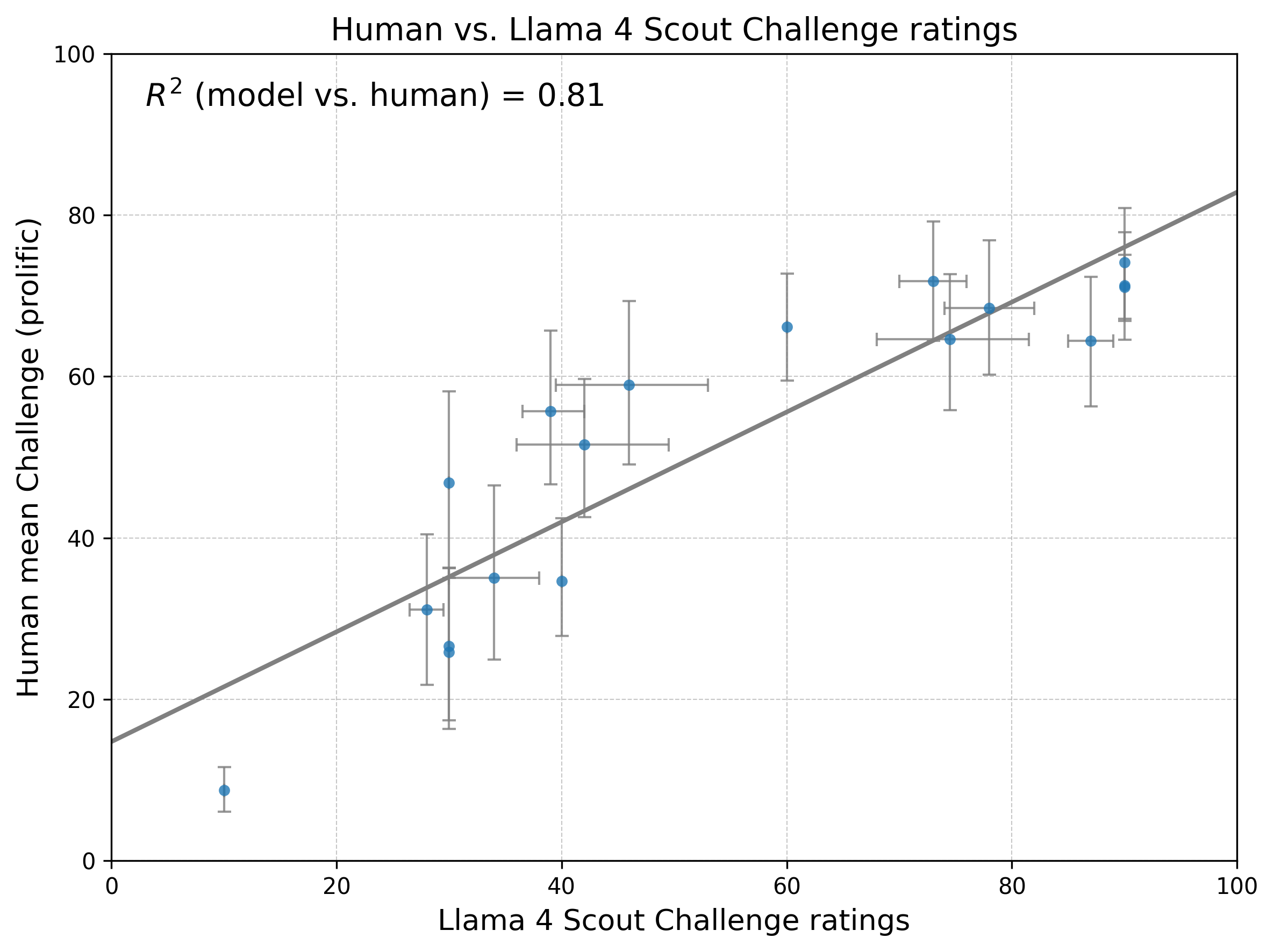} \\
    \end{tabular}
    \caption{LLaMA-Scout: Human vs LLM ratings}
\end{figure}

\begin{figure}[htbp]
    \centering
    \begin{tabular}{c|cc}
        & \textbf{Temp 0.3} & \textbf{Temp 1.0} \\ \hline
        \rotatebox{90}{Interestingness} &
        \includegraphics[width=0.48\linewidth]{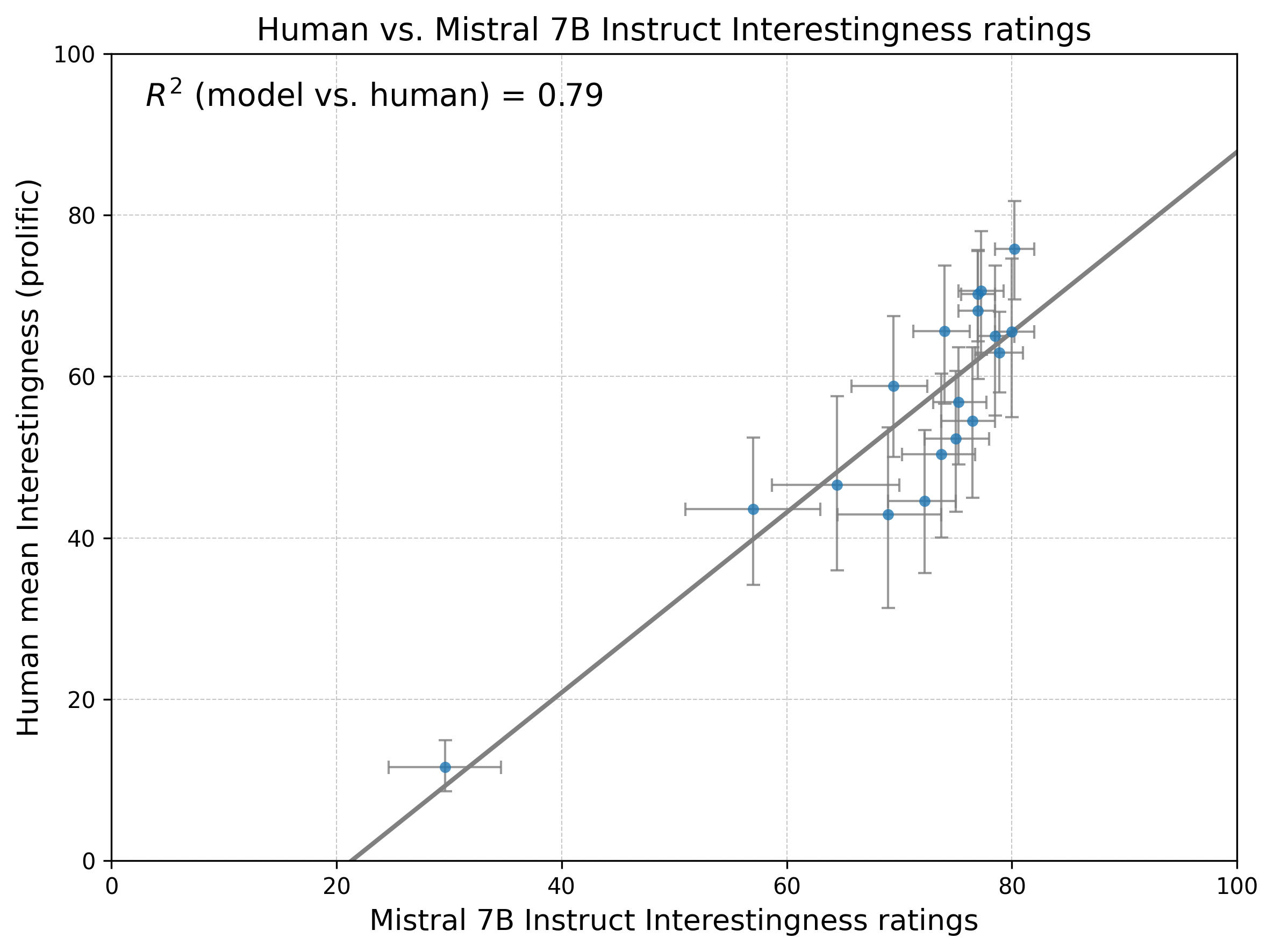} &
        \includegraphics[width=0.48\linewidth]{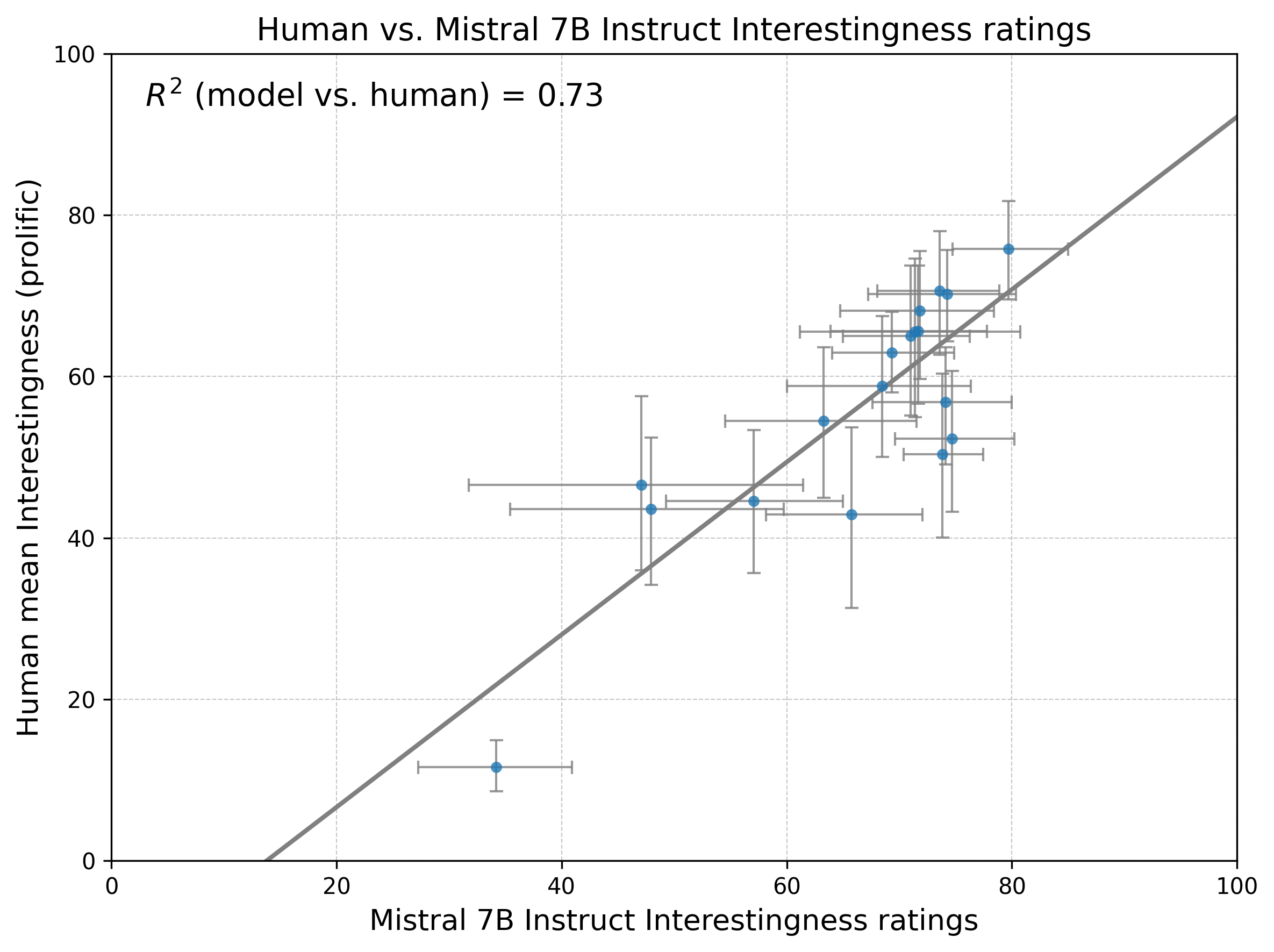} \\
        \rotatebox{90}{Difficulty} &
        \includegraphics[width=0.48\linewidth]{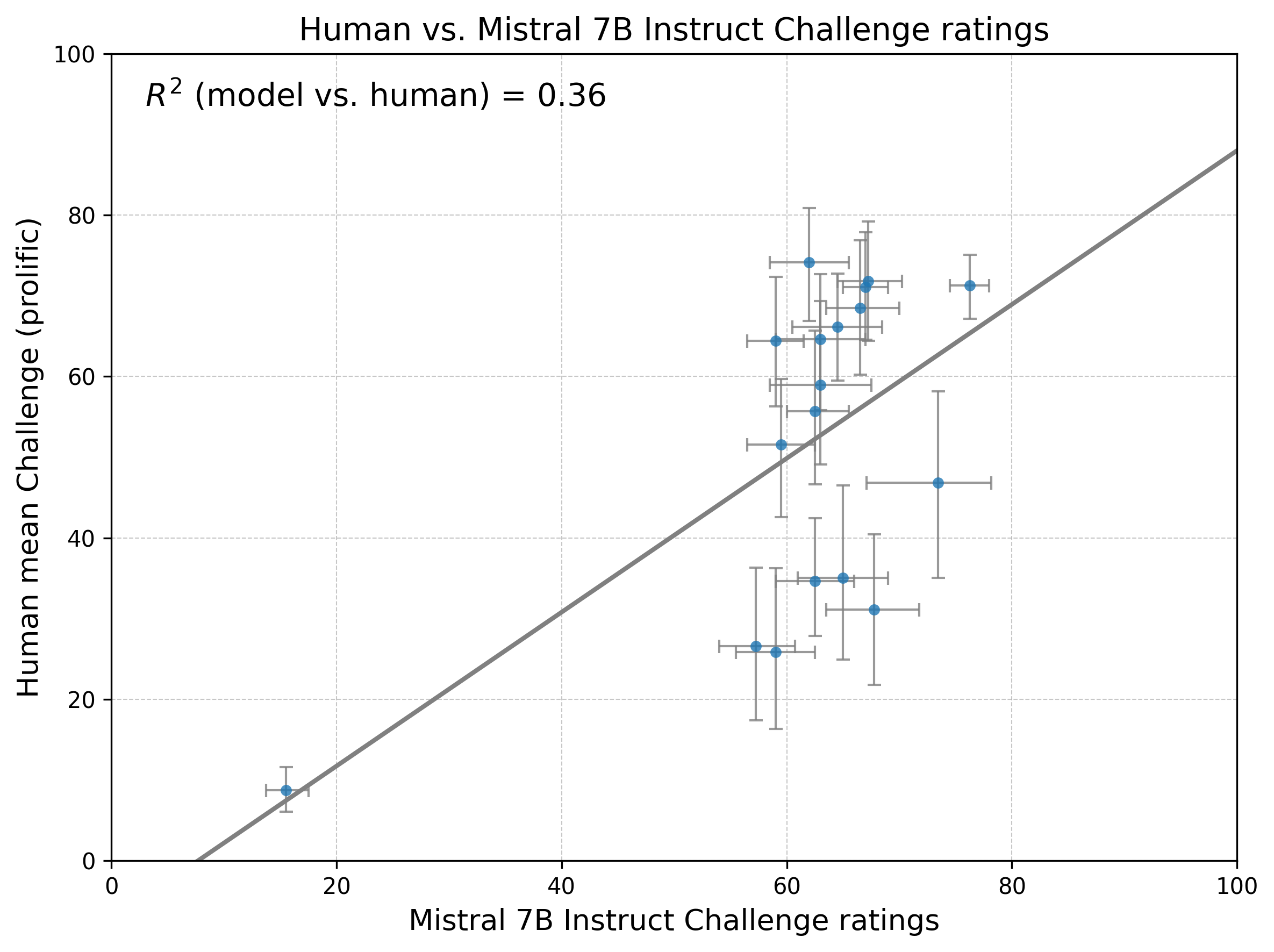} &
        \includegraphics[width=0.48\linewidth]{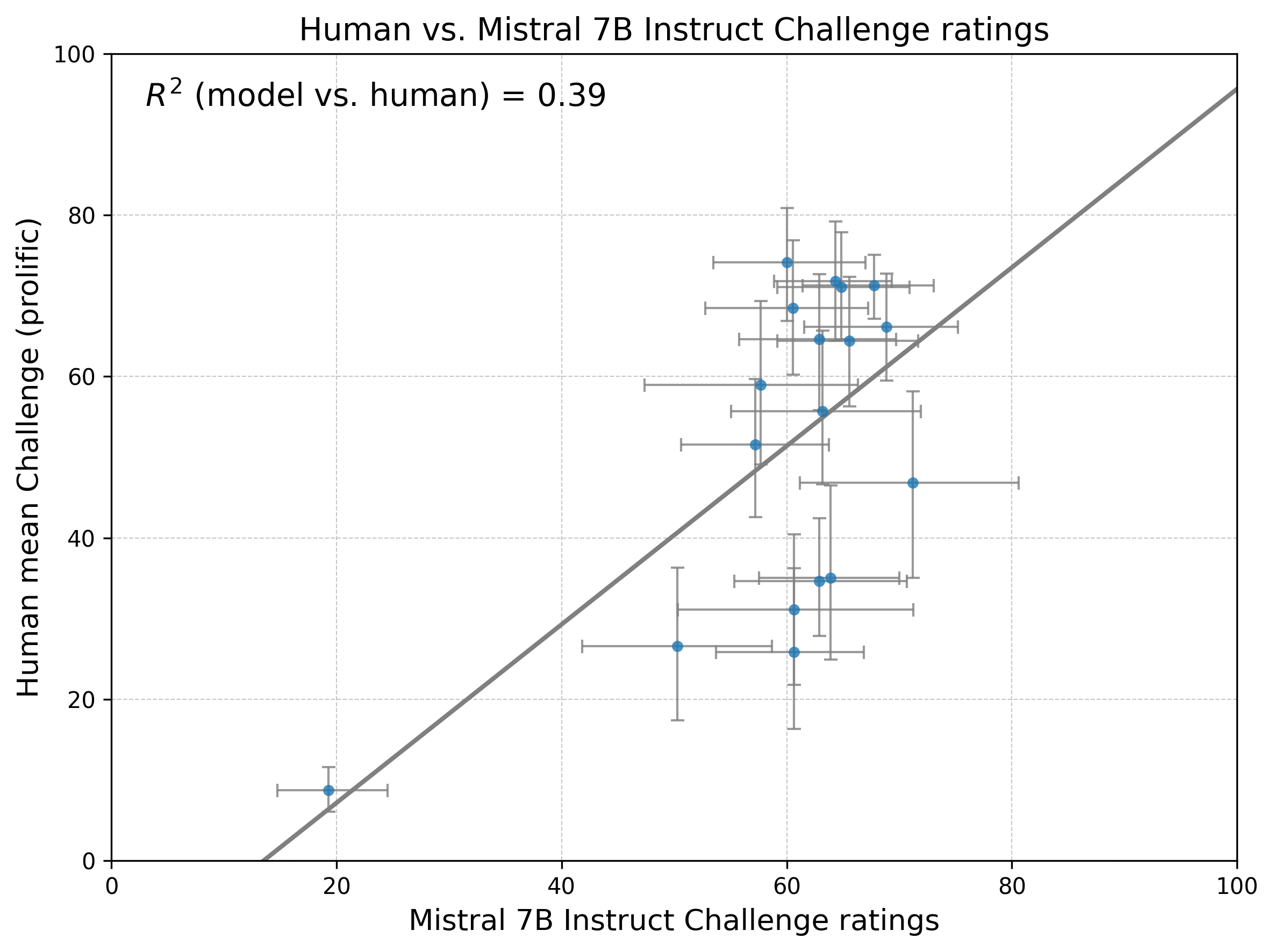} \\
    \end{tabular}
    \caption{Mistral-7B-Instruct: Human vs LLM ratings}
\end{figure}

\begin{figure}[htbp]
    \centering
    \begin{tabular}{c|cc}
        & \textbf{Temp 0.3} & \textbf{Temp 1.0} \\ \hline
        \rotatebox{90}{Interestingness} &
        \includegraphics[width=0.48\linewidth]{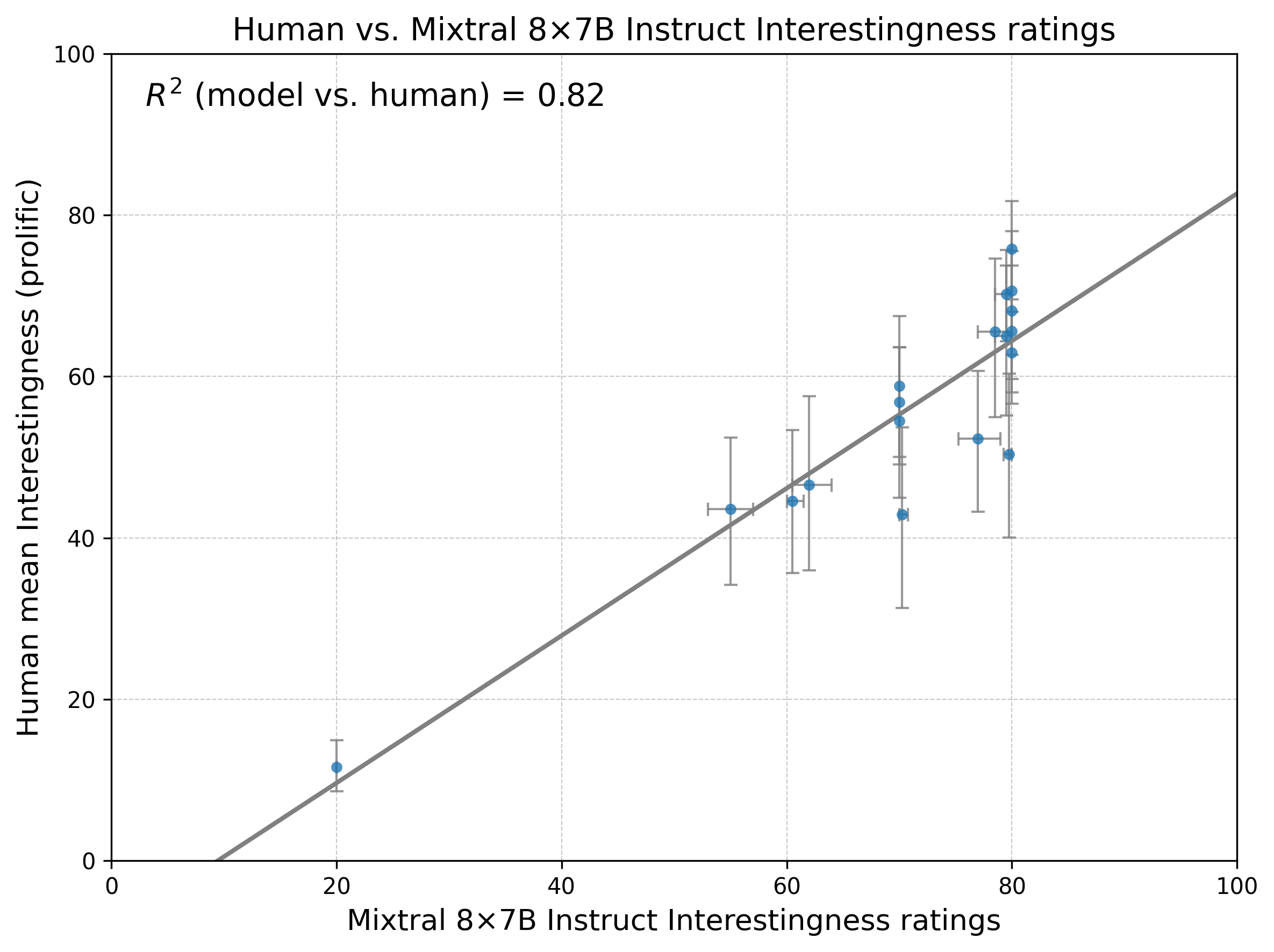} &
        \includegraphics[width=0.48\linewidth]{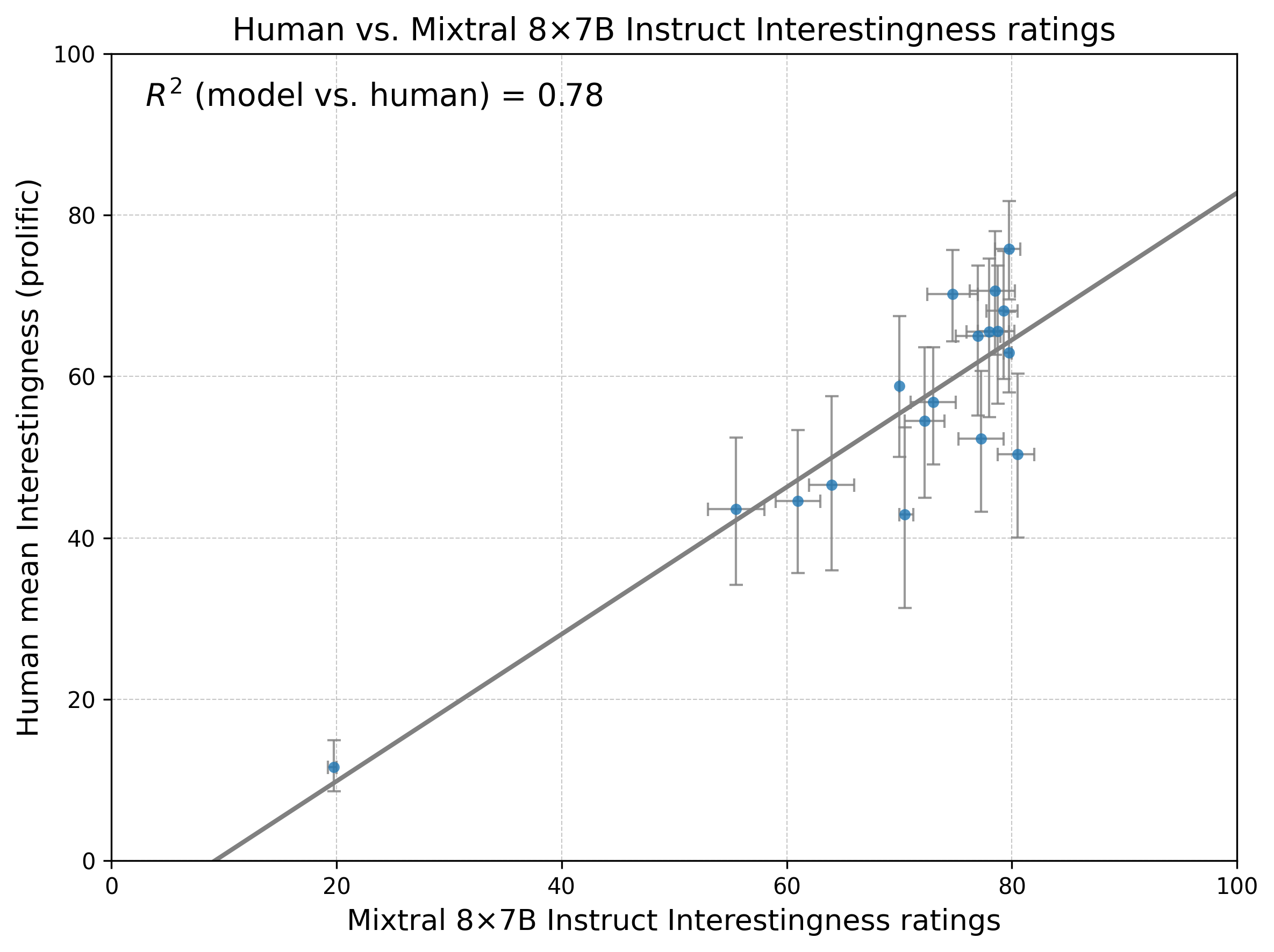} \\
        \rotatebox{90}{Difficulty} &
        \includegraphics[width=0.48\linewidth]{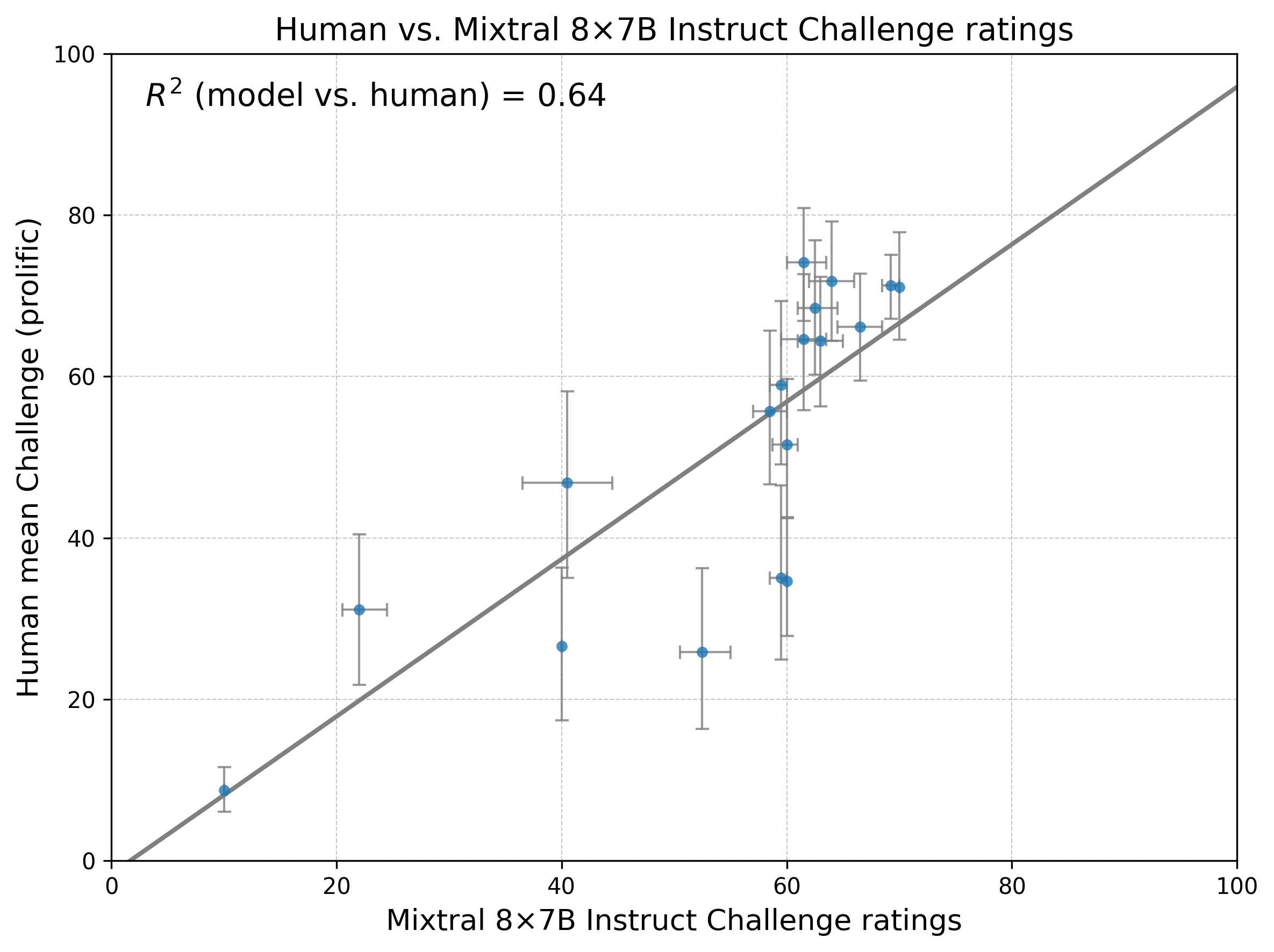} &
        \includegraphics[width=0.48\linewidth]{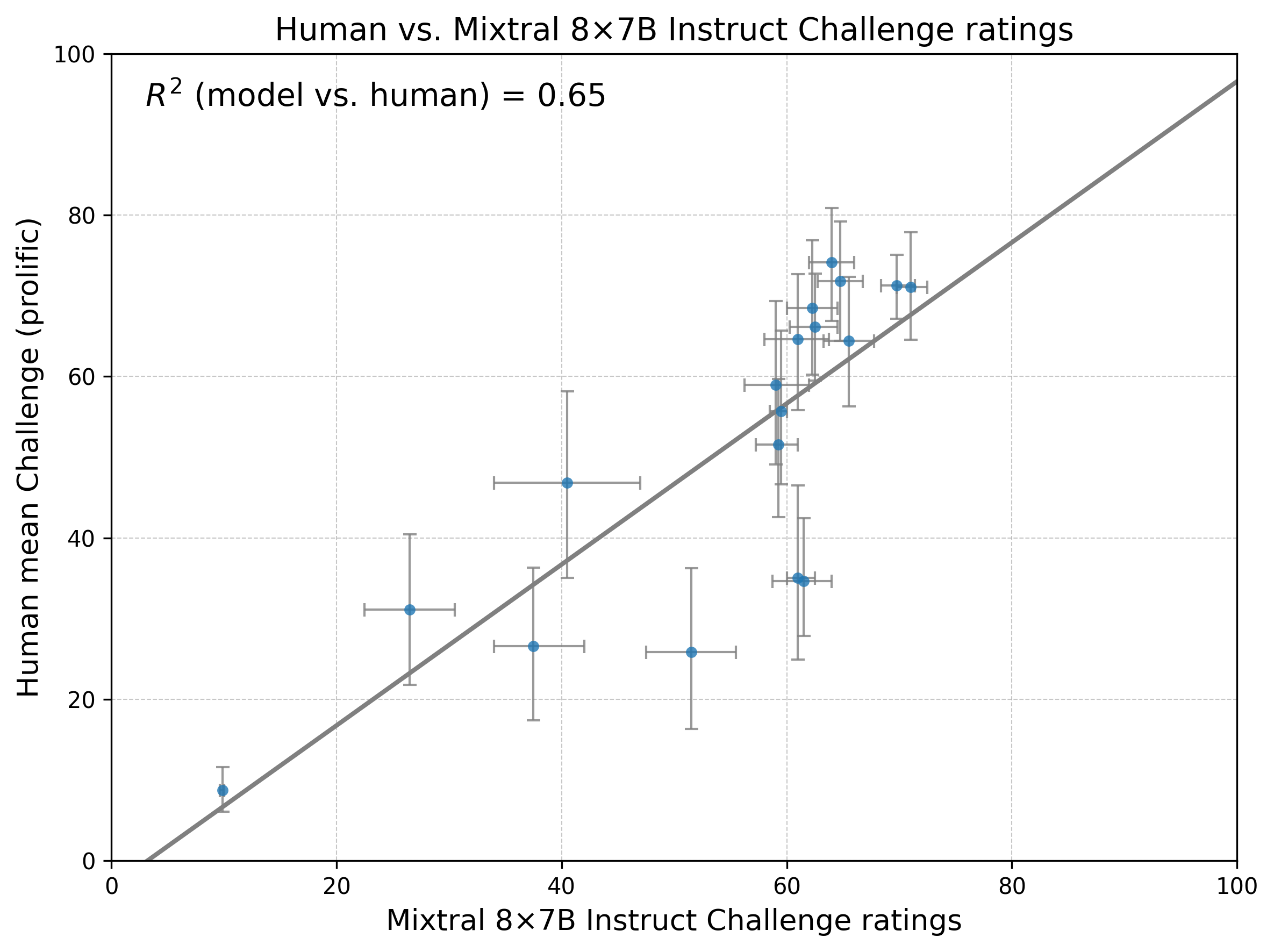} \\
    \end{tabular}
    \caption{Mixtral-8x7B-Instruct: Human vs LLM ratings}
\end{figure}

\begin{figure}[htbp]
    \centering
    \begin{tabular}{c|cc}
        & \textbf{Temp 0.3} & \textbf{Temp 1.0} \\ \hline
        \rotatebox{90}{Interestingness} &
        \includegraphics[width=0.48\linewidth]{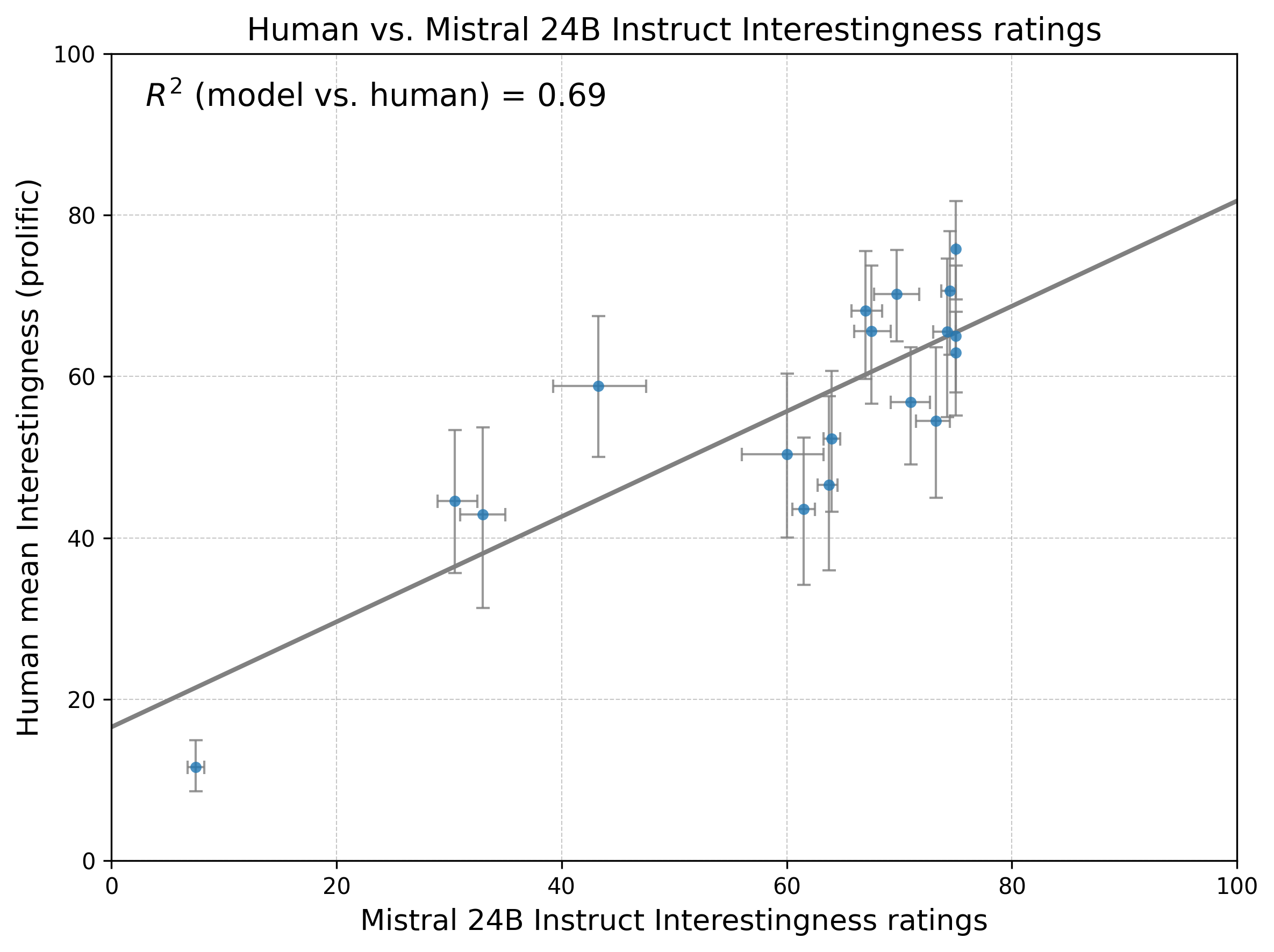} &
        \includegraphics[width=0.48\linewidth]{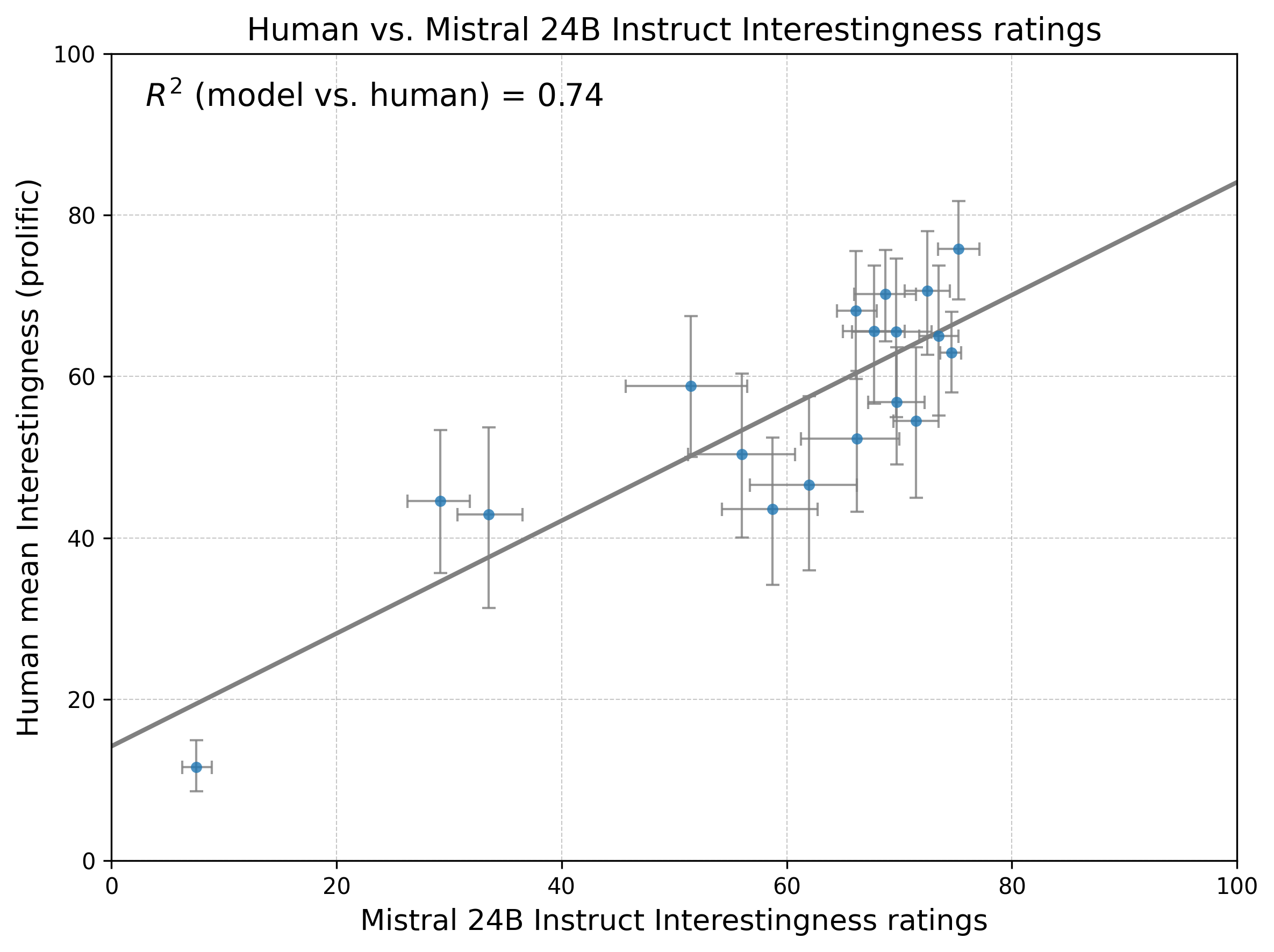} \\
        \rotatebox{90}{Difficulty} &
        \includegraphics[width=0.48\linewidth]{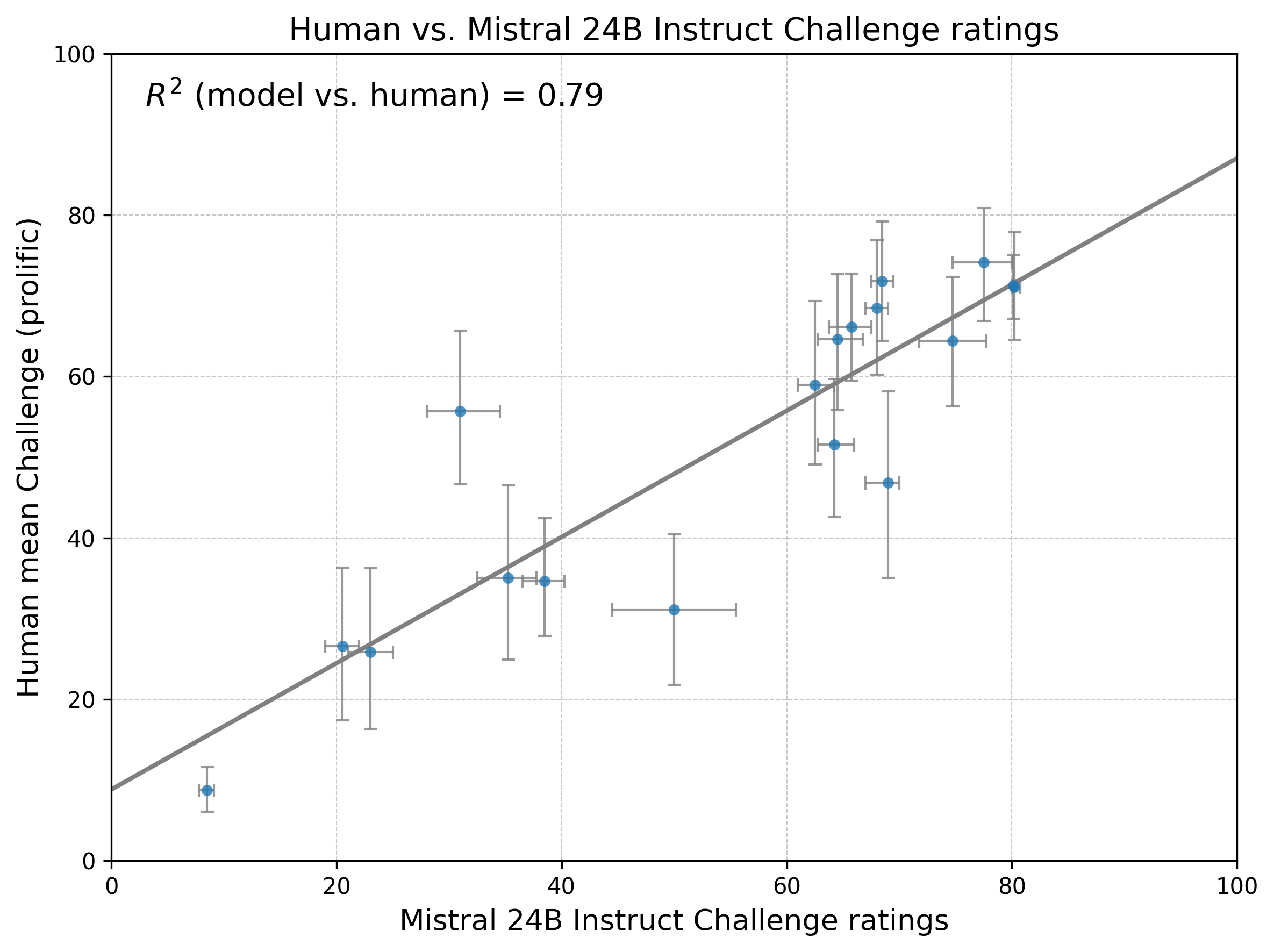} &
        \includegraphics[width=0.48\linewidth]{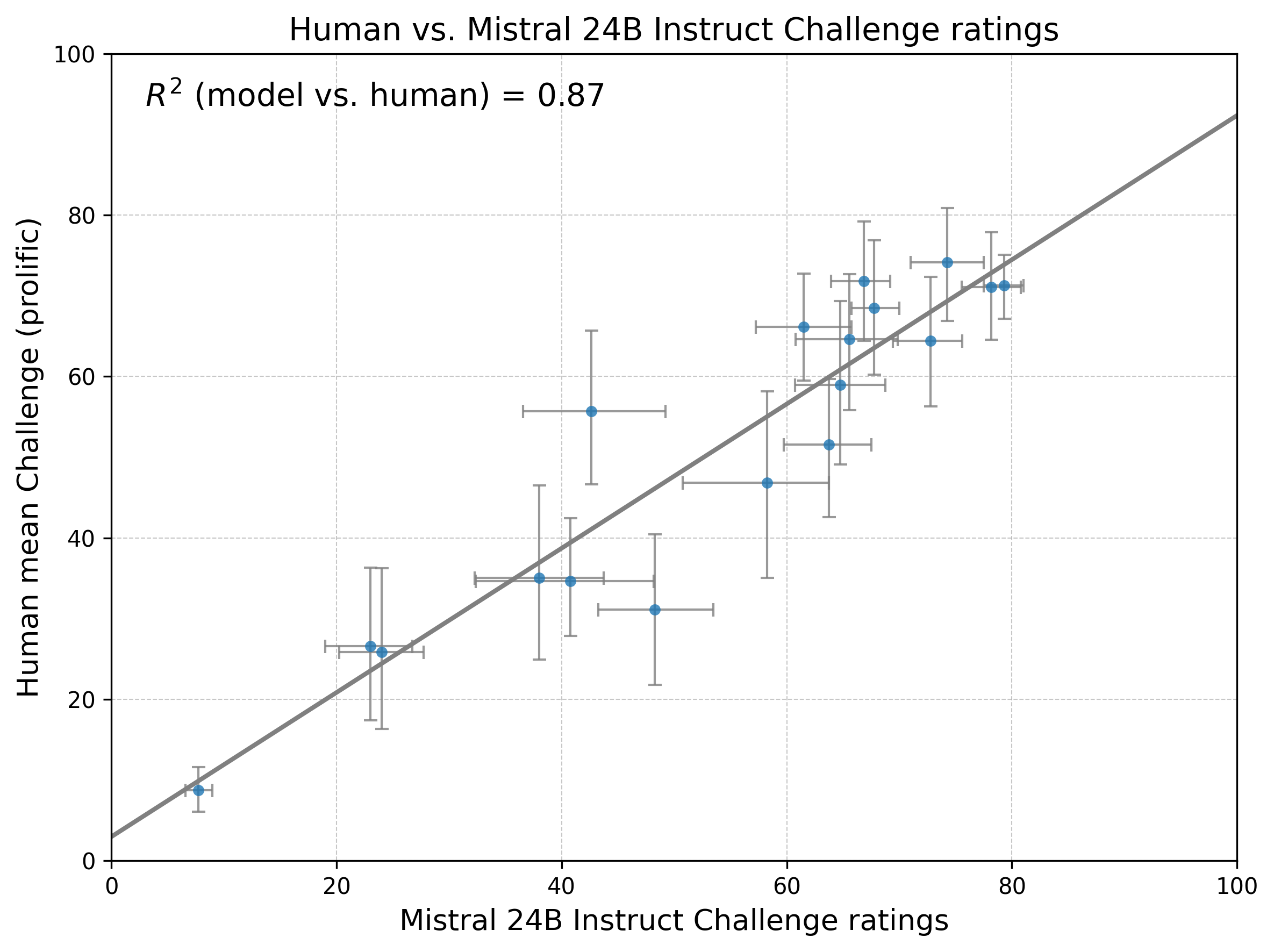} \\
    \end{tabular}
    \caption{Mistral-24B-Instruct: Human vs LLM ratings}
\end{figure}

\begin{figure}
    \centering
    \includegraphics[width=0.75\linewidth]{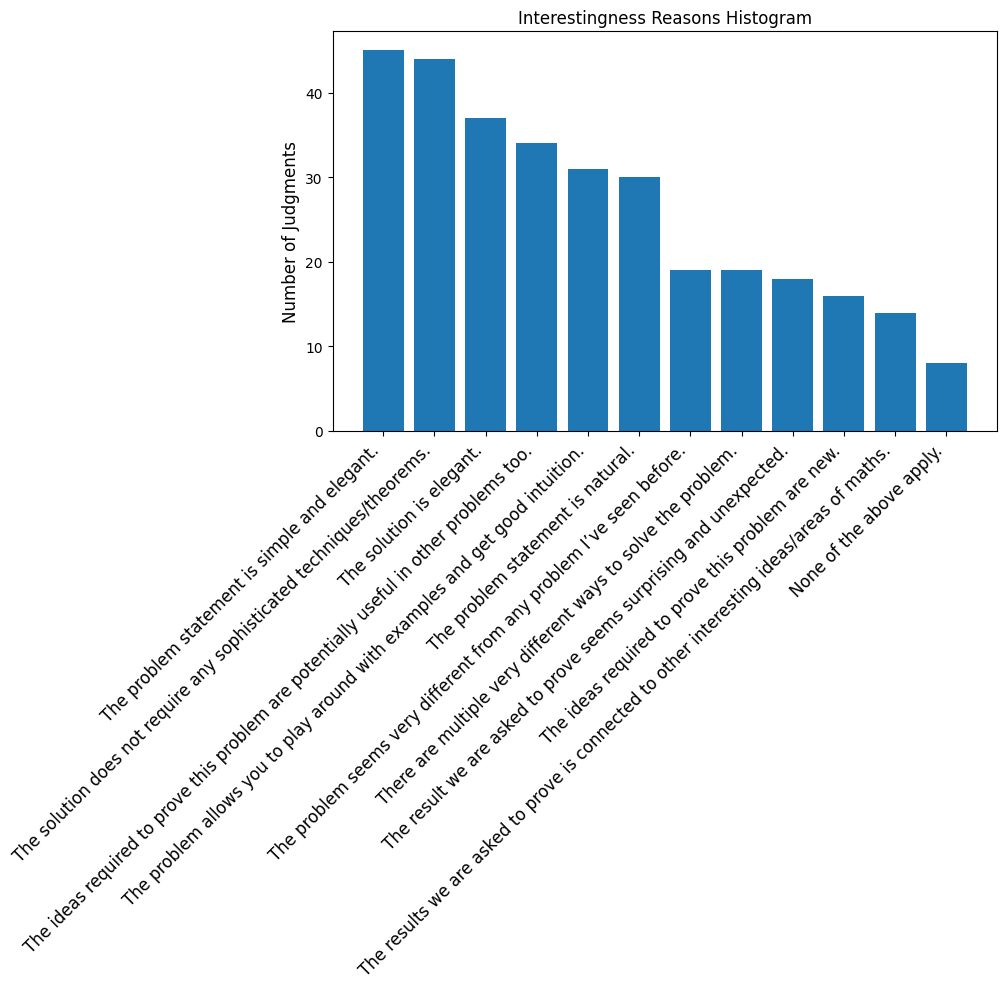}
    \includegraphics[width=0.75\linewidth]{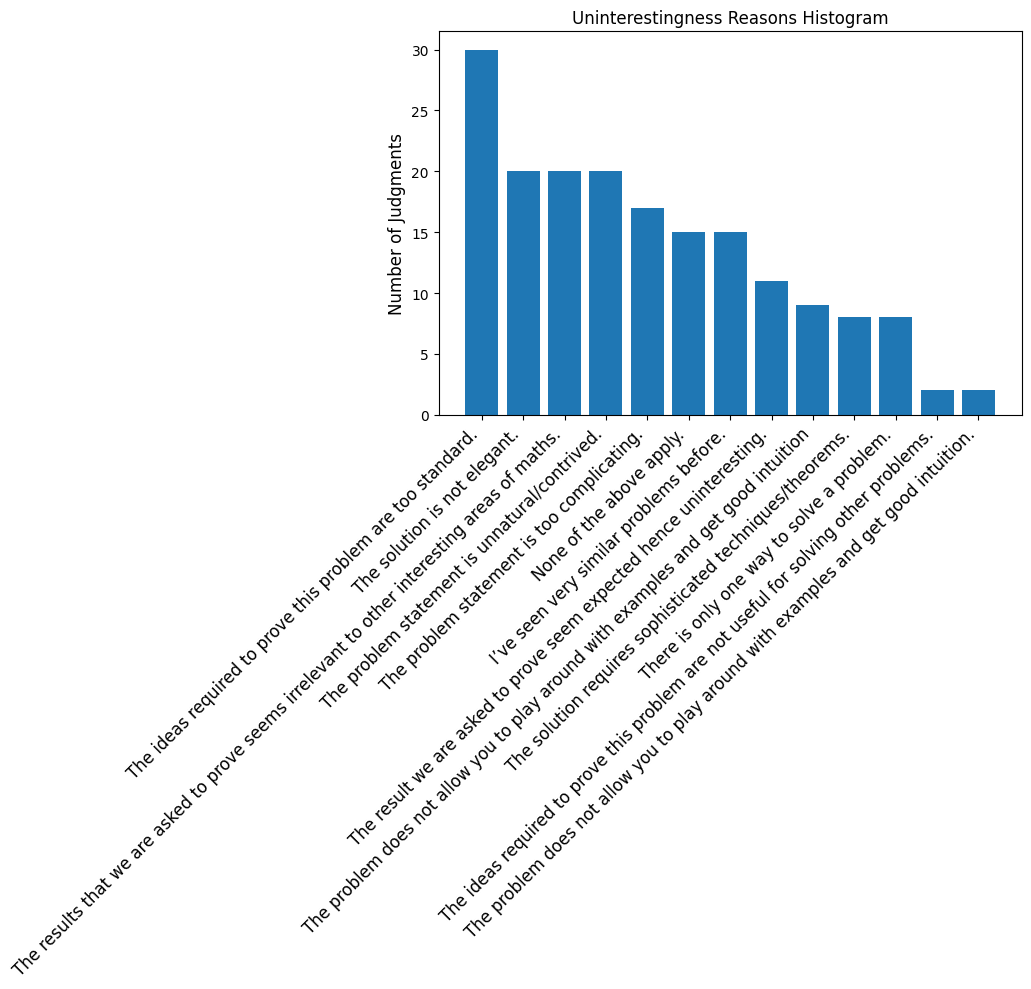}
    \caption{Histogram of how frequently interestingness (top) and uninterestingness (bottom) reasons were chosen by human participants across the survey.}
    \label{fig:hist_int_unint}
\end{figure}
\begin{figure}
    \centering
    \includegraphics[width=0.9\linewidth]{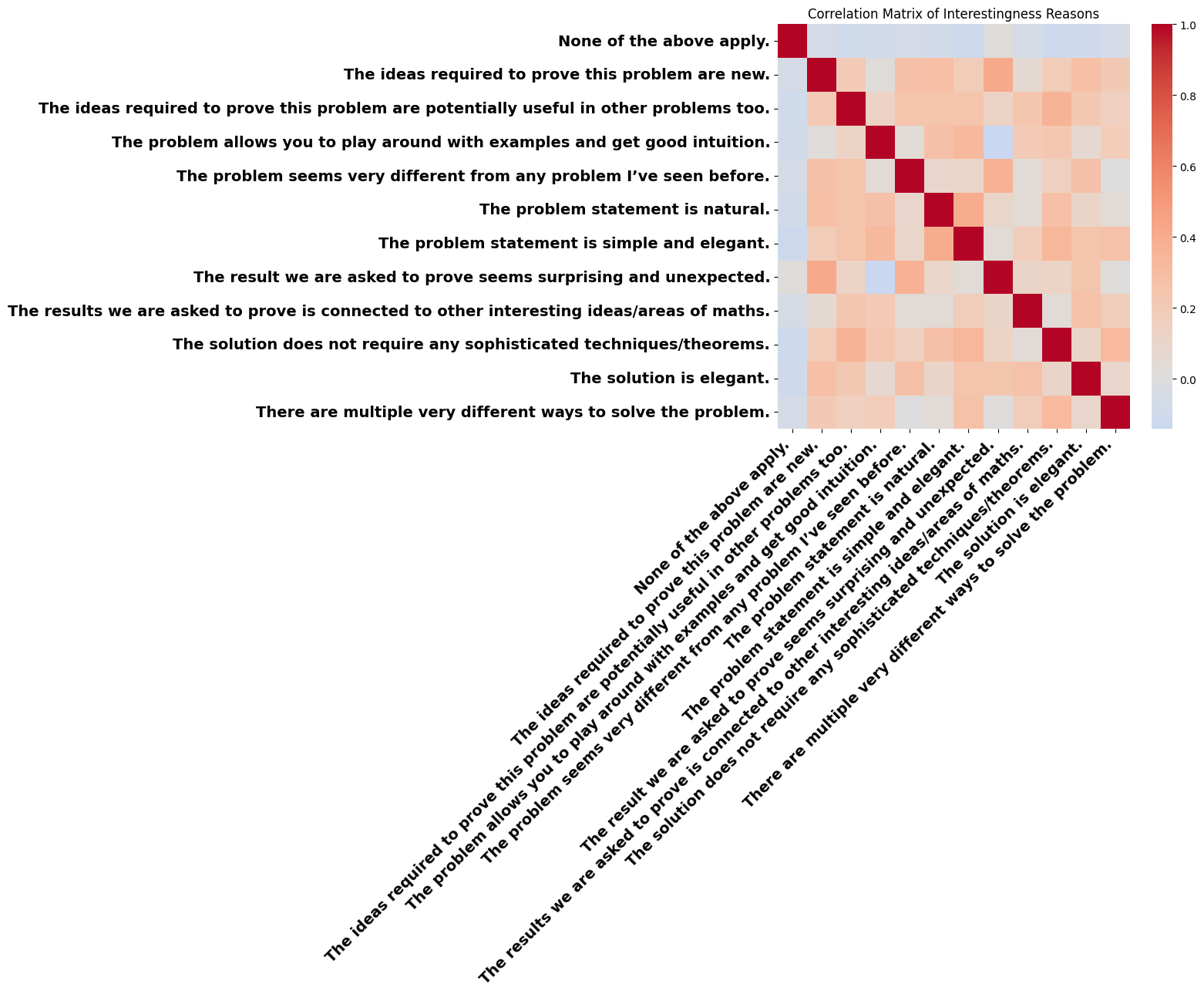}
    \includegraphics[width=0.9\linewidth]{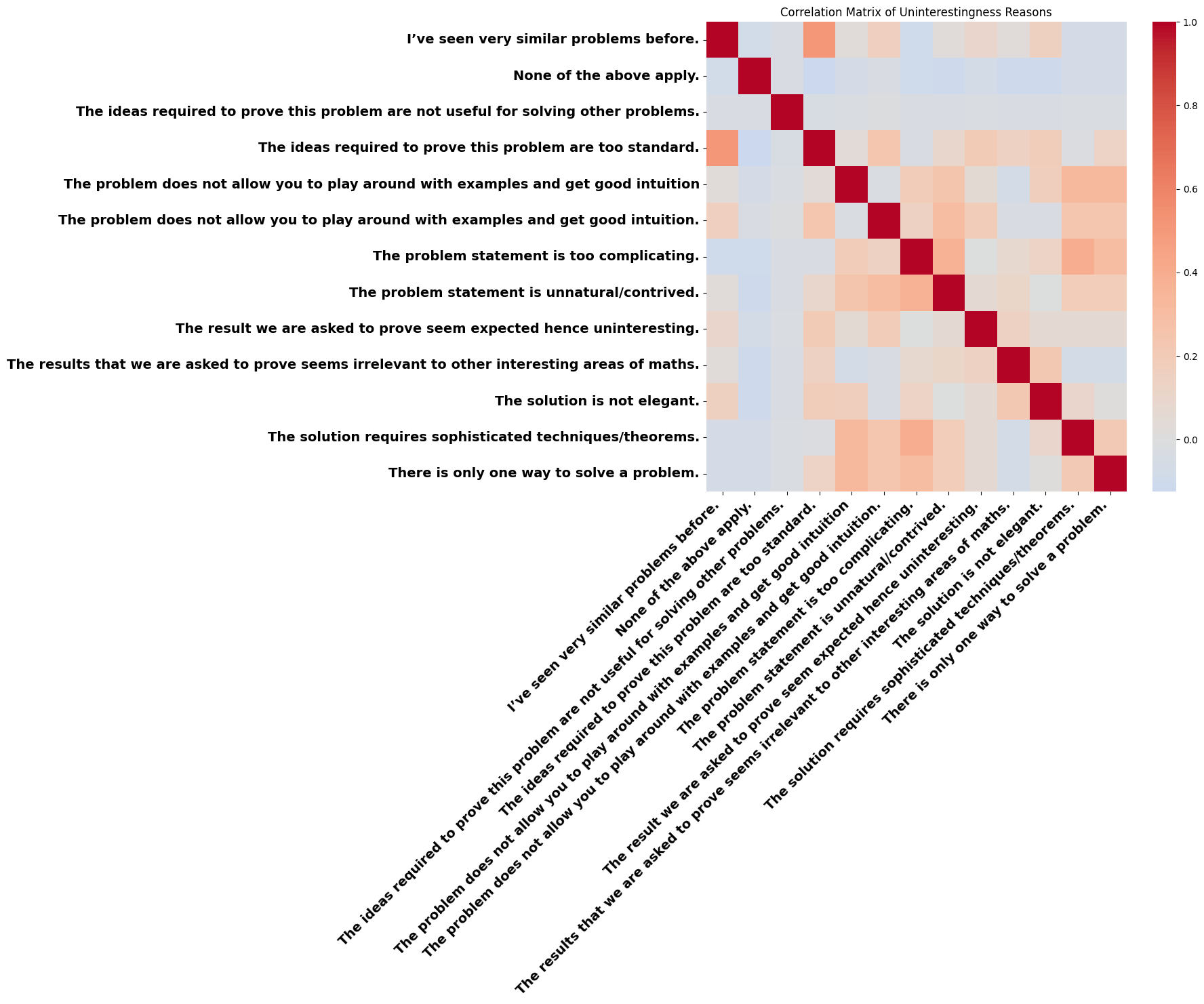}
    \caption{Correlation matrices showing when human participants in our IMO study chose multiple reasons for interestingness (top) or uninterestingness (bottom) for the same problem.}
    \label{fig:int_unint_corr}
\end{figure}

\begin{figure}[ht]
    \centering
    \includegraphics[width=\linewidth]{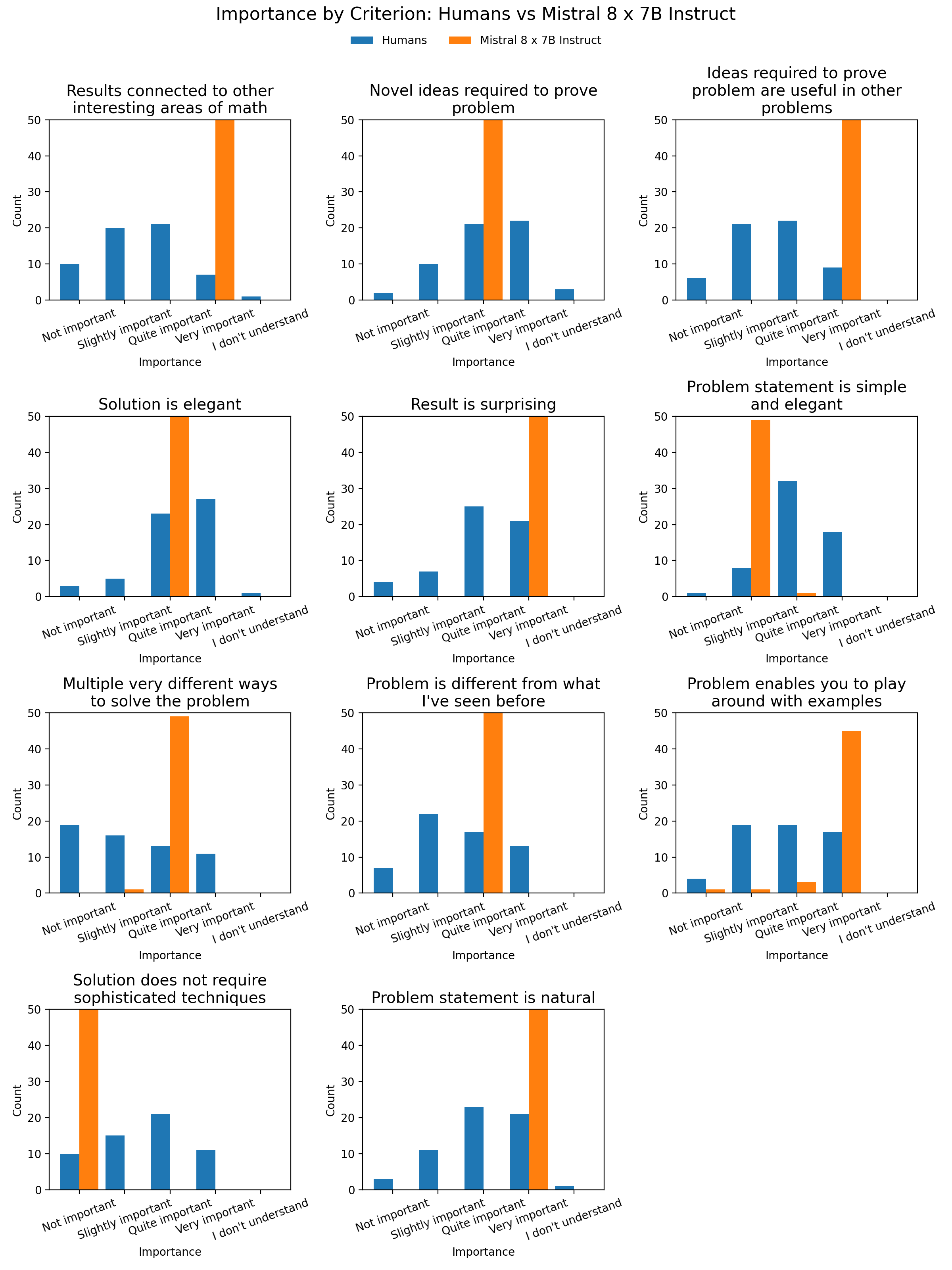}
    \caption{Importance ratings from \textbf{Mixtral 8$\times$7B Instruct}.}
\end{figure}

\begin{figure}[ht]
    \centering
    \includegraphics[width=\linewidth]{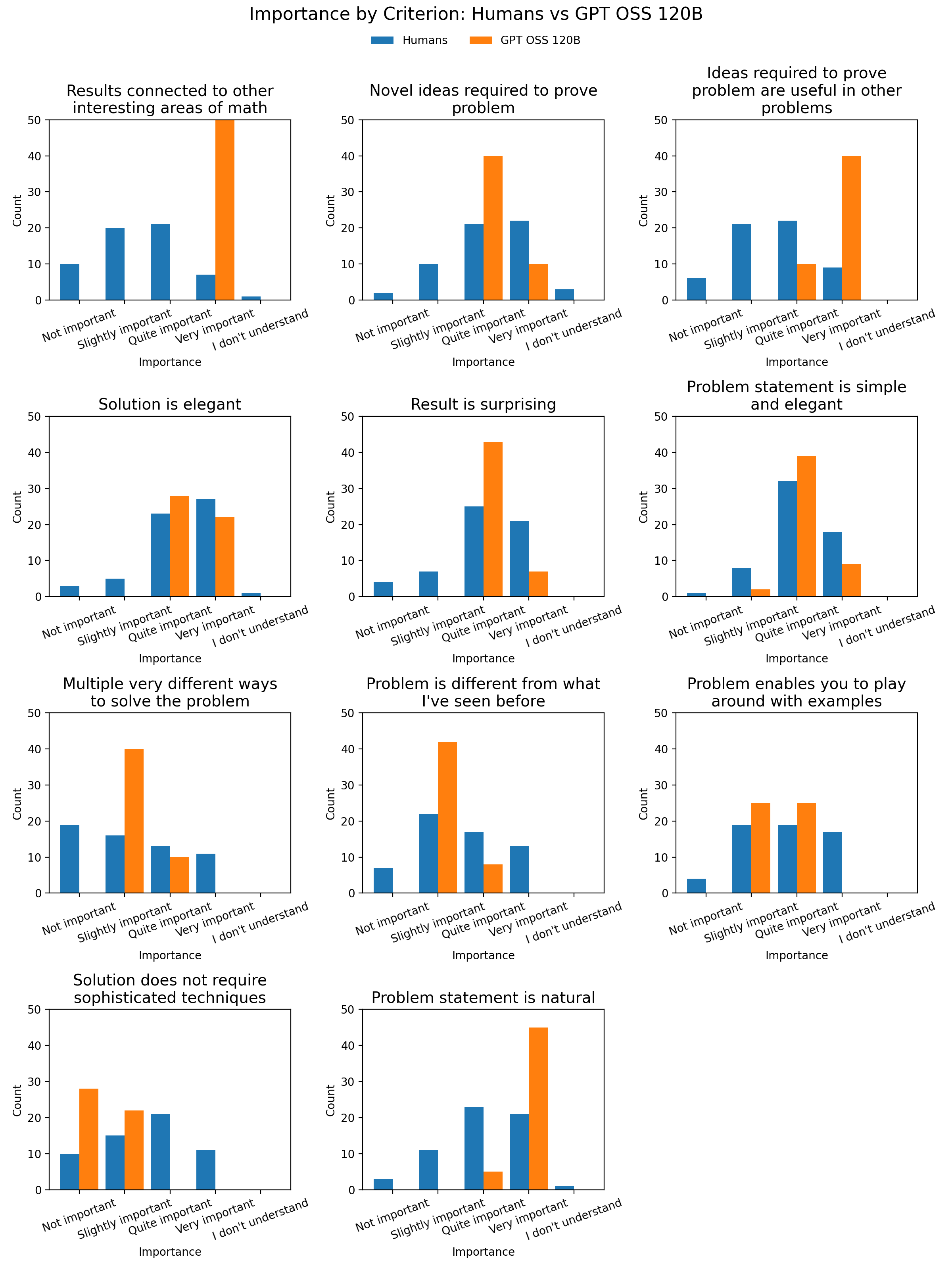}
    \caption{Importance ratings from \textbf{GPT-OSS 120B}.}
\end{figure}

\begin{figure}[ht]
    \centering
    \includegraphics[width=\linewidth]{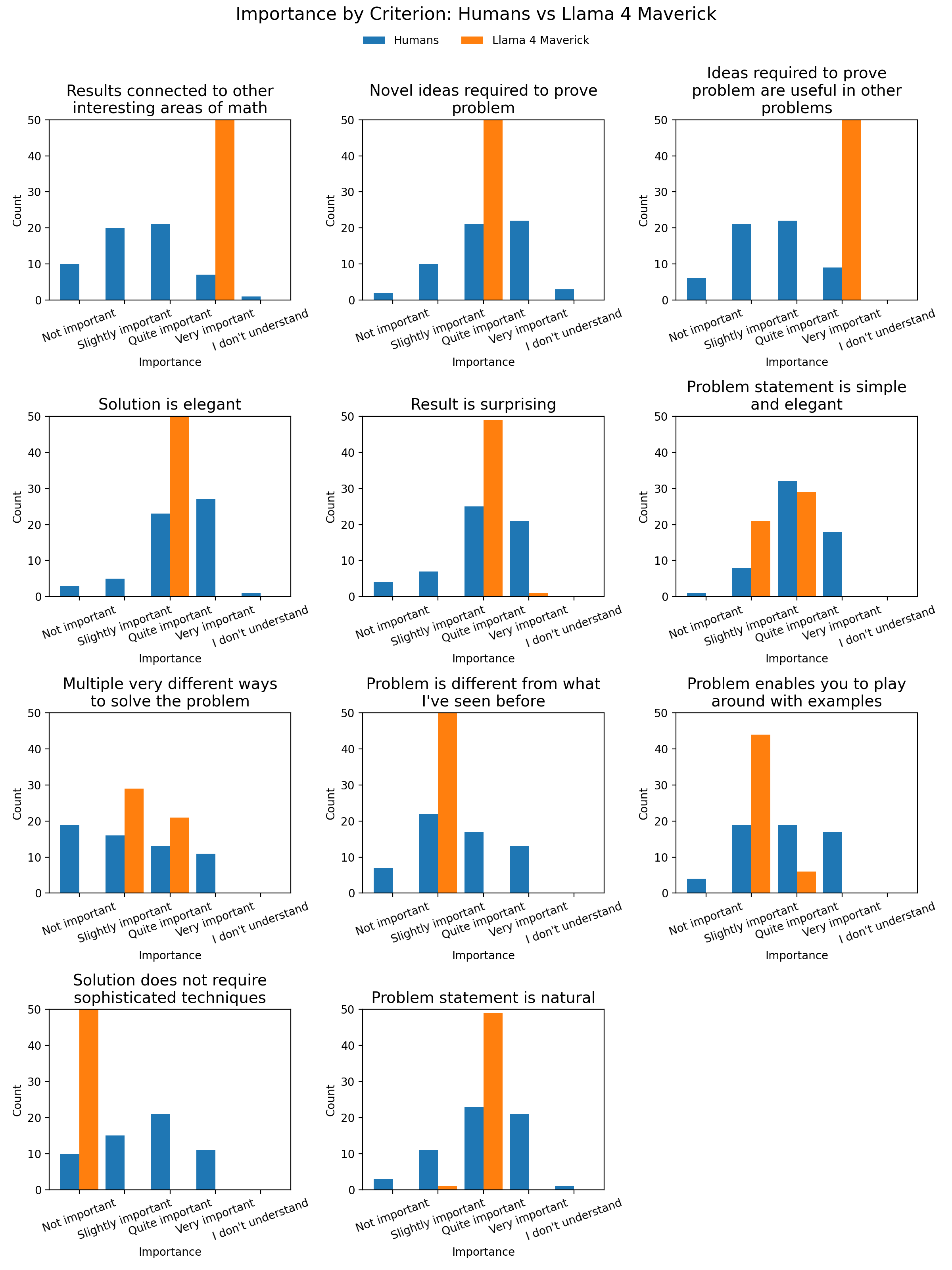}
    \caption{Importance ratings from \textbf{Llama 4 Maverick}.}
\end{figure}

\begin{figure}[ht]
    \centering
    \includegraphics[width=\linewidth]{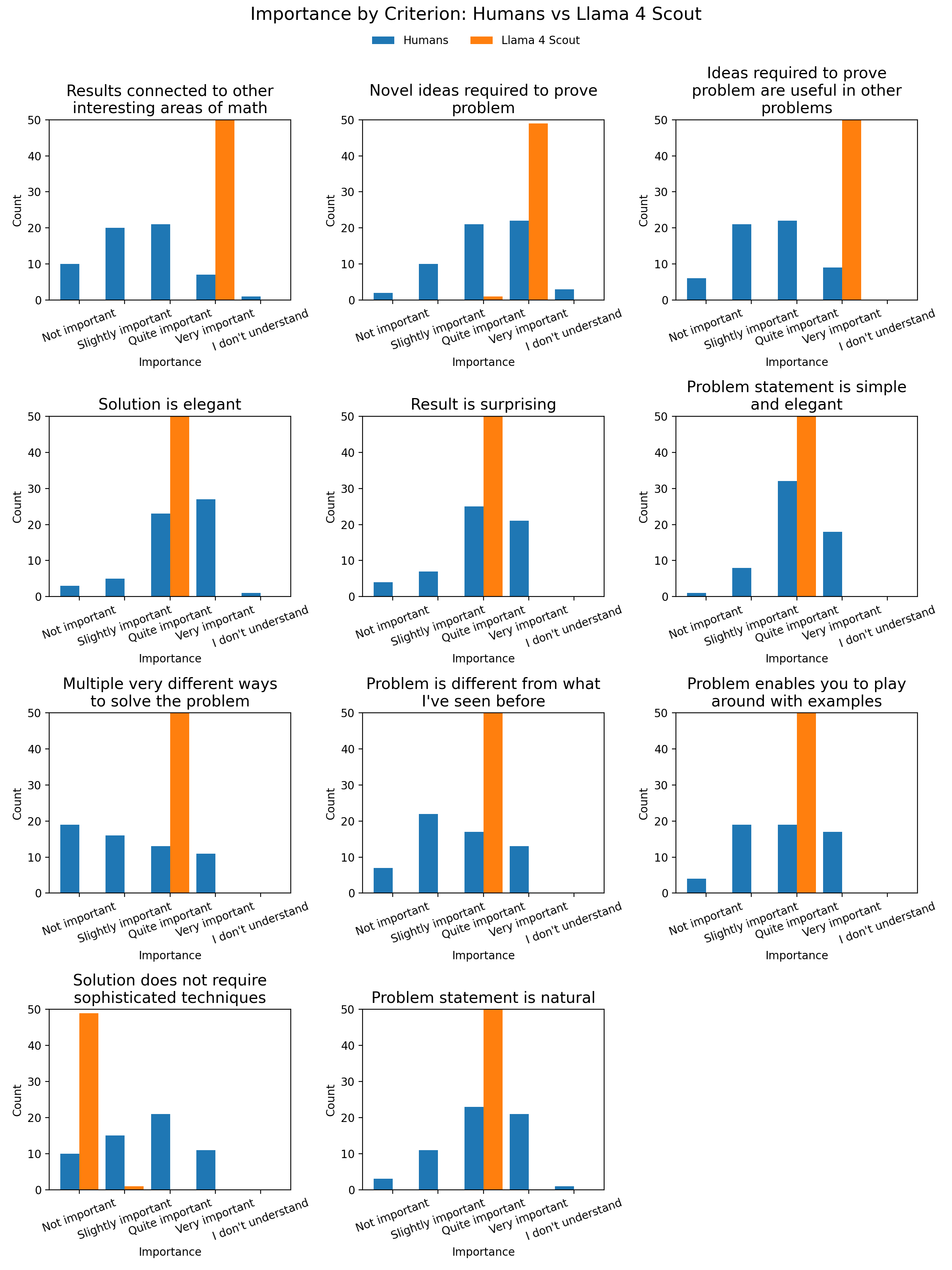}
    \caption{Importance ratings from \textbf{Llama 4 Scout}.}
\end{figure}

\begin{figure}[ht]
    \centering
    \includegraphics[width=\linewidth]{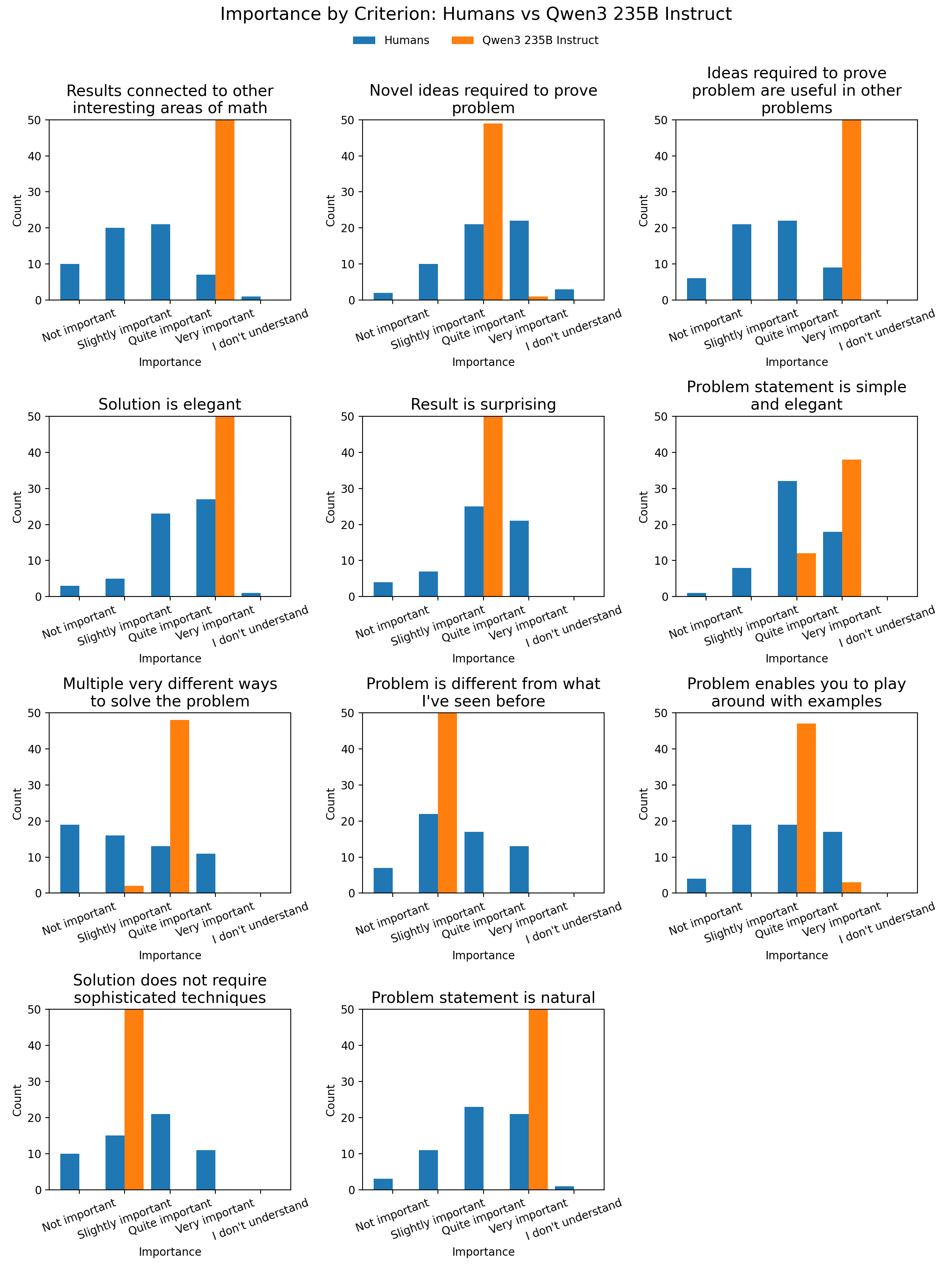}
    \caption{Importance ratings from \textbf{Qwen 235B Instruct}.}
\end{figure}

\begin{figure}[ht]
    \centering
    \includegraphics[width=\linewidth]{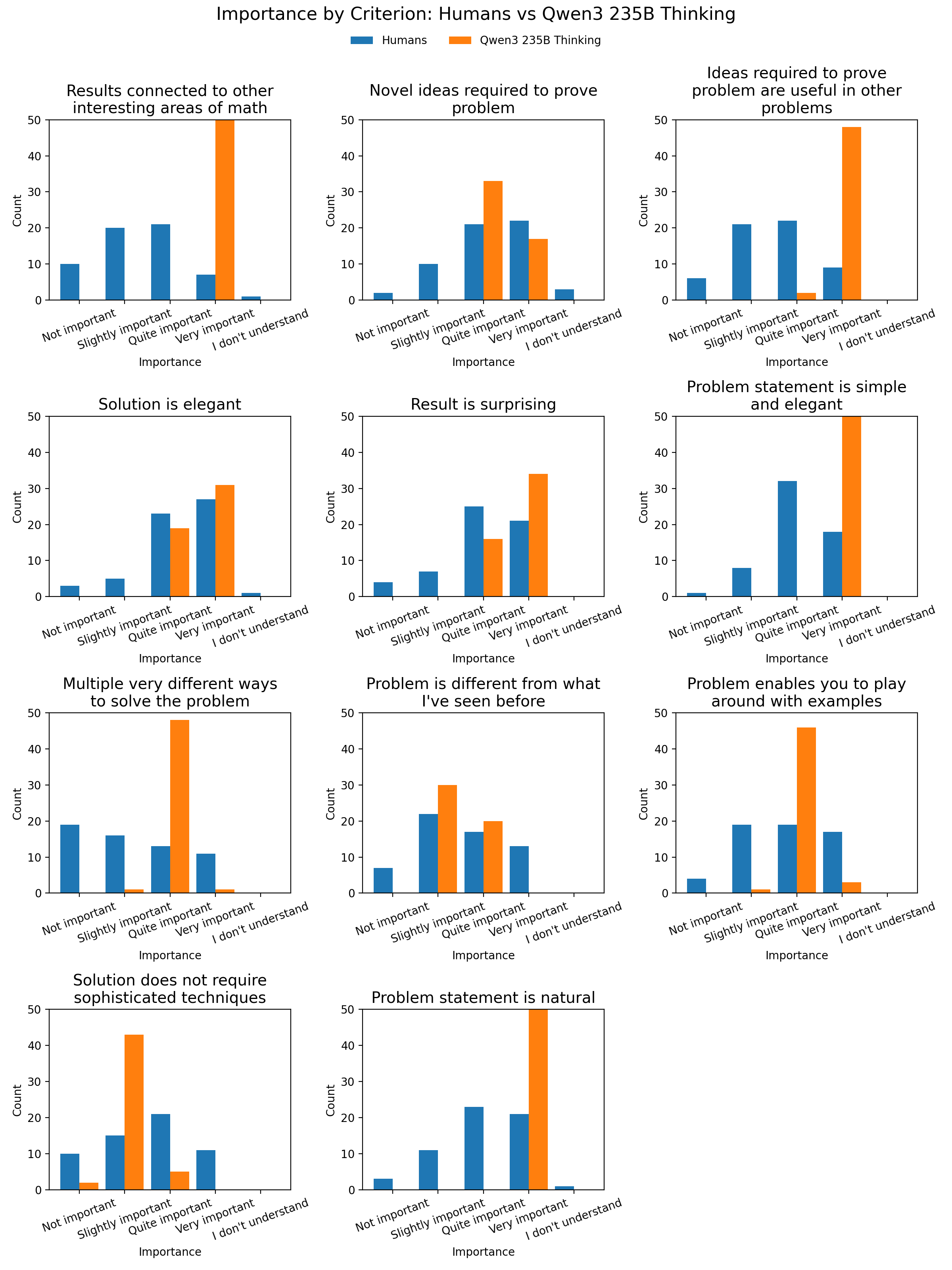}
    \caption{Importance ratings from \textbf{Qwen 235B Thinking}.}
\end{figure}

\begin{figure}[ht]
    \centering
    \includegraphics[width=\linewidth]{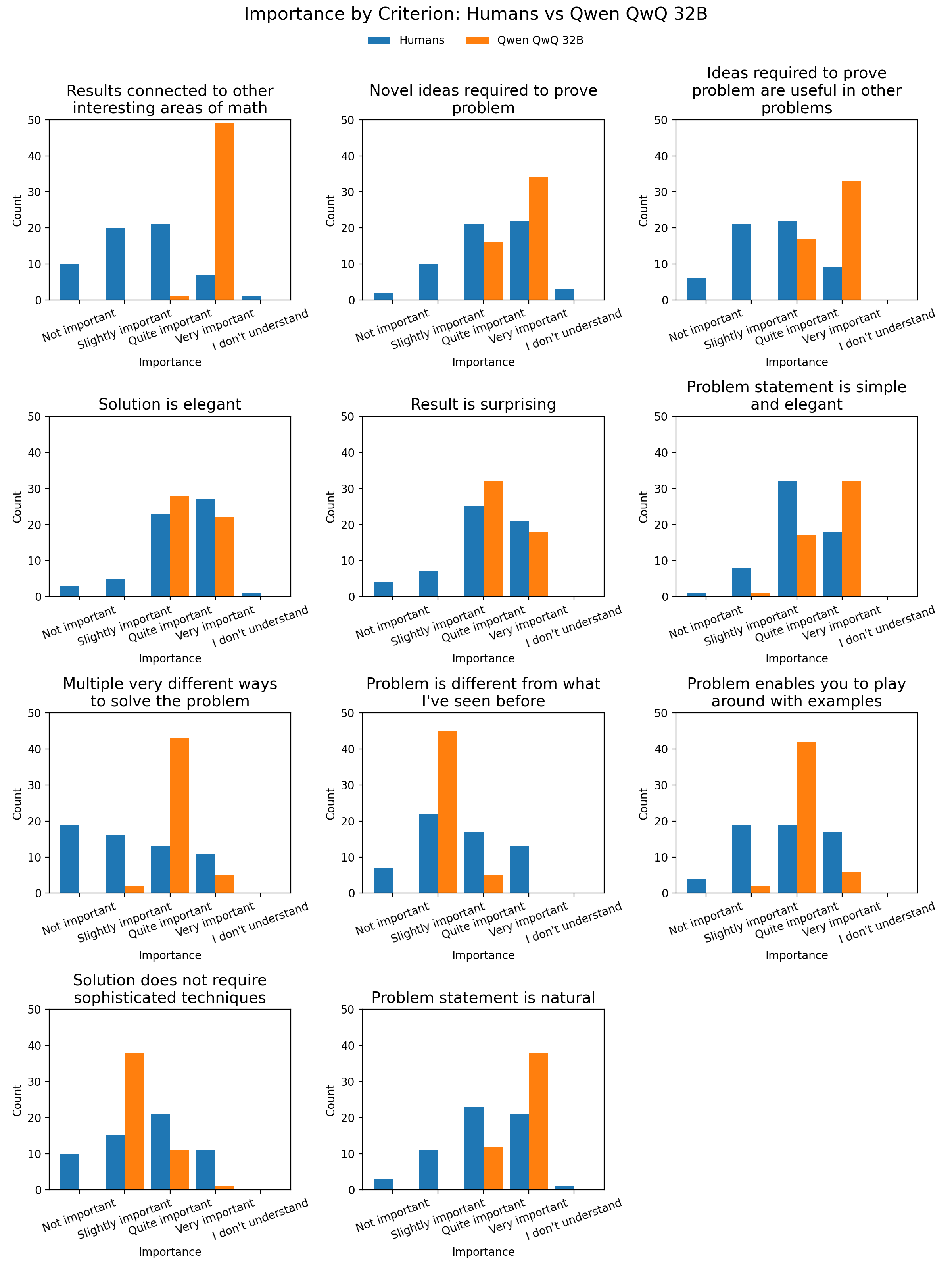}
    \caption{Importance ratings from \textbf{QwQ 32B}.}
\end{figure}

\begin{figure}[ht]
    \centering
    \includegraphics[width=\linewidth]{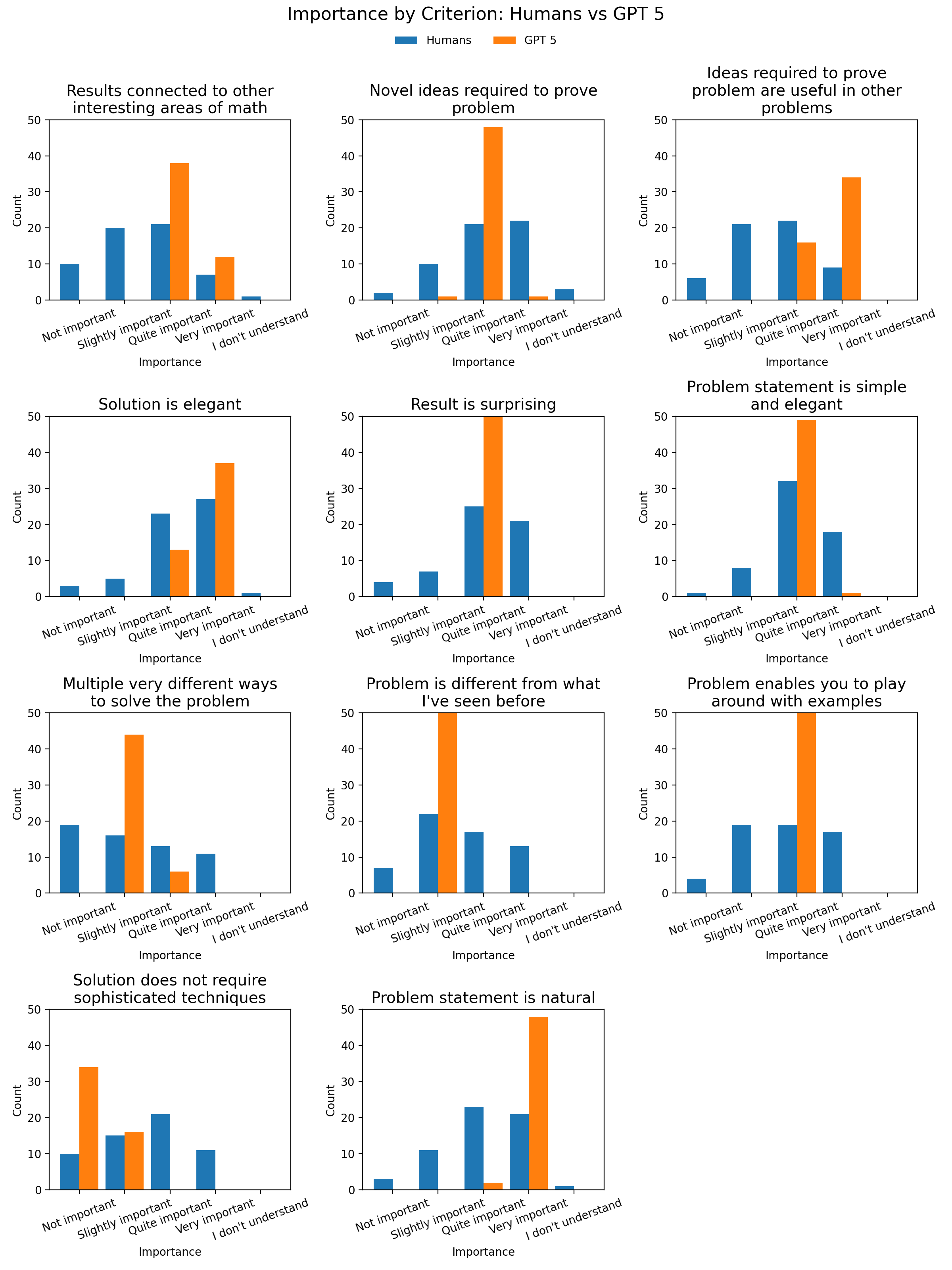}
    \caption{Importance ratings from \textbf{GPT-5}.}
\end{figure}

\begin{figure}[ht]
    \centering
    \includegraphics[width=\linewidth]{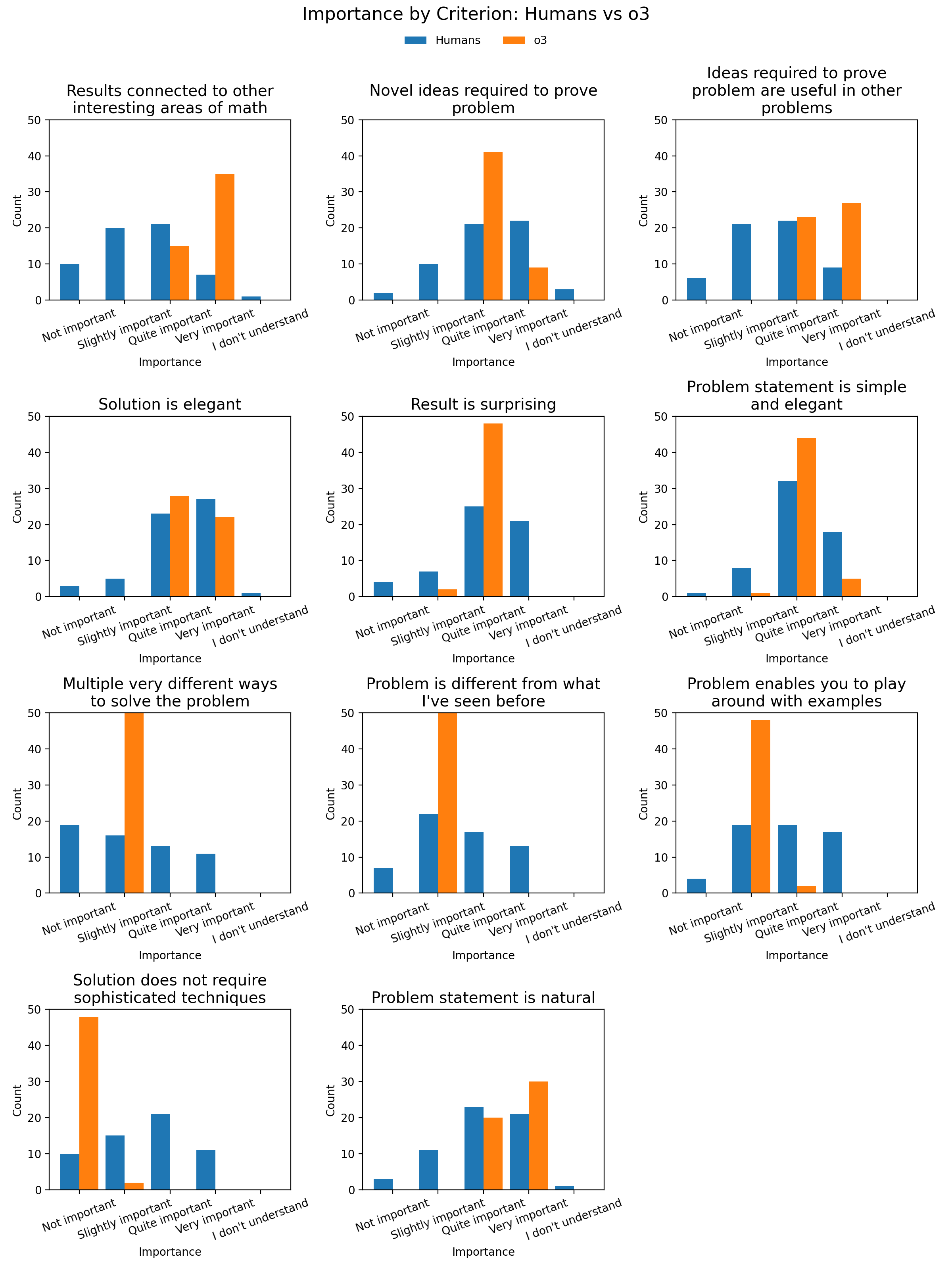}
    \caption{Importance ratings from \textbf{o3}.}
\end{figure}

\subsection{LLM Generated Problems Results}
\begin{figure}
    \centering
    \includegraphics[width=1\linewidth]{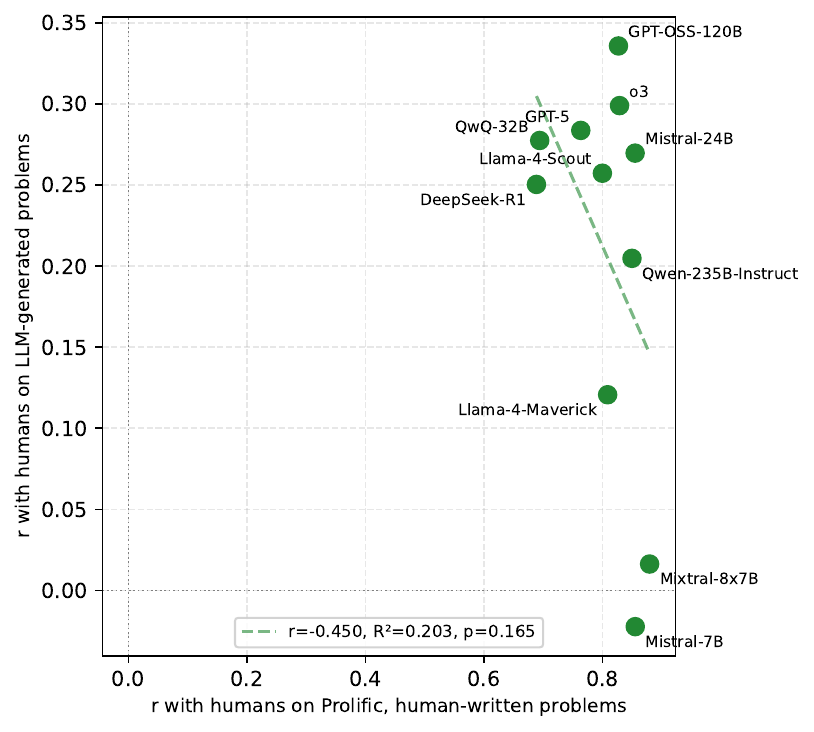}
    \caption{Comparing LLM-human alignment on interestingness judgments for human-written (Prolific) problems to LLM-human alignment on LLM-generated problems (i.e. are the models that are most aligned with humans on human-written problems also the most aligned with humans on LLM-generated problems?). We see no significant relationship between whether a model is more human-aligned on judging human-written problems and whether it is aligned on judging LLM-generated ones.}
    \label{fig:alignment_llm}
\end{figure}

In Figure~\ref{fig:alignment_llm}, we plot the correlation between LLMs' performance on aligning with humans on the interestingness of human-written problems vs. on the interestingness of those generated by LLMs. In~\cref{tab:top_bottom_generated_problems}, we include the top- and bottom-3 generated problems for the two batches of problems we run.
\begin{table*}[t]
\centering
\small
\setlength{\tabcolsep}{4pt}
\renewcommand{\arraystretch}{1.08}
\caption{Top- and bottom-rated generated problems by batch, according to human interestingness ratings. Scores are means with bootstrapped 95\% confidence intervals.}
\label{tab:top_bottom_generated_problems}
\begin{tabularx}{\textwidth}{cXlcc}
\toprule
Rank & Problem & Generator & Mean & 95\% CI \\
\midrule

\multicolumn{5}{l}{\textbf{Batch 3: Top-rated problems}} \\
\midrule
1 &
Find the sum of all integer solutions to the equation $(x^2 - 5x + 5)^{(x^2 - 11x + 30)} = 1$. &
Qwen3-235B & 77.8 & [68.1, 86.9] \\

2 &
Suppose $d = -2w + 17$, $2d - 6w - 6 = 0$. Suppose $-4i = 2u + 2$, $-10 = -i + u$. What is the remainder when $d$ is divided by $i$? &
Mistral-7B & 75.3 & [61.3, 87.0] \\

3 &
How many 5-digit positive integers, whose five digits are all different and whose digits add up to 10, are divisible by 11? &
o3 & 70.7 & [54.1, 84.9] \\

\midrule
\multicolumn{5}{l}{\textbf{Batch 3: Bottom-rated problems}} \\
\midrule
10 &
Let $a_n$ be the $n$th positive integer in increasing order whose digits sum to 10. Find $a_{15}$. &
Qwen3-235B & 43.5 & [29.6, 57.5] \\

9 &
An arithmetic sequence of positive integers has first term 3 and common difference $d$. Knowing that both the first term and the 10th term are divisors of 720, find the sum of all possible positive integers $d$. &
o3 & 55.5 & [40.0, 70.1] \\

8 &
Find the number of ordered pairs of positive integers $(x,y)$ such that $\frac{1}{x} + \frac{1}{y} = \frac{1}{2023}$. &
Qwen3-235B & 56.3 & [42.5, 70.5] \\

\midrule
\multicolumn{5}{l}{\textbf{Batch 4: Top-rated problems}} \\
\midrule
1 &
A 4-digit number is called balanced if the sum of its digits in odd positions equals the sum of its digits in even positions. How many balanced 4-digit numbers exist? &
Qwen3-235B & 82.6 & [74.1, 89.7] \\

2 &
During one complete 12-hour cycle, a digital clock displays every minute from 12:00 through 11:59. Removing the colon forms an integer, e.g., 7:05 becomes 705. How many of the 720 resulting integers are divisible by 3? &
o3 & 78.3 & [65.8, 88.3] \\

3 &
Find the positive integer $k$ such that the equation $\left|4x^2 - 20x\right| = k$ has exactly three distinct real solutions. &
Qwen3-235B & 64.2 & [50.1, 77.2] \\

\midrule
\multicolumn{5}{l}{\textbf{Batch 4: Bottom-rated problems}} \\
\midrule
10 &
Let $n(y) = 12y - 109$. What are the prime factors of $n(-8)$? &
Mistral-7B & 49.5 & [35.6, 64.7] \\

9 &
How many integers $n$ between 1 and 2023, inclusive, make $n^3 + 2023$ divisible by $n + 2023$? &
o3 & 52.0 & [39.1, 65.9] \\

8 &
A sequence is defined such that the first term is 1, and each subsequent term is the smallest positive integer not already in the sequence for which the sum of the current term and the previous term is a perfect square. What is the fifth term of this sequence? &
Qwen3-235B & 53.3 & [37.9, 67.4] \\

\bottomrule
\end{tabularx}
\end{table*}

\clearpage
\section{Models}
\label{app:models}
In~\cref{tab:models}, we list all models and the providers we use.
\begin{table}[ht]
\centering
\caption{Models evaluated in this study, their full identifiers, and providers.}
\begin{tabular}{l p{6.5cm} l}
\hline
\textbf{Model} & \textbf{Full Name} & \textbf{Provider} \\
\hline
llama\_scout & meta-llama/Llama-4-Scout-17B-16E-Instruct & Meta (TogetherAI) \\
llama\_maverick & meta-llama/Llama-4-Maverick-17B-128E-Instruct-FP8 & Meta (TogetherAI) \\
deepseek\_r1 & deepseek-ai/DeepSeek-R1 & DeepSeek (TogetherAI) \\
o3 & o3 & OpenAI \\
gpt\_5 & gpt-5 & OpenAI \\
gpt\_oss\_120b & openai/gpt-oss-120b & OpenAI (TogetherAI) \\
mixtral\_8x7b\_instruct & mistralai/Mixtral-8x7B-Instruct-v0.1 & Mistral (TogetherAI) \\
mistral\_7b\_instruct & mistralai/Mistral-7B-Instruct-v0.1 & Mistral (TogetherAI) \\
mistral\_24b\_instruct & mistralai/Mistral-Small-24B-Instruct-2501 & Mistral (TogetherAI) \\
qwen\_235b\_instruct & Qwen/Qwen3-235B-A22B-Instruct-2507-tput & Qwen (TogetherAI) \\
qwen\_235b\_thinking & Qwen/Qwen3-235B-A22B-Thinking-2507 & Qwen (TogetherAI) \\
qwq\_32b & Qwen/QwQ-32B & Qwen (TogetherAI) \\
\hline
\end{tabular}
\label{tab:models}
\end{table}

\section{Compute Information}
\label{app:compute}
Our analyses do not require GPU compute, and all LLM inference was done using TogetherAI and OpenAI's API. The device used to run the analyses relies on the Apple M3 Pro chip.

\section{Prolific Study}
\label{app:prolific}
Below, we describe additional information used to collect and adapt problems for the Prolific study. The majority of the problems were taken from the AMC contests, which are property of the Mathematical Association of America. 
\subsection{Variant types}
\label{app:prolific-variant-types}
\begin{itemize}
    \item \textbf{Increasing/decreasing numbers:} The numerical values in the problem are scaled up or down while keeping the structure intact.
    \item \textbf{Adding/removing steps:} The problem is modified to include additional intermediate steps, or simplified by removing steps.
    \item \textbf{Adding ambiguity:} The wording is adjusted to introduce multiple plausible interpretations.
\end{itemize}

\clearpage
\subsection{List of problems}
We list all problems for our Prolific study, their variants, and variant types in Table \ref{tab:prolific_probs}.
\label{app:problems}
\begin{table}[ht]
\centering
\small
\setlength{\tabcolsep}{6pt}
\caption{Prolific study problems and their variants.}
\begin{tabular}{p{0.42\linewidth} p{0.42\linewidth} p{0.12\linewidth}}
\toprule
\textbf{Original Problem} & \textbf{Variant Problem} & \textbf{Variant Type} \\
\midrule
 Real numbers $x$ and $y$ satisfy the following system:  $x^2 + y^2 = 25$ \\ $(x + y + 5)(-x + y + 5)(x - y + 5)(x + y - 5) = 100$  and $x + y = \sqrt(m)$. Determine $m$. &  & Positive control (no variant) \\
\midrule
 What is $28 + 13$? &  & Negative control (no variant) \\
\midrule
 What is the value of $9901 \cdot 101-99 \cdot 10101?$ & What is the value of $999001 \cdot 1001-999 \cdot 1001001?$ & Increase value \\
\midrule
 The number $2024$ is written as the sum of not necessarily distinct two-digit numbers. What is the least number of two-digit numbers needed to write this sum? & The number $500$ is written as the sum of not necessarily distinct two-digit numbers. What is the least number of two-digit numbers needed to write this sum? & Decrease value \\
\midrule
 A data set containing $20$ numbers, some of which are $6$, has mean $45$. When all the 6s are removed, the data set has mean $66$. How many 6s were in the original data set? & A data set containing $20$ numbers, 7 of which are $6$, has mean $45$. When all the 6s are removed, what is the mean of the dataset? & Remove step \\
\midrule
 Kei draws a $6$-by-$6$ grid. He colors $13$ of the unit squares silver and the remaining squares gold. Kei then folds the grid in half vertically, forming pairs of overlapping unit squares. Let $m$ and $M$ equal the least and greatest possible number of gold-on-gold pairs, respectively. What is the value of $m+M$? & Kei draws a $6$-by-$6$ grid. He colors $13$ of the unit squares silver and the remaining squares gold. Kei then folds the grid in half vertically, forming pairs of overlapping unit squares. Let $m$ equal the least possible number of gold-on-gold pairs. What is the value of $m$? &  Remove step \\
\midrule
 In a long line of people arranged left to right, the 1013th person from the left is also the 1010th person from the right. How many people are in the line? & In a long line of people, the 1013th person from one end is also the 1010th person from the other end. How many people are in the line? & Add ambiguity \\
\midrule
 Makayla finds all the possible ways to draw a path in a $5 \times 5$ square-shaped grid. Each path starts at the bottom left of the grid and ends at the top right, always moving one unit east or north. She computes the area of the region between each path and the right side of the grid. What is the sum of the areas determined by all possible paths? & Makayla finds all the possible ways to draw a path in a $2 \times 2$ square-shaped grid. Each path starts at the bottom left of the grid and ends at the top right, always moving one unit east or north. She computes the area of the region between each path and the right side of the grid. What is the sum of the areas determined by all possible paths? & Decrease value \\
\midrule
 Lucius is counting backward by $7$s. His first three numbers are $100$, $93$, and $86$. What is his $10$th number? & Lucius is counting backward by $7$s. His first three numbers are $100$, $93$, and $86$. What is his $5$th number? & Decrease value \\
\midrule
 $WXYZ$ is a rectangle with $WX=4$ and $WZ=8$. Point $M$ lies $\overline{XY}$, point $A$ lies on $\overline{YZ}$, and $\angle WMA$ is a right angle. The areas of $\triangle WXM$ and $\triangle WAZ$ are equal. What is the area of $\triangle WMA$? & $WXYZ$ is a rectangle with $WX=4$ and $WZ=8$. Point $M$ lies $\overline{XY}$, point $A$ lies on $\overline{YZ}$, and $\angle WMA$ is a right angle. The areas of $\triangle WXM$ and $\triangle WAZ$ are equal. What is the sum of the areas of $\triangle WMA$ and $\triangle WAZ$? & Add step \\
\midrule
\end{tabular}
\label{tab:prolific_probs}
\end{table}

\section{IMO Study}
\label{app:imo}
We conducted a survey of interestingness judgments made by participants at the 2024 IMO. Each of the 48 survey participants saw four problems. Each participant saw the same baseline problem. The rest of the problems were selected randomly from IMO shortlists, with each participant survey being unique and including problems from the same area (Algebra, Combinatorics, Number Theory, and Geometry). Participants were given the option to view the solution before rating the problem's \emph{interestingness} and \emph{difficulty}. They were also asked to select reasons for their interestingness and uninterestingness ratings from a multiple choice list (see Appendix~\ref{app:imo-reasons}), plus an additional free-text box to state their own reasons. Most problems only received one to two responses; this is too few to compare human and model judgments at a per-problem level. As such, IMO data comparisons are made over the interestingness \textit{criteria} that participants selected. 

One of our authors physically attended the IMO in 2024 and asked participants to complete a survey of their mathematics judgments. The survey received prior ethics approval by our institutional ethics review board. 

\subsection{List of Interestingness and Uninterestingness Reasons}
\label{app:imo-reasons}
\textbf{Interestingness reasons:}
\begin{itemize}
    \item The results we are asked to prove is connected to other interesting ideas/areas of maths.
    \item The ideas required to prove this problem are new.
    \item The ideas required to prove this problem are potentially useful in other problems too.
    \item The solution is elegant.
    \item The result we are asked to prove seems surprising and unexpected.
    \item The problem statement is simple and elegant.
    \item There are multiple very different ways to solve the problem.
    \item The problem seems very different from any problem I’ve seen before.
    \item The problem allows you to play around with examples and get good intuition.
    \item The solution does not require any sophisticated techniques/theorems.
    \item The problem statement is natural.
    \item None of the above apply.
\end{itemize}

\textbf{Uninterestingness reasons:}
\begin{itemize}
    \item The results that we are asked to prove seems irrelevant to other interesting areas of maths.
    \item The ideas required to prove this problem are too standard.
    \item The ideas required to prove this problem are not useful for solving other problems.
    \item The solution is not elegant.
    \item The result we are asked to prove seem expected hence uninteresting.
    \item The problem statement is too complicating.
    \item There is only one way to solve a problem.
    \item I’ve seen very similar problems before.
    \item The problem does not allow you to play around with examples and get good intuition.
    \item The solution requires sophisticated techniques/theorems.
    \item The problem statement is unnatural/contrived.
    \item None of the above apply.
\end{itemize}

\section{Problem Generation Study}
\label{app:prob_generation}
To begin to assess whether people find these problems interesting, we run an initial pilot problem evaluation study with Prolific participants, following our previous experimental design (to collect the judgments in RQ1-2), judging LLM-generated problems. Participants are only shown the problems and do not know the problems are LLMs generated to avoid bias. We choose LLMs along the range of human-alignment (as measured by WD in RQ2a): we use Mistral 7B instruct (the most human aligned), Qwen 235B Thinking (somewhat human aligned), and o3 (the least human aligned) as problem generators. This also allows us to assess whether models that are more aligned to people in their evaluations of problem interestingness are also better generators, or vice versa. As we sample 30 problems from each LLM, this creates a set of 90 problems, from which we manually filter invalid problems (e.g., problems with inconsistent information, no correct answer, etc.). This step removes 12 problems, all of which were generated by Mistral 7B Instruct. From these problems, we create 2 batches of 12 problems rated by 15 participants each (30 total participants and 24 total problems), resulting in 360 judgments we evaluate. 

We also ablate whether an automated problem verifier would also discard the problems as in our manual filtering, and find that the LLM we use (OpenAI's~O3) filters out the same cases. Specifically, we provide the following prompt:

\begin{verbatim}
You are a meticulous math contest grader.

Task:
1) Decide whether the problem statement is well-posed and solvable as 
written, with a determinate answer.
2) If it is NOT well-posed/solvable, set valid=false and give a concise 
reason (missing diagram/data, ambiguity, contradiction, non-unique, etc.). 
Do not attempt to guess missing info.
3) If it IS well-posed, solve it correctly and set valid=true and provide 
the final answer only in 'answer'.

Output must match the provided schema exactly.
\end{verbatim}
\subsection{t-SNE Visualization}
\label{app:tsne}
To create the t-SNE visualization in~\cref{fig:tsne}, we use text-embedding-3-small for embeddings, a perplexity of 15.0, and plot using cosine distance.

\section{Prompts to Humans and LLMs}
\label{app:human_llm_prompts}
\subsection{Survey Text - Prolific Study}
First, we include the pre-survey text seen by Prolific participants.
\begin{verbatim}
Welcome! We are conducting an experiment to understand how people think about 
math problems. Your answers will be used to inform cognitive science, 
mathematics, and AI research.

PLEASE ONLY PARTICIPATE IF YOU FIND MATH PROBLEMS INTERESTING. OTHERWISE 
THIS STUDY WILL NOT BE FUN! 

This experiment should take approximately 45 minutes.

You will be compensated at a base rate of $13.5/hour, with an optional bonus 
to bring the total up to a rate of $15/hr if you try your best throughout 
the experiment to answer each question

Please set the experiment to full screen.

We take your compensation and time seriously! The email for the main experimenter 
is [removed for anonymity]. 

Please write this down now, and email us with your Prolific ID and the subject 
line Human experiment compensation if you have problems submitting this task, 
or if it takes much more time than expected.

Please do not use any other Internet source or aid, including internet search, 
calculators, or chatbots — this experiment is designed to measure how people 
think. Unfortunately, if we find evidence that you did this, you may not 
receive the payment for this task.

There are two parts to this experiment. We present the instructions for Part I 
next. Instructions for a brief Part II will follow after you complete 
Part I.

In this experiment, you will be reading math problem statements and
answering a few questions about each problem.

For each problem, your task is to answer two questions.  

You will be asked to assess how interesting that problem is and how challenging
the problem is.

You can interpet interestigness however you wish.

You will answer the question by dragging a slider.

We also ask that you type your rationale for your judgment per problem (1-3 
sentences) in a text box.
  
Before you answer the question for each problem, you will have as much time 
as you want to think about the problem and its possible solutions. You must 
spend at least 60 seconds thinking about each problem. Please use scratch paper 
and focus while doing so.

After you feel like you understand the problem, please press the CONTINUE TO 
QUESTION to indicate that you are ready to answer the question.
  
We encourage you to take your time and carefully analyze the problem 
before providing your answer.

You will see a total of 10 problems.

When you are ready, please click "Next" to complete a quick comprehension 
check, before moving on to the experiment. 

Please make sure to window size is in full screen to properly view the questions.

\end{verbatim}

\subsection{Survey Text - IMO Study}
\begin{verbatim}
Your Task

The purpose of this survey is to study what mathematical olympiad problems people 
find interesting. In order to do so, we will ask you to judge the interestingness 
of various problems to get as detailed information as possible of what kinds of 
problem statements are interesting.

In the first part, you will identify some key notions which you think are 
most important for a maths olympiad problem to be interesting in general.

In the second part, you will rate the interestingness of four past problems from 
various IMO shortlists. You will evaluate how these specific problems are 
interesting or not interesting. 

Finally, you will be asked to self-evaluate what kind of a mathematician 
you are, which will provide more context for your interestingness judgement.

The rating process for each problem will be as follows: 
* First, you will read the problem statement, and decide whether you want to 
read the solution to the problem. This is to accommodate for people who want to 
get a gist of the solution before making a judgement of interestingness. 
You are welcome to skip this step if you like.
* You will then be asked to rate the interestingness of the problem, and 
elaborate on why the problem was interesting or not interesting.
* We will also ask how well you understand the problem, as well as how difficult 
you think the problem is, as this might have an effect on your judgement of 
interestingness.

The survey should take approximately 15 min. If you feel that you are taking 
too long, feel free to judge interestingness of problems just based on the 
problem statement without considering the solution.

Privacy and Data Usage

We very much value your privacy! We will not save any identifying information, 
beyond your self-reported level of mathematical olympiad expertise and your 
participation at the IMO. We plan to release these ratings anonymously 
open-source for other researchers. Please do not participate if you are not 
comfortable with those data sharing procedures. Please note that you are 
welcome to leave the study at any time, and that your participation 
in this research is voluntary. Please only proceed to the study if you are 
comfortable with the above, and acknowledge that you wish to participate in this research.

If you are comfortable with all of the above -- please read the instructions 
closely! -- then we welcome your participation in survey.

In this survey, you will be asked to judge the interestingness of the following 
four problems from various IMO shortlists. Before this, please tell us what 
you thinks is generally important for a maths Olympiad problem to be interesting.

Here are some possible notions of interestingness. Please indicate how important 
these factors generally are for a problem to be interesting to you.

[The following reasons each had a multiple choice selector that let participants 
choose one of the following: Very important, Quite important, Slightly important, 
Not important, and I don’t understand what this criteria means.]

1. The results we are asked to prove is connected to other interesting ideas/areas 
of mathematics.
2. The ideas required to prove this problem are novel.
3. The ideas required to prove this problem are potentially useful in other 
problems too.
4. The solution is elegant.
5. The result we are asked to prove seems surprising and unexpected.
6. The problem statement is simple and elegant.
7. There are multiple very different ways to solve the problem.
8. The problem seems very different from any problem I’ve seen before.
9. The problem allows you to play around with examples and get good intuition.
10. The solution does not require any sophisticated techniques/theorems.
11. The problem statement is natural.

[This was followed by all four problems being shown at once, and then each 
problem being shown one at a time, with the participant having the option to 
either view the solution or not.]

Would you like to see the solution to this problem before making a judgement? 
For geometry problems, choosing to see the solution will allow you to see the 
diagram too.

[0 - 7 scale of Least interesting problem I’ve ever seen - Most interesting 
problem I’ve ever seen]

What made this problem interesting? If this problem was not interesting at 
all, or if you think none of these reasons capture why this problem was 
interesting, tick the last option and move on.

- The results we are asked to prove is connected to other interesting ideas/areas 
of maths.  
- The ideas required to prove this problem are new.  
- The ideas required to prove this problem are potentially useful in other 
problems too.  
- The solution is elegant.  
- The result we are asked to prove seems surprising and unexpected.  
- The problem statement is simple and elegant.  
- There are multiple very different ways to solve the problem.  
- The problem seems very different from any problem I’ve seen before.  
- The problem allows you to play around with examples and get good intuition.  
- The solution does not require any sophisticated techniques/theorems.  
- The problem statement is natural.  
- None of the above apply.
   
Were there any other factors which made this problem interesting?
[open-ended response space]

What made this problem uninteresting for you? Please tick the last option if 
none of the other options apply.

- The results we are asked to prove seems irrelevant to other interesting areas 
of maths.  
- The ideas required to prove this problem are too standard.  
- The ideas required to prove this problem are not useful for solving other problems.  
- The solution is not elegant.  
- The result we are asked to prove seem expected hence uninteresting.  
- The problem statement is too complicating.  
- There is only one way to solve the problem.  
- I’ve seen very similar problems before.  
- The problem does not allow you to play around with examples and get good intuition.  
- The solution requires sophisticated techniques/theorems.  
- The problem statement is unnatural/contrived.  
- None of the above apply.  
   
Were there any other factors which made this problem uninteresting?
[open-ended response space]

Do you know the solution to this problem? [multiple choice]
No. My judgement is based purely on the problem statement.
Yes. I’ve solved the problem before.
Other:

How difficult is this problem?
0: too easy for an IMO
1: an easy IMO problem 1
2: a hard IMO problem 1
3: an easy IMO problem 2
4: a hard IMO problem 2
5: an easy IMO problem 3
6: a hard IMO problem 3
7: too hard for an IMO
Other:

\end{verbatim}
\subsection{LLM Prompt - Prolific Study}
Below, we include the prompt given to LLMs while rating problems from the Prolific study. The majority of it is similar to what the human participants saw.
\begin{verbatim}
In this experiment, you will be reading math problem statements
and answering a few questions about each problem.

For each problem, your task is to answer two questions.

You will be asked to assess how interesting that problem is and how 
challenging the problem is. The scale for both these ratings is between 0-100.

You can interpet interestigness however you wish.

We also ask that you type your rationale for your judgment per problem (1-3 
sentences).

We encourage you to take your time and carefully analyze the problem before 
providing your answer.

Your response should be in the following format.

### Interestingness: rating_between_0_and_100_here
### Int_rationale: 1-3 sentence rationale here
### Challenge: rating_between_0_and_100_here
### Chall_rationale: 1-3 sentence rationale here

{problem}
\end{verbatim}

\subsection{LLM Prompt - IMO Study}
Next, we include the prompt used to elicit interestingness and difficulty judgments for the IMO problems from LLMs.

\begin{verbatim}
In this experiment, you will be reading math problem statements and answering 
several questions about each problem.

For each problem, you will be asked to:

1. Rate how interesting the problem is on a scale of 0–7.  
    (0 = least interesting problem you’ve ever seen, 
    7 = most interesting problem you’ve ever seen)

2. Select reasons why you found the problem interesting 
(or select "None of the above" if none apply).  
   Options:  
   - The results we are asked to prove is connected to other interesting 
   ideas/areas of maths.  
   - The ideas required to prove this problem are new.  
   - The ideas required to prove this problem are potentially useful in 
   other problems too.  
   - The solution is elegant.  
   - The result we are asked to prove seems surprising and unexpected.  
   - The problem statement is simple and elegant.  
   - There are multiple very different ways to solve the problem.  
   - The problem seems very different from any problem I’ve seen before.  
   - The problem allows you to play around with examples and get good intuition.  
   - The solution does not require any sophisticated techniques/theorems.  
   - The problem statement is natural.  
   - None of the above apply.  

3. Select reasons why you found the problem uninteresting 
(or select "None of the above" if none apply).  
   Options:  
   - The results we are asked to prove seems irrelevant to other interesting 
   areas of maths.  
   - The ideas required to prove this problem are too standard.  
   - The ideas required to prove this problem are not useful for solving other problems.  
   - The solution is not elegant.  
   - The result we are asked to prove seem expected hence uninteresting.  
   - The problem statement is too complicating.  
   - There is only one way to solve the problem.  
   - I’ve seen very similar problems before.  
   - The problem does not allow you to play around with examples and get good 
   intuition.  
   - The solution requires sophisticated techniques/theorems.  
   - The problem statement is unnatural/contrived.  
   - None of the above apply.  

4. Rate how difficult the problem is on a scale of 0–7.  
   (0 = too easy for an IMO, 7 = too hard for an IMO)  

We encourage you to take your time and carefully analyze the problem before 
providing your answer.

Your response should be in the following format:

### Interestingness: rating_between_0_and_7_here
### Int_rationale: 1-3 sentence rationale here
### Interesting_reasons: list only the reasons that apply (ONLY from the provided 
options, exactly as they are written). Do not list reasons you don't agree with. 
### Uninteresting_reasons: list only the reasons that apply (ONLY from the provided 
options, exactly as they are written). Do not list reasons you don't agree with. 
### Challenge: rating_between_0_and_7_here
### Chall_rationale: 1-3 sentence rationale here

{problem}
\end{verbatim}
\subsection{LLM Prompt - Interestingness Reasons}
Below, we include the prompt used to prompt LLMs about the importance they assign to different reasons for interestingness.
\begin{verbatim}
Here are some possible notions of interestingness. 
For each criterion below, please indicate how important it generally is for a 
problem to be interesting.

Choose one of the following values for each criterion:
- Very important
- Quite important
- Slightly important
- Not important
- I don’t understand what this criteria means

Criteria:
1. The results we are asked to prove is connected to other interesting ideas/areas 
of mathematics.
2. The ideas required to prove this problem are novel.
3. The ideas required to prove this problem are potentially useful in other 
problems too.
4. The solution is elegant.
5. The result we are asked to prove seems surprising and unexpected.
6. The problem statement is simple and elegant.
7. There are multiple very different ways to solve the problem.
8. The problem seems very different from any problem I’ve seen before.
9. The problem allows you to play around with examples and get good intuition.
10. The solution does not require any sophisticated techniques/theorems.
11. The problem statement is natural.

Please respond in the following JSON format. Respond only from the options provided, 
nothing else. 
Don't return anything besides the JSON.

{
"criterion_1": "...",
"criterion_2": "...",
"criterion_3": "...",
"criterion_4": "...",
"criterion_5": "...",
"criterion_6": "...",
"criterion_7": "...",
"criterion_8": "...",
"criterion_9": "...",
"criterion_10": "...",
"criterion_11": "..."
}
\end{verbatim}

\newpage

\end{document}